# Provable Robustness Against a Union of $\ell_0$ Adversarial Attacks

Zayd Hammoudeh[*,1,2]     Daniel Lowd[1]

[1]*University of Oregon*
[2]*Qualtrics AI*

**Abstract**

Sparse or $\ell_0$ adversarial attacks arbitrarily perturb an unknown subset of the features. $\ell_0$ robustness analysis is particularly well-suited for heterogeneous (tabular) data where features have different types or scales. State-of-the-art $\ell_0$ certified defenses are based on randomized smoothing and apply to evasion attacks only. This paper proposes *feature partition aggregation* (FPA) – a certified defense against the union of $\ell_0$ evasion, backdoor, and poisoning attacks. FPA generates its stronger robustness guarantees via an ensemble whose submodels are trained on disjoint feature sets. Compared to state-of-the-art $\ell_0$ defenses, FPA is up to 3,000× faster and provides larger median robustness guarantees (e.g., median certificates of 13 pixels over 10 for CIFAR10, 12 pixels over 10 for MNIST, 4 features over 1 for Weather, and 3 features over 1 for Ames), meaning FPA provides the additional dimensions of robustness essentially for free.

## 1 Introduction

Machine learning models are vulnerable to numerous types of adversarial attacks, including (1) *evasion attacks* which manipulate a model by perturbing test instances [Sze+14], (2) *poisoning attacks* which manipulate predictions by perturbing a model's training set [BNL12], (3) *backdoor attacks* which combine training and test perturbations [Li+22], and (4) *patch attacks* – a specialized evasion attack where the adversarial perturbation is restricted to a specific shape [Bro+17]. *Certified defenses* provide provable guarantees of a prediction's robustness against adversarial attack.

This work focuses on $\ell_0$ or *sparse* attacks, where an adversary controls an unknown subset of the features. By certifying robustness w.r.t. the number of perturbed features, $\ell_0$ analysis is particularly well-suited to heterogeneous (tabular) data where the features have different types (e.g., numerical, categorical) or scales. Moreover, $\ell_0$ defenses provide provable robustness against real-world patch attacks [LF20a]. Several certified $\ell_0$ defenses have been proposed [Lee+19; LF20b; Cal+21; Jia+22b; LF22], but these methods apply to evasion only, which can be limiting. For example, consider a distributed sensor network where each (tabular) feature is independently measured by a different sensor. Under this type of *vertical partitioning* where features are sourced from multiple parties, an attacker that controls a single feature (i.e., sensor) can partially perturb every instance – training and test – up to 100% poisoning rate [LDD21; Wei+22]. Existing $\ell_0$ evasion defenses do not certify robustness over any training perturbation rendering them moot under such an attack. Moreover, existing $\ell_0$ defenses could not be combined with instance-wise poisoning defenses here since, typically, the latter are only provably robust under small poisoning rates, e.g., ≤1% [Rez+23].

To address these limitations, we propose *feature partition aggregation* (FPA) – a certified sparse defense jointly robust against both training and test feature perturbations. FPA uses a model ensemble approach, where each submodel is trained on a disjoint feature set, meaning any adversarially perturbed feature – training or test – affects at most one submodel prediction. Hence, FPA guarantees robustness over the *union* of $\ell_0$ evasion, backdoor, and poisoning attacks – a strictly stronger guarantee than existing $\ell_0$ methods [LF20b]. In our empirical evaluation, FPA's certified median guarantees are up to 4× larger than state-of-the-art $\ell_0$ defenses [Jia+22b] with little to no decrease in classification accuracy; FPA is also up to 3,000× faster. In other words, FPA provided additional dimensions of $\ell_0$ robustness essentially for free. Our primary contributions are summarized below; additional theoretical analysis and all proofs are in the supplement.

---

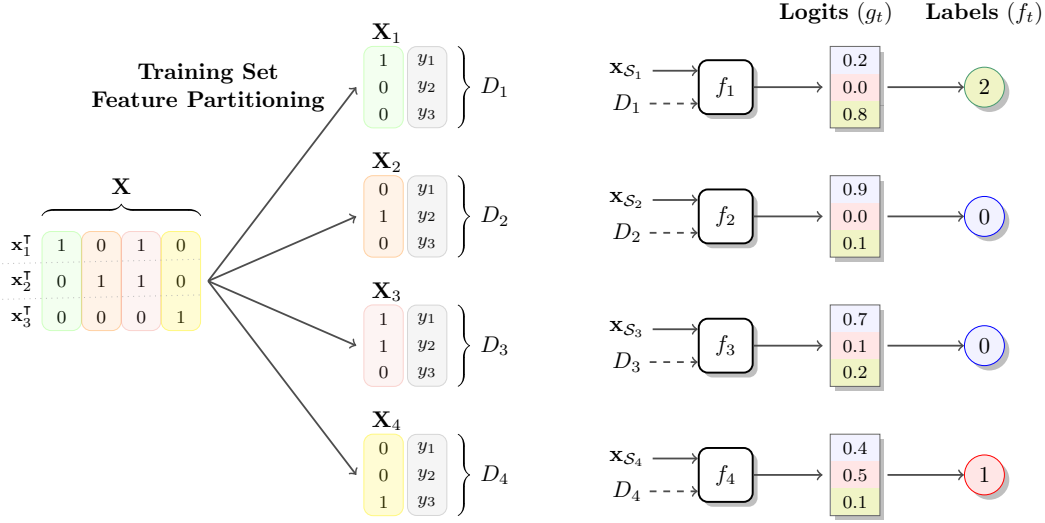

Figure 1: **Feature partition aggregation example** prediction for: test instance $\mathbf{x} \in \mathcal{X}$, $n = 3$, $d = 4$, and $|\mathcal{Y}| = 3$. Feature partitioning across $T = 4$ submodels, where the $t$-th submodel uses only feature dimensions $\mathcal{S}_t = \{t\} \subset [4]$ and training set $D_t$, i.e., the tuple containing the $t$-th column of feature matrix $\mathbf{X}$ (denoted $\mathbf{X}_t$) and label vector $\mathbf{y} := [y_1, y_2, y_3]$. $\mathbf{x}_{\mathcal{S}_t}$ denotes the subvector of $\mathbf{x}$ restricted to the feature dimensions in $\mathcal{S}_t$. Plurality label $y_{\text{pl}} = 0$; runner-up label $y_{\text{ru}} = 1$; and the predicted label with the run-off decision function is $y_{\text{RO}} = 0$. Under the plurality voting decision function (Sec. 4.1), $f(\mathbf{x})$ has certified feature robustness $r_{\text{pl}} = 0$. With the run-off decision function (Sec. 4.2), $f(\mathbf{x})$'s certified feature robustness is $r_{\text{RO}} = 1$.

- We define a new robustness paradigm we term *certified feature robustness* that generalizes $\ell_0$ (sparse) robustness to encompass training set feature perturbations.

- We propose feature partition aggregation, a certified feature defense that uses an ensemble of submodels trained on disjoint feature sets. We detail two certification schemes – a simple one based on plurality voting and the other based on multi-round elections.

- We empirically evaluate FPA on two classification and two regression datasets. FPA provided simultaneously larger and stronger median guarantees than the state-of-the-art certified $\ell_0$ defenses while also being 2 to 3 orders of magnitude faster.

## 2  Preliminaries

**Notation**  Supplemental Section A provides a full nomenclature reference. Let $[m]$ denote integer set $\{1, \ldots, m\}$. $\mathbb{1}[a]$ is the *indicator function*, which equals 1 if predicate $a$ is true and 0 otherwise. $\ell_0$ *norm* $\|\mathbf{w}\|_0$ is the number of non-zero elements in vector $\mathbf{w}$. Given some matrix $\mathbf{A}$, denote its $j$-th column as $\mathbf{A}_j$. In a slight abuse of notation, let $\mathbf{A} \ominus \mathbf{A}' := \{j : \mathbf{A}_j \neq \mathbf{A}'_j\}$ denote the set of column *indices* over which equal-size matrices $\mathbf{A}$ and $\mathbf{A}'$ differ. Similarly, let $\mathbf{v} \ominus \mathbf{v}' \subseteq [|\mathbf{v}|]$ denote the set of *dimensions* where vectors $\mathbf{v}$ and $\mathbf{v}'$ differ.

Let $\mathbf{x} \in \mathcal{X} \subseteq \mathbb{R}^d$ be a *feature vector* ($d := |\mathbf{x}|$) and $y \in \mathcal{Y} \subseteq \mathbb{N}$ a *label*. A *training set* $\{(\mathbf{x}_i, y_i)\}_{i=1}^n$ consists of $n$ instances. Denote the training set's *feature matrix* as $\mathbf{X} := [\, \mathbf{x}_1 \, \cdots \, \mathbf{x}_n \,]^\intercal$ where $\mathbf{X} \in \mathbb{R}^{n \times d}$, and denote the label vector $\mathbf{y} := [y_1, \ldots, y_n]$. Let $f : \mathcal{X} \to \mathcal{Y}$ be a *model*.

For our method, feature partition aggregation (FPA), $f$ is an ensemble of $T$ *submodels* (see Figure 1). A *decision function* aggregates the $T$ submodel predictions to form $f$'s overall prediction. The model architecture and decision function combined dictate how a prediction's *certified robustness* is calculated. For instance $(\mathbf{x}, y)$, let $g_t(\mathbf{x}, y)$ be the $t$-th submodel's *logit* value for label $y$, where $g_t : \mathcal{X} \times \mathcal{Y} \to \mathbb{R}$. Let $f_t(\mathbf{x})$ denote the $t$-th submodel's predicted *label* for $\mathbf{x}$, where $f_t : \mathcal{X} \to \mathcal{Y}$ and $f_t(\mathbf{x}) := \arg\max_{y \in \mathcal{Y}} g_t(\mathbf{x}, y)$. Throughout this work, all ties are broken by selecting the label with the smallest index.

*Feature set* $[d]$ is *partitioned* across FPA's $T$ submodels. Let $\mathcal{S}_t \subset [d]$ be the features used by the $t$-th submodel where $\bigsqcup_{t=1}^T \mathcal{S}_t = [d]$. In other words, each FPA submodel considers a fixed, disjoint subset of the features for all training and test instances. The $t$-th submodel's training set, $D_t$, consists of: label vector $\mathbf{y}$ and the $\mathcal{S}_t$ columns in $\mathbf{X}$. FPA submodels are *deterministic*, meaning fixing $D_t$, $\mathcal{S}_t$, and $\mathbf{x}$, in turn, fixes label $f_t(\mathbf{x})$ and logits $\forall_y\, g_t(\mathbf{x}, y)$.

Given $\mathbf{x}$ and $y$, the pointwise *submodel vote count* is $\dot{c}_y(\mathbf{x}) := \sum_{t=1}^{T} \mathbb{1}[f_t(\mathbf{x}) = y]$. The *plurality* and *runner-up* labels receive the most and second-most votes (resp.), i.e., $y_{\text{pl}} = \arg\max_{y \in \mathcal{Y}} \dot{c}_y(\mathbf{x})$ and $y_{\text{ru}} = \arg\max_{y \in \mathcal{Y} \setminus y_{\text{pl}}} \dot{c}_y(\mathbf{x})$. The pointwise *submodel vote gap* between labels $y, y' \in \mathcal{Y}$ is

$$\text{GAP}_{\text{vote}}(y, y'; \mathbf{x}) := \dot{c}_y(\mathbf{x}) - \dot{c}_{y'}(\mathbf{x}) - \mathbb{1}[y' < y], \tag{1}$$

with the indicator function used to break ties. Let $\ddot{c}_y(\mathbf{x}; y') := \sum_{t=1}^{T} \mathbb{1}[g_t(\mathbf{x}, y) > g_t(\mathbf{x}, y')]$ be $y$'s *logit vote count* w.r.t. $y' \in \mathcal{Y}$. The pointwise *logit vote gap* for $y$ w.r.t. $y'$ is

$$\text{GAP}_{\text{logit}}(y, y'; \mathbf{x}) := \ddot{c}_y(\mathbf{x}; y') - \ddot{c}_{y'}(\mathbf{x}; y) - \mathbb{1}[y' < y]. \tag{2}$$

Below, $\mathbf{x}$ is dropped from $\text{GAP}_{\text{vote}}$ and $\text{GAP}_{\text{logit}}$ when the feature vector of interest is clear from context.

**Threat Model** Given arbitrary $(\mathbf{x}, y)$, the attacker's objective is for $y \neq f(\mathbf{x})$. The adversary achieves this objective via two methods: (1) modify training features $\mathbf{X}$ or (2) modify test instance $\mathbf{x}$'s features.[1] An adversary may use either method individually or both methods jointly. An attacker can *perturb up to 100% of the training instances*.

**Our Objective** Given $(\mathbf{x}, y)$, determine the *certified feature robustness*, $r$ (defined below). *Pointwise* guarantees certify the robustness of each instance $(\mathbf{x}, y)$ individually.

**Def. 1. Certified Feature Robustness** *Given training set $(\mathbf{X}, \mathbf{y})$, model $f'$ trained on $(\mathbf{X}', \mathbf{y})$, and arbitrary feature vector $\mathbf{x}' \in \mathcal{X}$, certified feature robustness $r \in \mathbb{N}$ is a pointwise, deterministic guarantee w.r.t. instance $(\mathbf{x}, y)$ where $|\mathbf{X} \ominus \mathbf{X}' \cup \mathbf{x} \ominus \mathbf{x}'| \leq r \implies y = f'(\mathbf{x}')$.*

Certified robustness $r$ is <u>not</u> w.r.t. individual feature values. Rather, certified feature robustness provides a stronger guarantee allowing all values of a feature – training and test – to be perturbed.

## 3 Related Work

FPA marries ideas from two classes of certified adversarial defenses, which are discussed below. A more detailed discussion of related work is deferred to suppl. Section C.

$\ell_0$**-Norm Certified Evasion Defenses** Representing the work most closely related to ours, these methods certify $\ell_0$*-norm robustness* (also known as "*sparse robustness*"), which we formalize below.

**Def. 2. $\ell_0$-Norm Certified Robustness** *Given model $f$, $\alpha \in (0, 1)$, and arbitrary feature vector $\mathbf{x}' \in \mathcal{X}$, $\ell_0$-norm certified robustness $\rho \in \mathbb{N}$ is a pointwise guarantee w.r.t. instance $(\mathbf{x}, y)$ where if $\|\mathbf{x} - \mathbf{x}'\|_0 \leq \rho$, then $y = f(\mathbf{x}')$ with probability at least $1 - \alpha$.*

There are two main differences between certified $\ell_0$-norm robustness (Def. 2) and our certified feature robustness (Def. 1). (1) $\ell_0$-norm methods are not certifiably robust against any adversarial training perturbations (e.g., poisoning and backdoors). (2) $\ell_0$-norm robustness guarantees are *probabilistic*, while our feature guarantees are deterministic. Put simply, our certified feature guarantees are *strictly stronger* than $\ell_0$-norm guarantees.

*Randomized ablation* (RA) is the state-of-the-art certified $\ell_0$-norm defense [LF20b; Jia+22b]. RA adapts ideas from *randomized smoothing* [CRK19] to $\ell_0$ evasion attacks. Specifically, RA creates a single *smoothed classifier* by repeatedly evaluating different *ablated inputs*, each of which *keeps* a random subset of the features unchanged and masks out (*ablates*) all other features. RA's *ablated training* generally permits only stochastically trained, parametric model architectures. At inference, certifying a single prediction with RA requires evaluating up to 100k ablated inputs [Jia+22b]. Jia et al. [Jia+22b] improve RA's guarantees via novel certification analysis that is tight for top-1 predictions, meaning Jia et al.'s version of RA always performs at least as well as the original. Jia et al. [Jia+22b] also extend RA to certify $\ell_0$-norm robustness for top-$k$ predictions.

*Certified patch robustness* is a restricted form of $\ell_0$-norm robustness where the perturbed test features are constrained to a specific, contiguous shape, e.g., square [MY21]. Existing patch defenses include (de)randomized smoothing (DRS) [LF20a] – a specialized version of randomized ablation for patch attacks. Like RA, DRS performs ablated training and inference. By assuming a single patch shape, the number of possible attacks becomes linear in $d$, allowing DRS to only evaluate $\mathcal{O}(d)$ ablations during inference; this derandomizes the ablation set, making DRS's patch guarantees deterministic.[2] More recently, Metzen and Yatsura [MY21] propose BAGCERT – a certified patch defense that is less sensitive to patch shape than DRS. Note any certified feature or $\ell_0$-norm defense (e.g., FPA, RA) is also a certified patch defense, given the former's stronger guarantees.

---

[1] Our primary threat model assumes a *clean-label attacker* that does not modify training labels. Suppl. Sec. E provides additional theoretical results for an adversary that modifies training labels.

**Instance-wise Certified Poisoning Defenses** The second class of related defenses certify robustness under the arbitrary insertion or deletion of entire *instances* in the training set [Che+22] – generally a small poisoning rate (e.g., ≤1%). Like FPA, most instance-wise poisoning defenses are voting-based [JCG21; Jia+22a; WLF22a]. For example, *deep partition aggregation* (DPA) randomly partitions the training *instances* across an ensemble of $T$ submodels [LF21]. More recently, Rezaei et al. [Rez+23] propose *run-off elections*, a novel decision function for DPA that can improve DPA's certified robustness by several percentage points. While certified instance-wise poisoning defenses show promise, they are still vulnerable to test perturbations – even of a single feature.

## 4 Certifying Feature Robustness

Our certified defense, feature partition aggregation (FPA), can be viewed as the *transpose* of Levine and Feizi's [LF21] deep partition aggregation (DPA). Both defenses are (1) ensembles, (2) rely on voting-based decision functions, and (3) partition the training set; the key difference is in the partitioning operation. DPA horizontally partitions the set of training *instances* (rows of feature matrix $\mathbf{X}$), enabling DPA to certify *instance-wise* robustness. In contrast, FPA vertically partitions along an orthogonal dimension – the feature set (columns of $\mathbf{X}$) – enabling FPA to certify *feature-wise* robustness. Intuitively, *partitioning along orthogonal dimensions means that DPA and FPA certify orthogonal types of robustness*. Training FPA submodels on disjoint feature subsets (e.g., Figure 1) entails that a perturbed feature affects, at most, one submodel prediction. FPA leverages this property to certify feature robustness $r$. Below we describe two FPA *decision functions*: (1) a simpler scheme using plurality voting and (2) an enhanced multi-round voting procedure specialized for multiclass classification. The decision function combined with FPA's architecture dictates how our robustness guarantee is calculated.

### 4.1 Feature Robustness Under Plurality Voting

For $\mathbf{x} \in \mathcal{X}$, the *plurality voting* decision function defines the model prediction as $f(\mathbf{x}) \coloneqq y_{\text{pl}}$, i.e., the label that receives the most submodel votes. A successful attack requires perturbing enough submodels to change $y_{\text{pl}}$. Specifically, each submodel perturbation decreases the submodel vote gap (GAP$_{\text{vote}}$) between $y_{\text{pl}}$ and the adversary's selected label by two. Hence, the minimum number of submodel perturbations equals half the vote gap between $y_{\text{pl}}$ and runner-up label $y_{\text{ru}}$. Theorem 3 formalizes this idea as a deterministic feature robustness guarantee. Eq. (3)'s decomposed form is similar to other voting-based certified defenses, including DPA [LF21; Jia+22a].

**Theorem 3. Certified Feature Robustness with Plurality Voting** *For feature partition $\mathcal{S}_1, \ldots, \mathcal{S}_T$, let $f$ be an ensemble of $T$ submodels using the plurality-voting decision function, where the $t$-th submodel uses the features in $\mathcal{S}_t$. For instance $(\mathbf{x}, y)$, the pointwise certified feature robustness is*

$$r_{\text{pl}} \coloneqq \left\lfloor \frac{\text{GAP}_{\text{vote}}(y_{\text{pl}}, y_{\text{ru}})}{2} \right\rfloor. \tag{3}$$

**Understanding Theorem 3 More Intuitively** Let $\mathcal{A}_{\text{tr}} \subseteq [d]$ be the set of features (i.e., dimensions) an attacker modified in the training set, and let $\mathcal{A}_{\mathbf{x}} \subseteq [d]$ be the set of features the attacker modified in instance $\mathbf{x}$. As long as $|\mathcal{A}_{\text{tr}} \cup \mathcal{A}_{\mathbf{x}}| \leq r$, the adversarial perturbations did not change the model prediction. The union over the perturbed feature sets entails that a feature perturbed in both training and test counts only once against guarantee $r$. Theorem 3's certified guarantees are implicitly agnostic to the $\ell_0$ attack type. Certified feature robustness $r$ applies equally to an $\ell_0$ evasion attack ($\mathcal{A}_{\mathbf{x}}$ only) as it does to $\ell_0$ poisoning ($\mathcal{A}_{\text{tr}}$ only). Theorem 3's guarantees also encompass more complex $\ell_0$ backdoor attacks ($\mathcal{A}_{\text{tr}} \cup \mathcal{A}_{\mathbf{x}}$).

Against a *worst-case adversary* where a feature perturbation arbitrarily changes the corresponding submodel's prediction, Eq. (3)'s guarantee is tight under plurality-voting.

**Top-$k$ Certified Feature Robustness** In top-$k$ predictions, a classifier predicts $k$ labels for each instance $\mathbf{x}$, with the accuracy calculated based on whether $\mathbf{x}$'s true label is among the $k$ predicted labels. In line with Jia et al.'s [Jia+22b] extension of RA to top-$k$ predictions, supplemental Section D extends FPA with plurality voting to certify top-$k$ feature robustness.

---

[2](De)randomized smoothing's deterministic guarantees do not scale to RA which considers $\mathcal{O}(2^d)$ possible attacks.

## 4.2 Feature Robustness Under Run-Off Elections

Under plurality voting, only submodels that predict either $y_{\text{pl}}$ or $y_{\text{ru}}$ are considered when determining the certified feature robustness (Eq. (3)). In other words, submodels predicting other labels essentially contribute nothing to plurality voting's pointwise guarantees. Decision functions that leverage these "wasted" submodels may certify larger guarantees (see Figure 1). For instance, Rezaei et al. [Rez+23] propose *run-off elections*, an enhanced two-round DPA decision function for multiclass classification.[3] Since FPA and DPA share the same basic architecture (excluding the partitioning dimension), run-off can be directly combined with FPA to improve our certified robustness.

We now describe run-off. Our presentation is mostly similar to Rezaei et al.'s [Rez+23] beyond standardizing the formulation to align with previous work. Formally, run-off's decision function procedure is:

**Round #1**: Determine plurality and runner-up labels $y_{\text{pl}}$ and $y_{\text{ru}}$ (resp.) as above.

**Round #2**: Set *run-off prediction* $y_{\text{RO}}$ to either label $y_{\text{pl}}$ or $y_{\text{ru}}$ based on the logit vote gap where

$$f(\mathbf{x}) = y_{\text{RO}} \coloneqq \begin{cases} y_{\text{pl}} & \text{GAP}_{\text{logit}}(y_{\text{pl}}, y_{\text{ru}}) \geq 0 \\ y_{\text{ru}} & \text{Otherwise} \end{cases}. \qquad (4)$$

Under run-off, ensemble prediction $y_{\text{RO}}$ can only be perturbed in two ways: (1) overtake $y_{\text{RO}}$ in round #2 or (2) eject $y_{\text{RO}}$ from round #1's top-two labels, preventing $y_{\text{RO}}$ reaching round #2. Run-off's robustness is lower bounded by whichever of these two cases takes fewer submodel perturbations. Each case is analyzed separately below; Theorem 4 then combines the analyses to form run-off's overall robustness $r_{\text{RO}}$.

**Case #1: Overtake $y_{\text{RO}}$ in Round #2** Let $\widetilde{y}_{\text{RO}} \coloneqq \{y_{\text{pl}}, y_{\text{ru}}\} \setminus y_{\text{RO}}$ denote the label not selected in round #2. For a label $y$ to overtake $y_{\text{RO}}$ in round #2, $y$ must simultaneously satisfy two requirements: (a) be in round #1's top-two labels (in turn ejecting $\widetilde{y}_{\text{RO}}$ from the top two) and (b) receive more logit votes than $y_{\text{RO}}$ in round #2. Hence, the certified robustness for this case is bounded by whichever of these requirements requires more feature perturbations. Therefore, an attacker may control up to

$$r_{\text{RO}}^{\text{Case1}} \coloneqq \min_{y \in \mathcal{Y} \setminus y_{\text{RO}}} \max \left\{ \left\lfloor \frac{\text{GAP}_{\text{vote}}(\widetilde{y}_{\text{RO}}, y)}{2} \right\rfloor, \left\lfloor \frac{\text{GAP}_{\text{logit}}(y_{\text{RO}}, y)}{2} \right\rfloor \right\} \qquad (5)$$

features, without $y_{\text{RO}}$ being overtaken in round #2 (Lemma 6).

**Case #2: Eject $y_{\text{RO}}$ from Round #1's Top-Two Labels** In round #1, run-off prediction $y_{\text{RO}}$ is preferred over label $y$ iff $\text{GAP}_{\text{vote}}(y_{\text{RO}}, y) \geq 0$ (Lemma 5). For $y_{\text{RO}}$ to qualify for run-off's second round, then for any pair of labels $y, y' \in \mathcal{Y} \setminus y_{\text{RO}}$ either $\text{GAP}_{\text{vote}}(y_{\text{RO}}, y) \geq 0$ or $\text{GAP}_{\text{vote}}(y_{\text{RO}}, y') \geq 0$. Calculating case #2's certified robustness reduces to determining the maximum number of submodels that can be perturbed with the above property still holding for all $\binom{|\mathcal{Y}|}{2}$ label pairs.

Recall that perturbing a submodel's vote from $y_{\text{RO}}$ to $y$ decreases $\text{GAP}_{\text{vote}}(y_{\text{RO}}, y)$ by 2; this submodel perturbation also decreases $\text{GAP}_{\text{vote}}(y_{\text{RO}}, y')$ by 1 $\forall y' \in \mathcal{Y} \setminus \{y_{\text{RO}}, y\}$. We leverage this simple insight to determine case #2's certified robustness. Formally, let $\mathtt{dp}$ be a function that takes two submodel vote gaps (e.g., $\Delta, \Delta' \in \mathbb{N}$) and returns a lower bound on the number of possible submodel perturbations where either $\text{GAP}_{\text{vote}}(y_{\text{RO}}, y) \geq 0$ or $\text{GAP}_{\text{vote}}(y_{\text{RO}}, y') \geq 0$. Applying the insight above, suppl. Lemma 7 shows that

$$\mathtt{dp}[\Delta, \Delta'] = 1 + \min\{\mathtt{dp}[\Delta - 2, \Delta' - 1], \mathtt{dp}[\Delta - 1, \Delta' - 2]\}. \qquad (6)$$

Eq. (6)'s base case sets $\mathtt{dp}[\Delta, \Delta'] = 0$ when $\max\{\Delta, \Delta'\} \leq 1$ and $(\Delta, \Delta') \neq (1, 1)$; this ensures the vote gap non-negativity condition always holds for at least one of the two labels of interest.[4]

A worst-case adversary attacks whichever label pair, $(y, y')$, requires the fewest perturbations, making case #2's overall robustness

$$r_{\text{RO}}^{\text{Case2}} \coloneqq \min_{y, y' \in \mathcal{Y} \setminus y_{\text{RO}}} \mathtt{dp}\big[\text{GAP}_{\text{vote}}(y_{\text{RO}}, y), \text{GAP}_{\text{vote}}(y_{\text{RO}}, y')\big]. \qquad (7)$$

Eq. (6)'s recursive formulation is solvable using classic dynamic programming. $\mathcal{O}(T^2)$-space matrix $\mathtt{dp}$ is prepopulated once for all $\mathbf{x}$, making $r_{\text{RO}}^{\text{Case2}}$'s amortized time complexity $\mathcal{O}(|\mathcal{Y}|^2)$.

---

[3]Run-off only changes the decision function; no training or model architecture changes are required.

[4]This base case differs slightly from Rezaei et al.'s [Rez+23] run-off formulation to ensure our robustness guarantee is tight against a worst-case adversary.

**Combining Cases #1 and #2 to Certify Feature Robustness**  Theorem 4 provides the certified feature robustness for an FPA prediction using the run-off decision function. Intuitively, an optimal attacker selects whichever of the two cases above requires fewer feature perturbations; hence, Eq. (8) below takes the minimum of $r_\text{RO}^\text{Case1}$ and $r_\text{RO}^\text{Case2}$.

**Theorem 4. Certified Feature Robustness with Run-off**  *For feature partition $\mathcal{S}_1, \ldots, \mathcal{S}_T$, let $f$ be an ensemble of $T$ submodels using the run-off decision function, where the $t$-th submodel uses only the features in $\mathcal{S}_t$. Then, for instance $(\mathbf{x}, y)$, the pointwise certified feature robustness is*

$$r_\text{RO} = \min\{r_\text{RO}^\text{Case1}, r_\text{RO}^\text{Case2}\}. \tag{8}$$

## 4.3 Advantages of Feature Partition Aggregation

Below, we summarize FPA's advantages over state-of-the-art certified $\ell_0$-norm defense randomized ablation (RA). These advantages apply irrespective of whether FPA uses plurality voting or run-off.

> (1) **Stronger Guarantees**  FPA's certified feature robustness guarantee (Def. 1) is strictly stronger than RA's $\ell_0$-norm guarantee (Def. 2). First, FPA's guarantees apply equally to $\ell_0$ evasion, poisoning, and backdoor attacks while RA only applies to evasion. Second, FPA's guarantees are deterministic while RA's guarantees are only probabilistic.
>
> (2) **Faster**  RA requires up to 100k forward passes to certify one prediction. FPA requires only $T$ forward passes – one for each submodel – where $T < 200$ in general. FPA certification is, therefore, orders of magnitude faster than RA.
>
> (3) **Model Architecture Agnostic**  RA's feature ablation is specialized for parametric models like neural networks and generally prevents the use of tree-based models like gradient-boosted decision trees (GBDTs). By contrast, FPA supports any submodel architecture.

# 5 Feature Partitioning Strategies

The certification analysis above holds irrespective of the feature partitioning strategy. However, how the features are partitioned can have a *major* impact on the size of FPA's certified guarantees. Below, we very briefly describe two insights into the properties of good feature partitions.

**Insight #1**  *Ensure sufficient feature information is available to each submodel.* Each incorrect submodel or logit vote cancels out a correct vote, meaning the goal should be to maximize the number of correct submodel predictions while simultaneously minimizing incorrect ones. In other words, robustness is maximized when all submodels perform well, and feature information is divided equally.

**Insight #2**  *Limit information loss due to feature partitioning.* Models use (implicit) feature interaction information when making a prediction. Intuitively, if a pair of features is assigned to different FPA submodels, none of the submodels can use these features' pairwise interaction during inference. Put simply, feature partitioning causes some feature (interaction) information to be completely lost. Fixing $T$, some feature partitions are more lossy than others, and good partitions limit the total information lost.

## 5.1 Feature Partitioning Paradigms

Applying the above insights, we propose two general feature partitioning paradigms. In practice, the partitioning strategy is essentially a hyperparameter tunable on validation data. The validation set need not be clean so long as the perturbations are representative of the test distribution.

**Balanced Random Partitioning**  Given no domain-specific knowledge, each feature's expected information content is equal. *Balanced random partitioning* assigns each submodel a disjoint feature subset sampled uniformly at random, with subsets differing in size by at most one. Random partitioning has two primary benefits. First, each submodel has the same a priori expected information content. Second, random partitioning can be applied to any dataset. FPA with random partitioning is usually a good initial strategy and empirically performs quite well.

**Deterministic Partitioning** One may have application-related insights into quality feature partitions. For example, consider feature partitioning of images. Features (i.e., pixels) in an image are ordered, and that structure can be leveraged to design better feature partitions. Often the most salient features are clustered in an image's center. To ensure all submodels are high-quality, each submodel should be assigned as many highly salient features as possible. Moreover, adjacent pixels can be highly correlated, i.e., contain mostly the same information. Given a fixed set of pixels to analyze, the information contained in those limited features should be maximized, so a good strategy can be to select a subset of pixels spread uniformly across the image. Put simply, for images, random partitioning can have larger information loss than deterministic strategies.

Supplemental Section H.7 empirically compares random and deterministic partitioning. In summary, a simple strided strategy that distributes features regularly across an image tends to work well for vision. Formally, given $d$ pixels and $T$ submodels, the $t$-th submodel's feature set under *strided partitioning* is $\mathcal{S}_t = \{j \in [d] : j \bmod T = t - 1\}$.

## 5.2 Beyond Partitioned Feature Subsets

Everything above should *not* be interpreted to imply that FPA necessarily requires partitioned feature sets. Submodel feature sets can (partially) overlap, but determining optimal $r$ under overlapping feature sets is NP-hard in general. Supplemental Section F extends FPA to overlapping feature sets and provides an empirical comparison. In summary, overlapping submodel feature sets can marginally outperform random partitioning but often lags deterministic partitions.

# 6 Evaluation

Our empirical evaluation is modeled after Levine and Feizi's [LF20b] evaluation of randomized ablation. Due to space, additional results are deferred to supplement Section H including: the base (non-robust) accuracy for each dataset (H.1), full numerical results (H.2 and H.3), hyperparameter sensitivity analysis (H.4 and H.5), plurality voting vs. run-off comparison (H.6), random vs. deterministic feature partitioning comparison (H.7), and model training times (H.8). Our source code is available at https://github.com/ZaydH/feature-partition.

## 6.1 Experimental Setup

Most evaluation setup details are deferred to supplemental Section G with a brief summary below. We evaluate FPA with both the plurality-voting (Section 4.1) and run-off (Section 4.2) decision functions.

**Baselines** Randomized ablation (RA) is FPA's most closely related work and the primary baseline below. We report the performance of both Levine and Feizi's [LF20b] original version of RA (denoted "(LF'20b)") and Jia et al.'s [Jia+22b] improved version (denoted "(Jia'22b)") where the certification analysis is tight for top-1 predictions. RA performs feature ablation during training and inference. Each ablated input keeps $e$ randomly selected features unchanged and masks out the remaining $(d - e)$ features; RA evaluates up to 100k ablated inputs to certify each prediction. Recall that RA's $\ell_0$-norm robustness only applies to evasion attacks (Def. 2), while FPA provides strictly stronger guarantees covering both training and test perturbations (Def. 1).

We also compare FPA to three certified patch defenses: *(de)randomized smoothing* [LF20a], *patch interval bound propagation* (IBP) [Chi+20b], and BAGCERT [MY21].

**Performance Metrics** Certified defenses generally trade-off robustness and (clean) accuracy. Hence, following Levine and Feizi's [LF20b] evaluation of randomized ablation, performance is primarily measured using two complementary metrics: (1) *median certified robustness*, the median value of the certified robustness across a dataset's entire test set with misclassified instances assigned robustness $-\infty$ and (2) *classification accuracy*, the fraction of test predictions classified correctly. Below, $r_{\text{med}}$ and $\rho_{\text{med}}$ denote the median certified feature robustness (Def. 1) and median $\ell_0$-norm robustness (Def. 2), respectively.

*Mean certification time* measures the time to certify a single prediction. *Certified accuracy* is the fraction of correctly-classified test instances that satisfy some specific robustness criterion; this criterion can be patch robustness or certified robustness of at least $\psi \in \mathbb{N}$.

**Datasets** We compare the methods on standard datasets used in data poisoning evaluation. First, following Levine and Feizi's [LF20b] evaluation of baseline RA, we consider MNIST and CIFAR10[5] where each feature corresponds to one (RGB) pixel. Second, Hammoudeh and Lowd [HL23b] prove that certified regression *reduces* to certified *binary* classification when median is used as the regressor's decision function (see Section G.6 for details). We apply their reduction to both FPA and RA where for instance $(\mathbf{x}, y)$ and hyperparameter $\xi \in \mathbb{R}_{\geq 0}$, the goal is to certify that $y - \xi \leq f(\mathbf{x}) \leq y + \xi$. We consider two tabular regression datasets evaluated by Hammoudeh and Lowd [HL23b]. (1) Weather [Mal+21] predicts the temperature using features such as date, longitude, and latitude ($\xi = 3°C$). (2) Ames [De 11] predicts housing prices using features such as square footage ($\xi = 15\%y$). These two regression datasets serve as a stand-in for vertically partitioned data, which are commonly tabular and, as Section 1 mentions, particularly vulnerable to our union of $\ell_0$ attacks threat model. Note run-off and plurality voting are identical under binary classification so we only report FPA's plurality voting regression results.

---

[5]Existing certified poisoning defenses do not evaluate on full ImageNet due to the high training cost [Web+20; LF21; Jia+22a; WLF22a; WLF22b; Rez+23].

**Model Architectures**  For vision datasets CIFAR10 and MNIST, all methods used convolutional neural networks. Baseline randomized ablation models were trained using Levine and Feizi's [LF20a] published source code. Since FPA requires training multiple submodels, FPA uses small, simple CNNs that are fast to train. Specifically, the compact ResNet9 [Pag20] architecture was used for CIFAR10, allowing submodels to be trained from scratch in as little as 60 seconds [Col+17]. For MNIST, we follow Levine and Feizi's [LF21] evaluation of instance-wise poisoning defense DPA and use the Network-in-Network architecture (NiN) [LCY14].

Gradient-boosted decision trees (GBDTs) generally work exceptionally well on tabular data [BHL23] so for regression datasets Weather and Ames, FPA used LightGBM GBDTs [Ke+17]. In contrast, RA's feature ablation prevents the use of tree-based models like GBDTs, so RA instead used linear models for these two datasets (Hammoudeh and Lowd [HL23b] also used linear models for Weather). Even when restricted to linear submodels, FPA still had better median robustness and classification accuracy than RA; see suppl. Tables 24 and 25.

**Feature Partitioning Strategy**  Supplemental Section H.7 shows that deterministic, strided feature partitioning significantly outperforms random partitioning on vision data. Specifically, deterministic partitioning improved FPA's certified accuracy by up to 15.6 percentage points (pp) for CIFAR10 and 11.9pp for MNIST. Hence, this section exclusively reports FPA's performance on these two datasets using strided partitioning, with each submodel considering the full image dimensions and any pixels not in $\mathcal{S}_t$ set to 0.

Section H.7 shows no consistent advantage when using deterministic partitioning for unordered tabular features. As such, for Weather and Ames, this section reports FPA's performance using balanced random partitioning.

**Hyperparameters**  Hyperparameters $T$ (FPA's submodel count) and $e$ (RA's kept feature count) control the corresponding method's robustness vs. accuracy tradeoff. When optimizing patch and median robustness, hyperparameters $T$ and $e$ were tuned on validation data.[6]

**Patch Robustness**  We consider two CIFAR10 patch attacks: (1) a $5 \times 5$ pixel square [LF20a] and (2) all 24-pixel rectangles (e.g., $1 \times 24$ pixels, $24 \times 1$, $2 \times 12$, etc.), reporting each method's minimum and maximum certified accuracies across the eight valid shapes [MY21].

**Hardware Requirements**  Both FPA and baseline RA have minimal hardware requirements. Experiments in this section were performed on a consumer desktop system containing an NVIDIA 3090 GPU, AMD 5950X CPU, and 64GB of RAM.

## 6.2 Main Results

Tables 1 and 2 summarize the median certified robustness and classification accuracy (resp.) for FPA and baseline RA. Table 3 details each method's mean certification time. Due to space, Tables 2 and 3 only report results for Jia et al.'s [Jia+22b] (significantly) better performing version of baseline RA. Table 4 analyzes FPA as a patch defense. We briefly summarize the experiments' takeaways below. See supplemental Sections H.2 and H.3 for the full numerical results, including comparing the methods at additional robustness levels.

**Takeaway #1**  *FPA simultaneously provided larger and stronger median robustness guarantees than RA.* As Table 1 details, FPA's median certified robustness was 20–30% larger than RA for classification and 3 to 4× larger for regression. Importantly, FPA's certified feature guarantees apply to evasion, poisoning, and backdoor attacks, while baseline RA only covers evasion attacks.

---
[6]Sections H.2 and H.3 compare each method's certified accuracy across a range of hyperparameter settings.

Table 1: **Median certified robustness** (larger is better). Each dataset's best performing method is in **bold**. Our median robustness was 20–30% larger for classification and 3 to 4× larger for regression while simultaneously providing stronger guarantees. For detailed results, see Section H.2.

Table 2: **Classification accuracy** (% – larger is better). We report FPA's accuracy at both RA's (middle, **bold**) and FPA's (blue) best median robustness levels. At RA's best median robustness, FPA had better classification accuracy for all four datasets. For full results, see Section H.2.

| Dataset | Dim. ($d$) | FPA (ours) | | Random. Ablate. | |
|---|---|---|---|---|---|
| | | Plural | Run-Off | (LF'20b) | (Jia'22b) |
| CIFAR10 | 1024 | 11 | **13** | 7 | 10 |
| MNIST | 784 | 9 | **12** | 8 | 10 |
| Weather | 128 | **4** | – | 0 | 1 |
| Ames | 352 | **3** | – | 1 | 1 |

| Dataset | FPA (ours) | | | | RA (Jia'22b) | |
|---|---|---|---|---|---|---|
| | $r_{med}$ | Acc. | $r_{med}$ | Acc. | $\rho_{med}$ | Acc. |
| CIFAR10 | 13 | 62.4 | 10 | **75.0** | 10 | 64.7 |
| MNIST | 12 | 87.2 | 10 | **96.1** | 10 | 93.1 |
| Weather | 4 | 76.1 | 1 | **85.3** | 1 | 75.2 |
| Ames | 3 | 65.5 | 1 | **84.6** | 1 | 67.2 |

Table 3: **Mean certification time** in seconds for FPA and Jia et al.'s [Jia+22b] randomized ablation (RA). FPA is 2 to 3 orders of magnitude faster than baseline RA.

| Dataset | RA (Jia'22b) | | FPA (ours) | | Speedup |
|---|---|---|---|---|---|
| | $e$ | Time | $T$ | Time | |
| CIFAR10 | 15 | 5.4E+0 | 115 | 7.3E−3 | **743×** |
| MNIST | 25 | 6.8E−1 | 60 | 2.9E−3 | **235×** |
| Weather | 45 | 3.1E−1 | 21 | 1.0E−4 | **3,134×** |
| Ames | 60 | 3.8E−1 | 21 | 3.5E−4 | **1,082×** |

**Takeaway #2** *FPA's median robustness gains come at little cost in classification accuracy.* Table 2 reports FPA's classification accuracy at two robustness levels: (1) FPA's best median robustness (blue) and (2) RA's best median robustness (**bold**). Table 2 also reports RA's classification accuracy at its own best median robustness (last column). For CIFAR10 at median robustness of 10 pixels, FPA's classification accuracy was 10.2 percentage points (pp) better than RA (75.0% vs. 64.7%). At $r_{med} = 13$, FPA's CIFAR10 classification accuracy was 62.4%, only 2.3pp lower than RA's classification accuracy at $\rho_{med} = 10$. For Weather at median robustness 1, FPA's classification accuracy was 10.1pp better than RA (85.3% vs. 75.2%); even at $r_{med} = 4$, FPA's classification accuracy was 76.1%, 0.9pp better than RA at $\rho_{med} = 1$. For MNIST at median robustness 10, FPA's classification accuracy was 3pp better than RA (96.1% vs. 93.1%). At $r_{med} = 12$, FPA's MNIST classification accuracy was 5.9pp lower than RA's classification accuracy at $\rho_{med} = 10$ (87.2% vs. 93.1%).

**Takeaway #3** *FPA certifies predictions 2 to 3 orders of magnitude faster than RA.* Table 3 compares the mean certification times using the hyperparameter settings with the best median robustness. To certify one prediction, Jia et al.'s [Jia+22b] improved RA evaluates 100k ablated inputs. In contrast, FPA requires exactly $T$ forward passes per prediction (one per submodel).

**Takeaway #4** *FPA provides strong patch robustness without any assumptions about patch shape or the number of patches.* As Table 4 details, FPA certifies 41.6% of CIFAR10 predictions at $r = 24$ perturbed pixels (2.3% of $d$) – regardless of patch shape or number of patches. In contrast, (de)randomized smoothing's [LF20a] 24-pixel certified accuracy varies between 0% to 72.7% based on patch shape alone. BAGCERT's certified accuracy drops as low as 43.1% for 24-pixel column and row patches – only 1.5pp better than FPA. Unlike FPA, patch defenses' certified accuracy guarantees decline further or even evaporate under (1) multiple patches, (2) training data perturbations, and (3) amorphous shapes. While less effective in some settings than dedicated patch defenses that make stronger assumptions and weaker guarantees, FPA is still competitive, providing patch guarantees essentially for free.

**Takeaway #5** *FPA is the first integrated defense to provide significant pointwise robustness guarantees over the union of evasion, backdoor, and poisoning attacks – $\ell_0$ or otherwise.* Consider CIFAR10 ($n = 50{,}000$) where FPA feature robustness $r \geq 25$ (Table 4) certifies 41.0% of predictions' robustness against 1.25M arbitrarily perturbed pixels. In contrast, the only other certified defense robust over the union of evasion, backdoor, and poisoning attacks [Web+20] certifies the equivalent of 3 or fewer arbitrarily perturbed CIFAR10 pixels (i.e., a total training and test $\ell_2$ perturbation distance of $\leq 3$). Moreover, FPA certifies $r \geq 7$ for 35.1% of Weather predictions ($n > 3M$ – supplemental Table 28) – a pointwise guaranteed robustness of up to 21M arbitrarily perturbed feature values.

Table 4: **CIFAR10 certified patch accuracy** (% – larger is better) for FPA, RA, and three dedicated patch defenses. FPA is competitive despite making fewer assumptions and providing stronger guarantees than patch defenses.

| Method | 24 Pixel Rect. | | Square |
|---|---|---|---|
| | Min. | Max. | $5 \times 5$ |
| FPA Plurality ($T = 180$, ours) | $\longleftarrow$ 38.53 $\longrightarrow$ | | 37.77 |
| FPA Run-Off ($T = 180$, ours) | $\longleftarrow$ 41.60 $\longrightarrow$ | | 40.95 |
| Randomized Ablation (LF'20b) | $\longleftarrow$ 28.95 $\longrightarrow$ | | 28.21 |
| Randomized Ablation (Jia'22b) | $\longleftarrow$ 37.31 $\longrightarrow$ | | 36.43 |
| (De)Random. Smoothing [LF20a] | 0.0 | 72.68 | 57.69 |
| BagCert [MY21] | **43.11** | 60.17 | 59.95 |
| Patch IBP [Chi+20b] | — | — | 30.30 |

### 6.3 Model Training Time

Recall that baseline randomized ablation (RA) trains only a single model, while FPA requires training $T$ submodels. However, FPA's training time is generally *not* $T$ times longer than RA's. As supplemental Section H.8 details, an FPA ensemble is sometimes *faster* to train than RA models since FPA does not require ablated training. For instance, for tabular regression datasets Weather and Ames, FPA supports GBDTs out of the box, while RA requires the use of parametric models. This translates to FPA being $18\times$ and $60\times$ faster to train than RA for Weather ($T = 31$) and Ames ($T = 51$), respectively.

Moreover, Section 6.1 explains that for datasets CIFAR10 and MNIST, FPA used small, simple CNN submodels that are fast to train. When $T = 25$, FPA is $2\times$ and $4\times$ slower to train than RA for CIFAR10 and MNIST, respectively. At $T = 115$ for CIFAR10 and $T = 60$ for MNIST, FPA is only about $10\times$ slower to train than RA. Note that since each FPA submodel is independent, FPA training is embarrassingly parallel. Hence, while FPA is slower to train than RA for CIFAR10 and MNIST on a single system, FPA at maximum parallelism takes significantly less wall time to train than RA.

## 7 Conclusions

This paper proposes *feature partition aggregation* – a certified defense against the union of $\ell_0$ evasion, poisoning, and backdoor attacks. FPA provided stronger and larger robustness guarantees than the state-of-the-art $\ell_0$ evasion defense, randomized ablation. FPA's certified feature guarantees are particularly important for *vertically partitioned* data where a single compromised data source allows an attacker to arbitrarily modify a limited number of features for all instances – training and test.

To our knowledge, FPA is the first integrated defense that provides non-trivial pointwise robustness guarantees against the union of evasion, poisoning, and backdoor attacks – $\ell_0$ or otherwise [Web+20]. Future work remains to develop other $\ell_p$ certified defenses over this union of attack types.

## Acknowledgments

The authors thank Jonathan Brophy for helpful discussions and feedback on earlier drafts of this manuscript. This work was supported by a grant from the Air Force Research Laboratory and the Defense Advanced Research Projects Agency (DARPA) — agreement number FA8750-16-C-0166, subcontract K001892-00-S05, as well as a second grant from DARPA, agreement number HR00112090135. This work benefited from access to the University of Oregon high-performance computer, Talapas.

# References

[Bar17]     Jonathan T. Barron. *Continuously Differentiable Exponential Linear Units*. 2017. arXiv: 1704.07483 [cs.LG].

[BNL12]     Battista Biggio, Blaine Nelson, and Pavel Laskov. "Poisoning Attacks against Support Vector Machines". In: *Proceedings of the 29th International Conference on Machine Learning*. ICML'12. Edinburgh, Great Britain: PMLR, 2012. URL: https://arxiv.org/abs/1206.6389.

[BHL23]     Jonathan Brophy, Zayd Hammoudeh, and Daniel Lowd. "Adapting and Evaluating Influence-Estimation Methods for Gradient-Boosted Decision Trees". In: *Journal of Machine Learning Research* 24 (2023), pp. 1–48. URL: http://jmlr.org/papers/v24/22-0449.html.

[Bro+17]    Tom B. Brown, Dandelion Mané, Aurko Roy, Martín Abadi, and Justin Gilmer. *Adversarial Patch*. 2017. arXiv: 1712.09665 [cs.CV].

[Cal+21]    Stefano Calzavara, Claudio Lucchese, Federico Marcuzzi, and Salvatore Orlando. "Feature Partitioning for Robust Tree Ensembles and their Certification in Adversarial Scenarios". In: *EURASIP Journal on Information Security* (Dec. 2021), pp. 245–317.

[Che+22]    Ruoxin Chen, Zenan Li, Jie Li, Chentao Wu, and Junchi Yan. "On Collective Robustness of Bagging Against Data Poisoning". In: *Proceedings of the 39th International Conference on Machine Learning*. ICML'22. PMLR, 2022. URL: https://arxiv.org/abs/2205.13176.

[Che+17]    Xinyun Chen, Chang Liu, Bo Li, Kimberly Lu, and Dawn Song. *Targeted Backdoor Attacks on Deep Learning Systems Using Data Poisoning*. 2017. arXiv: 1712.05526 [cs.CR].

[Chi+20a]   Ping-yeh Chiang, Michael J. Curry, Ahmed Abdelkader, Aounon Kumar, John Dickerson, and Tom Goldstein. "Detection as Regression: Certified Object Detection by Median Smoothing". In: *Proceedings of the 34th Conference on Neural Information Processing Systems*. NeurIPS'20. Virtual Only: Curran Associates, Inc., 2020. URL: https://arxiv.org/abs/2007.03730.

[Chi+20b]   Ping-yeh Chiang, Renkun Ni, Ahmed Abdelkader, Chen Zhu, Christoph Studor, and Tom Goldstein. "Certified Defenses for Adversarial Patches". In: *Proceedings of the 8th International Conference on Learning Representations*. ICLR'20. 2020. URL: https://openreview.net/forum?id=HyeaSkrYPH.

[CRK19]     Jeremy Cohen, Elan Rosenfeld, and Zico Kolter. "Certified Adversarial Robustness via Randomized Smoothing". In: *Proceedings of the 36th International Conference on Machine Learning*. ICML'19. PMLR, 2019. URL: https://proceedings.mlr.press/v97/cohen19c.html.

[Col+17]    Cody A. Coleman, Deepak Narayanan, Daniel Kang, Tian Zhao, Jian Zhang, Luigi Nardi, Peter Bailis, Kunle Olukotun, Chris Ré, and Matei Zaharia. "DAWNBench: An End-to-End Deep Learning Benchmark and Competition". In: *Proceedings of the 2017 NeurIPS Workshop on Machine Learning Systems*. Long Beach, California, USA: Curran Associates, Inc., 2017. URL: https://dawn.cs.stanford.edu/benchmark/.

[De 11]     Dean De Cock. "Ames, Iowa: Alternative to the Boston Housing Data as an End of Semester Regression Project". In: *Journal of Statistics Education* 19.3 (2011).

[Gu+19]     Tianyu Gu, Kang Liu, Brendan Dolan-Gavitt, and Siddharth Garg. "BadNets: Evaluating Backdooring Attacks on Deep Neural Networks". In: *IEEE Access* 7 (2019), pp. 47230–47244. URL: https://ieeexplore.ieee.org/document/8685687.

[HL22]      Zayd Hammoudeh and Daniel Lowd. "Identifying a Training-Set Attack's Target Using Renormalized Influence Estimation". In: *Proceedings of the 29th ACM SIGSAC Conference on Computer and Communications Security*. CCS'22. Los Angeles, CA: Association for Computing Machinery, 2022. URL: https://arxiv.org/abs/2201.10055.

[HL23a]     Zayd Hammoudeh and Daniel Lowd. "Feature Partition Aggregation: A Fast Certified Defense Against a Union of $\ell_0$ Attacks". In: *Proceedings of the 2nd ICML Workshop on New Frontiers in Adversarial Machine Learning*. AdvML-Frontiers'23. 2023. URL: https://arxiv.org/abs/2302.11628.

[HL23b]     Zayd Hammoudeh and Daniel Lowd. "Reducing Certified Regression to Certified Classification for General Poisoning Attacks". In: *Proceedings of the 1st IEEE Conference on Secure and Trustworthy Machine Learning*. SaTML'23. 2023. URL: https://arxiv.org/abs/2208.13904.

[HL24]      Zayd Hammoudeh and Daniel Lowd. "Provable Robustness Against a Union of $\ell_0$ Attacks". In: *Proceedings of the 38th AAAI Conference on Artificial Intelligence*. AAAI'24. 2024.

[Hua+20]   W. Ronny Huang, Jonas Geiping, Liam Fowl, Gavin Taylor, and Tom Goldstein. "MetaPoison: Practical General-purpose Clean-label Data Poisoning". In: *Proceedings of the 34th Conference on Neural Information Processing Systems*. NeurIPS'20. Virtual Only: Curran Associates, Inc., 2020. URL: https://arxiv.org/abs/2004.00225.

[JCG21]   Jinyuan Jia, Xiaoyu Cao, and Neil Zhenqiang Gong. "Intrinsic Certified Robustness of Bagging against Data Poisoning Attacks". In: *Proceedings of the 35th AAAI Conference on Artificial Intelligence*. AAAI'21. 2021. URL: https://arxiv.org/abs/2008.04495.

[Jia+22a]  Jinyuan Jia, Yupei Liu, Xiaoyu Cao, and Neil Zhenqiang Gong. "Certified Robustness of Nearest Neighbors against Data Poisoning and Backdoor Attacks". In: *Proceedings of the 36th AAAI Conference on Artificial Intelligence*. AAAI'22. 2022. URL: https://arxiv.org/abs/2012.03765.

[Jia+22b]  Jinyuan Jia, Binghui Wang, Xiaoyu Cao, Hongbin Liu, and Neil Zhenqiang Gong. "Almost Tight $\ell_0$-norm Certified Robustness of Top-k Predictions against Adversarial Perturbations". In: *Proceedings of the 10th International Conference on Learning Representations*. ICLR'22. 2022. URL: https://openreview.net/forum?id=gJLEXy3ySpu.

[Ke+17]    Guolin Ke, Qi Meng, Thomas Finley, Taifeng Wang, Wei Chen, Weidong Ma, Qiwei Ye, and Tie-Yan Liu. "LightGBM: A Highly Efficient Gradient Boosting Decision Tree". In: *Proceedings of the 31st International Conference on Neural Information Processing Systems*. NeurIPS'17. 2017.

[KT06]     Jon Kleinberg and Éva Tardos. *Algorithm Design*. Addison Wesley, 2006.

[KNH14]    Alex Krizhevsky, Vinod Nair, and Geoffrey Hinton. *The CIFAR-10 Dataset*. 2014.

[LeC+98]   Yann LeCun, Léon Bottou, Yoshua Bengio, and Patrick Haffner. "Gradient-Based Learning Applied to Document Recognition". In: *Proceedings of the IEEE*. Vol. 86. 1998, pp. 2278–2324.

[Lee+19]   Guang-He Lee, Yang Yuan, Shiyu Chang, and Tommi Jaakkola. "Tight Certificates of Adversarial Robustness for Randomly Smoothed Classifiers". In: *Proceedings of the 33rd Conference on Neural Information Processing Systems*. NeurIPS'19. 2019. URL: https://arxiv.org/abs/1906.04948.

[LF20a]    Alexander Levine and Soheil Feizi. "(De)Randomized Smoothing for Certifiable Defense against Patch Attacks". In: *Proceedings of the 34th International Conference on Neural Information Processing Systems*. NeurIPS'20. Red Hook, NY, USA: Curran Associates Inc., 2020. URL: https://arxiv.org/abs/2002.10733.

[LF20b]    Alexander Levine and Soheil Feizi. "Robustness Certificates for Sparse Adversarial Attacks by Randomized Ablation". In: *Proceedings of the 34th AAAI Conference on Artificial Intelligence*. AAAI Press, 2020. URL: https://arxiv.org/abs/1911.09272.

[LF21]     Alexander Levine and Soheil Feizi. "Deep Partition Aggregation: Provable Defenses against General Poisoning Attacks". In: *Proceedings of the 9th International Conference on Learning Representations*. ICLR'21. Virtual Only, 2021. URL: https://arxiv.org/abs/2006.14768.

[LF22]     Alexander J. Levine and Soheil Feizi. "Provable Adversarial Robustness for Fractional Lp Threat Models". In: *Proceedings of the 25th International Conference on Artificial Intelligence and Statistics*. AISTATS'22. 2022. URL: https://arxiv.org/abs/2203.08945.

[LXL23]    Linyi Li, Tao Xie, and Bo Li. "SoK: Certified Robustness for Deep Neural Networks". In: *Proceedings of the 44th IEEE Symposium on Security and Privacy*. SP'23. IEEE, 2023. URL: https://arxiv.org/abs/2009.04131.

[LDD21]    Xiling Li, Rafael Dowsley, and Martine De Cock. "Privacy-Preserving Feature Selection with Secure Multiparty Computation". In: *Proceedings of the 38th International Conference on Machine Learning*. ICML'21. 2021. URL: https://arxiv.org/abs/2102.03517.

[Li+22]    Yiming Li, Baoyuan Wu, Yong Jiang, Zhifeng Li, and Shu-Tao Xia. "Backdoor Learning: A Survey". In: *IEEE Transactions on Neural Networks and Learning Systems* (2022). DOI: 10.1109/TNNLS.2022.3182979. URL: https://arxiv.org/abs/2007.08745.

[LCY14]    Min Lin, Qiang Chen, and Shuicheng Yan. "Network in Network". In: *Proceedings of the 2nd International Conference on Learning Representations*. ICLR'14. 2014. URL: https://arxiv.org/abs/1312.4400.

[Mal+21]  Andrey Malinin, Neil Band, Yarin Gal, Mark Gales, Alexander Ganshin, German Chesnokov, Alexey Noskov, Andrey Ploskonosov, Liudmila Prokhorenkova, Ivan Provilkov, Vatsal Raina, Vyas Raina, Denis Roginskiy, Mariya Shmatova, Panagiotis Tigas, and Boris Yangel. "Shifts: A Dataset of Real Distributional Shift Across Multiple Large-Scale Tasks". In: *Proceedings of the 35th Conference on Neural Information Processing Systems*. NeurIPS'21. Curran Associates, Inc., 2021. URL: https://arxiv.org/abs/2107.07455.

[MY21]  Jan Hendrik Metzen and Maksym Yatsura. "Efficient Certified Defenses Against Patch Attacks on Image Classifiers". In: *Proceedings of the 9th International Conference on Learning Representations*. ICLR'21. 2021. URL: https://openreview.net/forum?id=hr-3PMvDpil.

[Pag20]  David Page. "How to Train Your ResNet". In: (May 2020). URL: https://myrtle.ai/learn/how-to-train-your-resnet/.

[Pas+19]  Adam Paszke, Sam Gross, Francisco Massa, Adam Lerer, James Bradbury, Gregory Chanan, Trevor Killeen, Zeming Lin, Natalia Gimelshein, Luca Antiga, Alban Desmaison, Andreas Kopf, Edward Yang, Zachary DeVito, Martin Raison, Alykhan Tejani, Sasank Chilamkurthy, Benoit Steiner, Lu Fang, Junjie Bai, and Soumith Chintala. "PyTorch: An Imperative Style, High-Performance Deep Learning Library". In: *Proceedings of the 33rd Conference on Neural Information Processing Systems*. NeurIPS'19. 2019. URL: https://arxiv.org/abs/1912.01703.

[Rez+23]  Keivan Rezaei, Kiarash Banihashem, Atoosa Chegini, and Soheil Feizi. "Run-Off Election: Improved Provable Defense against Data Poisoning Attacks". In: *Proceedings of the 40th International Conference on Machine Learning*. ICML'23. 2023. URL: https://arxiv.org/abs/2302.02300.

[Ros+20]  Elan Rosenfeld, Ezra Winston, Pradeep Ravikumar, and J. Zico Kolter. "Certified Robustness to Label-Flipping Attacks via Randomized Smoothing". In: *Proceedings of the 37th International Conference on Machine Learning*. ICML'20. 2020. URL: https://arxiv.org/abs/2002.03018.

[Sha+18]  Ali Shafahi, W. Ronny Huang, Mahyar Najibi, Octavian Suciu, Christoph Studer, Tudor Dumitras, and Tom Goldstein. "Poison Frogs! Targeted Clean-Label Poisoning Attacks on Neural Networks". In: *Proceedings of the 32nd Conference on Neural Information Processing Systems*. NeurIPS'18. 2018. URL: https://arxiv.org/abs/1804.00792.

[SD20]  Cecilia Summers and Michael J. Dinneen. "Four Things Everyone Should Know to Improve Batch Normalization". In: *Proceedings of the 8th International Conference on Learning Representations*. ICLR'20. Virtual Only, 2020. URL: https://arxiv.org/abs/1906.03548.

[Sze+14]  Christian Szegedy, Wojciech Zaremba, Ilya Sutskever, Joan Bruna, Dumitru Erhan, Ian Goodfellow, and Rob Fergus. "Intriguing Properties of Neural Networks". In: *Proceedings of the 2nd International Conference on Learning Representations*. ICLR'14. 2014. URL: https://arxiv.org/abs/1312.6199.

[Wal+21]  Eric Wallace, Tony Z. Zhao, Shi Feng, and Sameer Singh. "Concealed Data Poisoning Attacks on NLP Models". In: *Proceedings of the North American Chapter of the Association for Computational Linguistics*. NAACL'21. 2021. URL: https://arxiv.org/abs/2010.12563.

[WF23]  Wenxiao Wang and Soheil Feizi. *Temporal Robustness Against Data Poisoning*. 2023. arXiv: 2302.03684 [cs.LG].

[WLF22a]  Wenxiao Wang, Alexander Levine, and Soheil Feizi. "Improved Certified Defenses against Data Poisoning with (Deterministic) Finite Aggregation". In: *Proceedings of the 39th International Conference on Machine Learning*. ICML'22. 2022. URL: https://arxiv.org/abs/2202.02628.

[WLF22b]  Wenxiao Wang, Alexander Levine, and Soheil Feizi. "Lethal Dose Conjecture on Data Poisoning". In: *Proceedings of the 36th Conference on Neural Information Processing Systems*. NeurIPS'22. Curran Associates, Inc., 2022. URL: https://arxiv.org/abs/2208.03309.

[Web+20]  Maurice Weber, Xiaojun Xu, Bojan Karlaš, Ce Zhang, and Bo Li. *RAB: Provable Robustness Against Backdoor Attacks*. 2020. arXiv: 2003.08904 [cs.LG].

[Wei+22]  Kang Wei, Jun Li, Chuan Ma, Ming Ding, Sha Wei, Fan Wu, Guihai Chen, and Thilina Ranbaduge. *Vertical Federated Learning: Challenges, Methodologies and Experiments*. 2022. arXiv: 2202.04309 [cs.LG].

[YHL23]  Wencong You, Zayd Hammoudeh, and Daniel Lowd. "Large Language Models Are Better Adversaries: Exploring Generative Clean-Label Backdoor Attacks Against Text Classifiers". In: *Findings of the Association for Computational Linguistics*. EMNLP'23. 2023.

# Provable Robustness Against a Union of $\ell_0$ Adversarial Attacks

Supplemental Materials

## Organization of the Appendix

# A Nomenclature Reference

Scalars and functions are denoted with lowercase italics letters. Vectors are denoted as lowercase bold letters. Matrices are denoted as uppercase bold letters. The $j$-th column of a matrix $\mathbf{A}$ is denoted $\mathbf{A}_j$.

Table 5: **Nomenclature Reference**: Related symbols are grouped together. For example, the first group lists the acronyms of methods evaluated in this work. This table also includes nomenclature symbols that only appear in the supplement.

| | |
|---|---|
| FPA | Our certified defense, feature partition aggregation, against sparse poisoning, backdoor, evasion, and patch attacks |
| RA | Randomized ablation. Certified $\ell_0$-norm evasion defense. Proposed by Levine and Feizi [LF20b] and subsequently improved by Jia et al. [Jia+22b] |
| DPA | Deep partition aggregation certified instance-wise poisoning defense proposed by Levine and Feizi [LF21] |
| DRS | (De)randomized smoothing certified patch defense proposed by Levine and Feizi [LF20a]. Based on randomized ablation |
| Patch IBP | Certified patch defense based on interval bound propagation proposed by Chiang et al. [Chi+20b] |
| BAGCERT | Certified patch defense proposed by Metzen and Yatsura [MY21] |
| RAB | Robustness against backdoors certified defense proposed by Weber et al. [Web+20] |
| LightGBM | Gradient-boosted decision tree model architecture [Ke+17] |
| $r$ | Pointwise certified feature robustness – feature partition aggregation's certification objective (Def. 1) |
| $r_{\text{med}}$ | Median certified feature robustness w.r.t. a dataset's test set |
| $\rho$ | Pointwise $\ell_0$-norm certified evasion-only robustness (Def. 2). A weaker guarantee than certified feature robustness. |
| $\rho_{\text{med}}$ | Median $\ell_0$-norm certified evasion-only robustness w.r.t. a dataset's test set |
| $\bar{\rho}$ | Certified instance-wise poisoning robustness. (Def. 9 – Sec. C). |
| $\widetilde{r}$ | Pointwise certified feature and label-flipping robustness (Def. 11 – Sec. E) |
| $[m]$ | Integer set $\{1, \ldots, m\}$ where $m \in \mathbb{N}$ |
| $\mathbb{1}[q]$ | Indicator function where $\mathbb{1}[q] = 1$ if $q$ is true and 0 otherwise |
| $\|\mathbf{w}\|_0$ | $\ell_0$ norm for vector $\mathbf{w}$, i.e., the number of non-zero elements in $\mathbf{w}$ |
| $\mathbf{X}_j$ | $j$-th column of matrix $\mathbf{X}$ where $j \in [d]$ and $\mathbf{X}_j \in \mathbb{R}^n$ |
| $\mathbf{X} \ominus \mathbf{X}'$ | Set of column indices over which equal-size matrices $\mathbf{X}$ and $\mathbf{X}'$ differ, where $\mathbf{X} \ominus \mathbf{X}' = \{j \in [d] : \mathbf{X}_j \neq \mathbf{X}'_j\}$ |
| $x_j$ | $j$-th dimension of vector $\mathbf{x}$ where $j \in [d]$ and $x_j \in \mathbb{R}$ |
| $\mathbf{x} \ominus \mathbf{x}'$ | Set of dimensions over which vectors $\mathbf{x}$ and $\mathbf{x}'$ differ where $\mathbf{x} \ominus \mathbf{x}' = \{j \in [d] : x_j \neq x'_j\}$ |
| $d_{\text{sym}}(\mathcal{D}, \mathcal{D}')$ | Symmetric difference between sets $\mathcal{D}$ and $\mathcal{D}'$ |
| pp | Percentage points |
| $n$ | Number of training instances |
| $\mathcal{X}$ | Feature domain where $\mathcal{X} \subseteq \mathbb{R}^d$ |
| $\mathbf{x}$ | Feature vector where $\forall_{\mathbf{x}} \mathbf{x} \in \mathcal{X}$ |
| $d$ | Feature dimension where $\forall_{\mathbf{x}} |\mathbf{x}| = d$ |
| $[d]$ | Complete feature set |
| $\mathcal{Y}$ | Label set where $\mathcal{Y} \subseteq \mathbb{N}$ |
| $y$ | Instance label where $\forall_y y \in \mathcal{Y}$ |
| $(\mathbf{x}_i, y_i)$ | Arbitrary training instance where $\mathbf{x}_i \in \mathcal{X}$, $y_i \in \mathcal{Y}$, and $i \in [n]$ |
| $\mathbf{X}$ | Training feature matrix where $\mathbf{X} := [\, \mathbf{x}_1 \; \cdots \; \mathbf{x}_n \,]^\intercal$ and $\mathbf{X} \in \mathbb{R}^{n \times d}$ |
| $\mathbf{y}$ | Training label vector where $\mathbf{y} := [y_1, \ldots, y_n]$ |

(Continued ...)

Table 5: **Nomenclature Reference (Continued)**: Related symbols are grouped together.

| Symbol | Description |
|---|---|
| $f$ | Voting-based, ensemble classifier trained over partitioned feature sets where $f : \mathcal{X} \to \mathcal{Y}$ |
| $T$ | Number of submodels in ensemble $f$ |
| $\mathcal{S}_t$ | Feature subset considered by the $t$-th submodel during training and test where $\mathcal{S}_t \subset [d]$ and $\bigsqcup_{t=1}^{T} \mathcal{S}_t = [d]$ |
| $\mathbf{x}_{\mathcal{S}_t}$ | Subvector of $\mathbf{x} \in \mathcal{X}$ restricted to feature subset $\mathcal{S}_t \subset [d]$ |
| $D_t$ | Training set for the $t$-th submodel |
| $\phi$ | Spread degree of the (overlapping) feature subsets $D_1, D_2, \ldots$; by default, $\phi = 1$ (Sec. F). |
| $f(\mathbf{x})$ | Model prediction for instance $\mathbf{x} \in \mathcal{X}$ and $f(\mathbf{x}) \in \mathcal{Y}$ |
| $f_t(\mathbf{x})$ | Label predicted by the $t$-th submodel for instance $\mathbf{x} \in \mathcal{X}$ where $f_t(\mathbf{x}) \coloneqq \arg\max_{y \in \mathcal{Y}} g_t(\mathbf{x}, y)$ |
| $\dot{c}_y(\mathbf{x})$ | Submodel vote count for label $y$ and feature vector $\mathbf{x}$ where $\dot{c}_y(\mathbf{x}) \coloneqq \sum_{t=1}^{T} \mathbb{1}[f_t(\mathbf{x}) = y]$ |
| $\mathrm{GAP}_{\mathrm{vote}}(y, y'; \mathbf{x})$ | Submodel vote gap for instance $\mathbf{x} \in \mathcal{X}$ and labels $y, y' \in \mathcal{Y}$ where $\mathrm{GAP}_{\mathrm{vote}}(y, y'; \mathbf{x}) \coloneqq \dot{c}_y(\mathbf{x}) - \dot{c}_{y'}(\mathbf{x}) - \mathbb{1}[y' < y]$ |
| $y_{\mathrm{pl}}$ | Submodel plurality label where $y_{\mathrm{pl}} \coloneqq \arg\max_{y \in \mathcal{Y}} \dot{c}_y(\mathbf{x})$ and ties broken by preferring the smaller label. FPA ensemble prediction under the plurality label decision function (Sec. 4.1) |
| $y_{\mathrm{ru}}$ | Label with the second-most submodel votes (i.e., the "runner up") where $y_{\mathrm{ru}} \coloneqq \arg\max_{y' \in \mathcal{Y} \setminus y_{\mathrm{pl}}} \dot{c}_{y'}(\mathbf{x})$ |
| $g_t(\mathbf{x}, y)$ | Logit value predicted by the $t$-th submodel for instance $\mathbf{x} \in \mathcal{X}$ and label $y \in \mathcal{Y}$ where $g_t(\mathbf{x}, y) \in [0, 1]$ |
| $y_{\mathrm{RO}}$ | FPA ensemble prediction under the run-off decision function (Sec. 4.2). |
| $\widetilde{y}_{\mathrm{RO}}$ | Label in the run-off decision function's second round that is not selected as the run-off prediction where $\widetilde{y}_{\mathrm{RO}} \coloneqq \{y_{\mathrm{pl}}, y_{\mathrm{ru}}\} \setminus y_{\mathrm{RO}}$ |
| $\ddot{c}_{\mathbf{x}}(y; y')$ | Pairwise logit count for instance $\mathbf{x}$ and label $y \in \mathcal{Y}$ w.r.t. label $y' \in \mathcal{Y}$ where $\ddot{c}_y(\mathbf{x}; y') \coloneqq \sum_{t=1}^{T} \mathbb{1}[g_t(\mathbf{x}, y) > g_t(\mathbf{x}, y')]$ |
| $\mathrm{GAP}_{\mathrm{logit}}(y, y'; \mathbf{x})$ | Submodel logit vote gap for labels $y, y' \in \mathcal{Y}$ where $\mathrm{GAP}_{\mathrm{logit}}(y, y'; \mathbf{x}) \coloneqq \ddot{c}_y(\mathbf{x}; y') - \ddot{c}_{y'}(\mathbf{x}; y) - \mathbb{1}[y' < y]$ |
| $f(\mathbf{x}; k)$ | Top-$k$ model prediction for instance $\mathbf{x} \in \mathcal{X}$ (Sec. D) |
| $\widetilde{y}$ | Label with the $(k+1)$-th most submodel votes (Sec. D) |
| $h_{\mathrm{tr}}$ | Instance space mapping function where $h_{\mathrm{tr}} : \mathcal{X} \times \mathcal{Y} \to [T]$ (Sec. E). |
| $h_{\mathcal{S}}$ | Feature subset mapping function for overlapping feature sets where $h_{\mathcal{S}} : [\phi T] \to [\phi T]$ (Sec. F) |
| $e$ | Randomized ablation hyperparameter – number of kept features with the other $(d - e)$ ablated where $e \in \mathbb{N}$. |
| BS | Blocking smoothing ablation paradigm used by (de)randomized smoothing [LF20a] |

# B  Proofs

This section contains all proofs for our theoretical contributions. Sec. B.1 provides the proofs for the main paper's theoretical contributions. Due to space, some of our theoretical contributions appear only in the supplement. Sec. B.2 contains the proofs for these supplement-only theoretical contributions.

## B.1  Theorems from the Main Paper

This section provides the proofs for our theoretical contributions in the main paper.

**Proof of Theorem 3**

*Proof.* Let
$$\Delta := \dot{c}_{y_{\text{pl}}}(\mathbf{x}) - \dot{c}_{y_{\text{ru}}}(\mathbf{x}) \leq \forall_{y' \notin \mathcal{Y} \setminus \{y_{\text{pl}}, y_{\text{ru}}\}} \dot{c}_{y_{\text{pl}}}(\mathbf{x}) - \dot{c}_{y'}(\mathbf{x}). \tag{9}$$
In words, vote-count difference $\Delta$ between plurality label $y_{\text{pl}}$ and runner-up label $y_{\text{ru}}$ is at least as small as the gap between $y_{\text{pl}}$ and any other label.

In the worst case, a single feature perturbation changes a single submodel's vote from plurality label $y_{\text{pl}}$ to a label of the adversary's choosing. Each perturbed submodel prediction reduces the gap between the plurality label and the adversary's chosen label by two. By Eq. (9), it takes the fewest number of vote changes for $y_{\text{ru}}$ to overtake plurality label $y_{\text{pl}}$ with the proof following by induction. $\Delta$ then lower bounds the certified robustness. When determining $r$, $\Delta$ may be even or odd. We separately consider both cases below.

**Case #1**: $\Delta$ is odd.

Since $\Delta$ is odd, there can never be a tie between labels $y_{\text{pl}}$ and $y_{\text{ru}}$, simplifying the analysis. Then, the maximum number of submodel predictions that can change without changing the plurality label is any $r \in \mathbb{N}$ satisfying

$$\dot{c}_{y_{\text{ru}}}(\mathbf{x}) + 2r < \dot{c}_{y_{\text{pl}}}(\mathbf{x}) \tag{10}$$

$$r < \frac{\dot{c}_{y_{\text{pl}}}(\mathbf{x}) - \dot{c}_{y_{\text{ru}}}(\mathbf{x})}{2} \tag{11}$$

$$r = \left\lfloor \frac{\dot{c}_{y_{\text{pl}}}(\mathbf{x}) - \dot{c}_{y_{\text{ru}}}(\mathbf{x})}{2} \right\rfloor \qquad \triangleright r \text{ must be a whole number} \tag{12}$$

$$= \left\lfloor \frac{\dot{c}_{y_{\text{pl}}}(\mathbf{x}) - \dot{c}_{y_{\text{ru}}}(\mathbf{x}) - \mathbb{1}[y_{\text{ru}} < y_{\text{pl}}]}{2} \right\rfloor \qquad \triangleright \text{Subtracting 1 has no effect when } \Delta \text{ odd} \tag{13}$$

$$= \left\lfloor \frac{\text{GAP}_{\text{vote}}(y_{\text{pl}}, y_{\text{ru}}; \mathbf{x})}{2} \right\rfloor \qquad \triangleright \text{Eq. (1).} \tag{14}$$

**Case #2**: $\Delta$ is even.

For even-valued $\Delta$, ties can occur. If $y_{\text{ru}} < y_{\text{pl}}$, the tie between $y_{\text{pl}}$ and $y_{\text{ru}}$ is broken in favor of $y_{\text{ru}}$. Then, the number of submodel predictions that can change without changing the plurality label is any $r \in \mathbb{N}$ satisfying

$$\dot{c}_{y_{\text{ru}}}(\mathbf{x}) + \mathbb{1}[y_{\text{ru}} < y_{\text{pl}}] + 2r < \dot{c}_{y_{\text{pl}}}(\mathbf{x}) \tag{15}$$

$$r \leq \frac{\dot{c}_{y_{\text{pl}}}(\mathbf{x}) - \dot{c}_{y_{\text{ru}}}(\mathbf{x}) - \mathbb{1}[y_{\text{ru}} < y_{\text{pl}}]}{2} \tag{16}$$

$$r = \left\lfloor \frac{\dot{c}_{y_{\text{pl}}}(\mathbf{x}) - \dot{c}_{y_{\text{ru}}}(\mathbf{x}) - \mathbb{1}[y_{\text{ru}} < y_{\text{pl}}]}{2} \right\rfloor \qquad \triangleright r \text{ must be a whole number} \tag{17}$$

$$= \left\lfloor \frac{\text{GAP}_{\text{vote}}(y_{\text{pl}}, y_{\text{ru}}; \mathbf{x})}{2} \right\rfloor \qquad \triangleright \text{Eq. (1).} \tag{18}$$

□

Theorem 3's definition of $r$ follows the same basic structure as that of *deep partition aggregation* [LF21, Eq. (10)].

**Proof of Claims Related to Theorem 4**

**Lemma 5.** *Let $f_1, \ldots, f_T$ be a set of $T$ models where $\forall_{t \in [T]} f_t : \mathcal{X} \to \mathcal{Y}$. Under submodel voting, label $y \in \mathcal{Y}$ is preferred over label $y' \in \mathcal{Y} \setminus y$ w.r.t. instance $\mathbf{x} \in \mathcal{X}$ if and only if $\mathrm{GAP}_{\mathrm{vote}}(y, y'; \mathbf{x}) \geq 0$.*

*Proof.* Label $y$ is preferred over label $y'$ in only two cases:

1. $y$ receives more (sub)model votes than $y'$, i.e., $\dot{c}_y(\mathbf{x}) > \dot{c}_{y'}(\mathbf{x})$.
2. $y$ and $y'$ receive the same number of votes and $y < y'$.

In the first case,

$$\begin{aligned} \mathrm{GAP}_{\mathrm{vote}}(y, y'; \mathbf{x}) &:= \dot{c}_y(\mathbf{x}) - \dot{c}_{y'}(\mathbf{x}) - \mathbb{1}[y' < y] \\ &\geq 1 - \mathbb{1}[y' < y] \\ &\geq 1 - 1 = 0. \end{aligned}$$

In the second case,

$$\begin{aligned} \mathrm{GAP}_{\mathrm{vote}}(y, y'; \mathbf{x}) &:= \dot{c}_y(\mathbf{x}) - \dot{c}_{y'}(\mathbf{x}) - \mathbb{1}[y' < y] \\ &= 0 - \mathbb{1}[y' < y] \\ &= 0 - 0 = 0. \end{aligned}$$

The reverse direction where $\mathrm{GAP}_{\mathrm{vote}}(y, y'; \mathbf{x}) \geq 0 \implies y$ is preferred over $y'$ can be proven by contradiction using similar logic as above. If $y'$ receives more votes than $y$, then $\mathrm{GAP}_{\mathrm{vote}}(y, y'; \mathbf{x}) < 0$, a contradiction. Similarly, if $\dot{c}_y(\mathbf{x}) = \dot{c}_{y'}(\mathbf{x})$ then necessarily $y' < y$. This also leads to a contradiction as $\mathrm{GAP}_{\mathrm{vote}}(y, y'; \mathbf{x})$ would be negative. □

**Lemma 6. Runoff Elections Case #1 Certified Feature Robustness** *Given submodel feature partition $\mathcal{S}_1, \ldots, \mathcal{S}_T$, let $f$ be a voting-based ensemble of $T$ submodels, where the $t$-th submodel uses only the features in $\mathcal{S}_t$. For instance $\mathbf{x} \in \mathcal{X}$, let $y_{\mathrm{RO}}$ be the label selected by the run-off decision function. The certified feature robustness of $y_{\mathrm{RO}}$ getting overtaken in round #2 of the run-off election is*

$$r_{\mathrm{RO}}^{\mathrm{Case1}} := \min_{y \in \mathcal{Y} \setminus y_{\mathrm{RO}}} \max \left\{ \left\lfloor \frac{\mathrm{GAP}_{\mathrm{vote}}(\widetilde{y}_{\mathrm{RO}}, y)}{2} \right\rfloor, \left\lfloor \frac{\mathrm{GAP}_{\mathrm{logit}}(y_{\mathrm{RO}}, y)}{2} \right\rfloor \right\}$$

*Proof.* For a label $y \in \mathcal{Y} \setminus y_{\mathrm{RO}}$ to overtake $y_{\mathrm{RO}}$, two requirements must be simultaneously met:

- $y$ and $y_{\mathrm{RO}}$ must be round #1's top-two labels, and
- $y$ must be preferred over $y_{\mathrm{RO}}$ in round #2.

Let $\widetilde{y}_{\mathrm{RO}} \in \mathcal{Y} \setminus y_{\mathrm{pl}}$ denote the other top-two label in round #1. Note that $\widetilde{y}_{\mathrm{RO}}$ may or may not be the same as $y$. The robustness of $\widetilde{y}_{\mathrm{RO}}$ to being overtaken by $y$ in round #1 follows directly from Theorem 3 and equals

$$r' = \left\lfloor \frac{\mathrm{GAP}_{\mathrm{vote}}(\widetilde{y}_{\mathrm{RO}}, y; \mathbf{x})}{2} \right\rfloor. \tag{19}$$

Concerning the second requirement, $y_{\mathrm{RO}}$ is preferred over $y$ in round #2 so long as $\mathrm{GAP}_{\mathrm{logit}}(y_{\mathrm{RO}}, y; \mathbf{x}) \geq 0$. Following similar logic as above, $y_{\mathrm{RO}}$'s certified feature robustness in round #2 is

$$r'' = \left\lfloor \frac{\mathrm{GAP}_{\mathrm{logit}}(y_{\mathrm{RO}}, y; \mathbf{x})}{2} \right\rfloor. \tag{20}$$

Since both requirements must hold, the certified feature robustness is lower bounded by both (i.e., the maximum) of Eqs. (19) and (20). Moreover, the optimal label $y \in \mathcal{Y} \setminus y_{\mathrm{RO}}$ is not determined a priori meaning all labels need to be checked. □

**Lemma 7. Runoff Elections Case #2 Certified Feature Robustness** *Given submodel feature partition $\mathcal{S}_1, \ldots, \mathcal{S}_T$, let $f$ be a voting-based ensemble of $T$ submodels, where the $t$-th submodel uses only the features in $\mathcal{S}_t$. For instance $\mathbf{x} \in \mathcal{X}$, let $y_{\text{RO}}$ be the label selected by the run-off decision function. Define recursive function* $\mathtt{dp}$ *as*

$$\mathtt{dp}[\Delta, \Delta'] = \begin{cases} 0 & \min\{\Delta, \Delta'\} \leq 1 \text{ and } (\Delta, \Delta') \neq (1,1) \\ 1 + \min\{\mathtt{dp}[\Delta-2, \Delta'-1], \mathtt{dp}[\Delta-1, \Delta'-2]\} & \text{Otherwise} \end{cases} \quad (21)$$

*Then $y_{\text{RO}}$'s certified feature robustness of remaining in the top-two round #1 labels predicted by the submodels is*

$$r_{\text{RO}}^{\text{Case2}} := \min_{y, y' \in \mathcal{Y} \setminus y_{\text{RO}}} \mathtt{dp}\big[\text{GAP}_{\text{vote}}(y_{\text{RO}}, y; \mathbf{x}), \text{GAP}_{\text{vote}}(y_{\text{RO}}, y'; \mathbf{x})\big]$$

*Proof.* Lemma 5 proves that a label $y$ is preferred over another label $y'$ iff $\text{GAP}_{\text{vote}}(y, y'; \mathbf{x}) \geq 0$. For label $y_{\text{RO}}$ to be in round #1's top two, no pair of labels can both have negative submodel vote gaps w.r.t. $y_{\text{RO}}$. Determining $y_{\text{RO}}$'s round #1 certified feature robustness reduces to determining the maximum number of submodel votes that can be perturbed with it remaining guaranteed that no pair of labels both have negative submodel vote gaps.

In the best case for an attacker, perturbing a single feature changes a submodel's predicted label from $y_{\text{RO}}$ to a label of the attacker's choosing, e.g., $y \neq y_{\text{RO}}$; this perturbation decreases $\text{GAP}_{\text{vote}}(y_{\text{RO}}, y; \mathbf{x})$ by 2. For all other $y' \in \mathcal{Y} \setminus \{y_{\text{RO}}, y\}$, this perturbation also decreases $\text{GAP}_{\text{vote}}(y_{\text{RO}}, y'; \mathbf{x})$ by 1.

By definition, $y_{\text{RO}}$ is in the top-two round #1 labels, meaning $r_{\text{RO}}^{\text{Case2}} \geq 0$. Consider first when $\max\{\text{GAP}_{\text{vote}}(y_{\text{RO}}, y), \text{GAP}_{\text{vote}}(y_{\text{RO}}, y')\} \leq 1$ and $(\Delta, \Delta') \neq (1,1)$. The attacker perturbs whichever label $y, y'$ has the larger submodel vote gap. Since at most one of these two labels has a positive gap, an additional submodel perturbation could make *both* $\text{GAP}_{\text{vote}}(y_{\text{RO}}, y; \mathbf{x})$ and $\text{GAP}_{\text{vote}}(y_{\text{RO}}, y'; \mathbf{x})$ negative meaning no further feature perturbations are possible. In the special case of $\Delta = \Delta' = 1$, perturbing a submodel predicting either label $y$ or $y'$ never causes the other label's submodel vote gap to be negative meaning one additional submodel feature perturbation is possible. When $\max\{\text{GAP}_{\text{vote}}(y_{\text{RO}}, y; \mathbf{x}), \text{GAP}_{\text{vote}}(y_{\text{RO}}, y'; \mathbf{x})\} > 1$, the proof follows by induction where recursive function $\mathtt{dp}$ returns the fewest number of submodel perturbations required given $y, y' \in \mathcal{Y}$.

Since the attacker's optimal pair of labels $y, y'$ is not determined a priori, Eq. (7)'s feature guarantee considers all pairs of labels and returns the robustness of the pair of labels most advantageous to the attacker. □

**Proof of Theorem 4**

*Proof.* For a given $\mathbf{x} \in \mathcal{X}$, there are only two possible ways that run-off prediction $y_{\text{RO}} \in \mathcal{Y}$ can be perturbed, namely:

1. $y_{\text{RO}}$ loses in run-off's second round.
2. $y_{\text{RO}}$ fails to qualify for the second round by not being in the top two labels in round #1.

These two cases align directly with Lemmas 6 and 7, respectively. An optimal attacker targets whichever of the two cases requires fewer feature perturbations. Therefore, run-off's certified feature robustness is the minimum of Eqs. (5) and (7). □

## B.2 Lemmas from the Supplemental Materials

This section provides the proofs for our theoretical contributions that appear only in the supplement.

**Proof of Theorem 10**

Alg. 1's iterative greedy strategy is formalized below.

**Def. 8. Certified Feature Robustness Greedy Strategy** *Given target label $y \in \mathcal{Y}$, plurality label $y_{\text{pl}} \in \mathcal{Y}$, and label $\widetilde{y} \in \mathcal{Y}$ with the $(k+1)$-th most votes, if $\dot{c}_y(\mathbf{x}) > 0$, decrement $\dot{c}_y(\mathbf{x})$ by 1; otherwise, decrement $\dot{c}_{y_{\text{pl}}}(\mathbf{x})$ by 1. Increment both $\dot{c}_{\widetilde{y}}(\mathbf{x})$ and certified feature robustness $r$ by 1.*

Theorem 10's proof references Def. 8 for brevity.

*Proof.* We follow the classic "*greedy stays ahead*" proof strategy [KT06]. In short, given some iterative greedy strategy, the greedy algorithm always does better at each iteration than any other algorithm. Also, observe that the order that the greedy strategy perturbs the labels does not affect the optimality of the bound since each perturbation is strictly increasing, additive, and fully commutative.

In short, Def. 8's greedy strategy minimizes at each iteration the margin between $y$'s vote count, $\dot{c}_y(\mathbf{x})$, and the vote count of the label with the $(k+1)$-th most votes, i.e., $\dot{c}_{\widetilde{y}}(\mathbf{x})$. Recall that Theorem 3's proof above for top-1 certified robustness only considers the runner-up label $y_{\text{ru}}$ since all other labels $y' \notin \mathcal{Y} \setminus \{y, y_{\text{ru}}\}$ require at least as many label changes as runner-up $y_{\text{ru}}$ to overtake plurality label $y$. Def. 8's greedy strategy generalizes this idea where now only the top $(k+1)$ labels are considered and the rest of the labels ignored.

Each iteration of Alg. 1 may have a different label with the $(k+1)$-th most votes. For a given iteration, denote this label $\widetilde{y}$, making label $y$'s margin of remaining in the top $k$

$$\Delta := \dot{c}_y(\mathbf{x}) - \dot{c}_{\widetilde{y}}(\mathbf{x}). \tag{22}$$

Trivially, maximally reducing $\dot{c}_y(\mathbf{x})$ and maximally increasing $\dot{c}_{\widetilde{y}}(\mathbf{x})$ has the effect of maximally reducing their difference $\Delta$. While it is always possible to increase $\dot{c}_{\widetilde{y}}(\mathbf{x})$, it is not always possible to always reduce $\dot{c}_y(\mathbf{x})$. Our greedy approach, as implemented in Alg. 1, conditions each iteration's strategy based on whether $\dot{c}_y(\mathbf{x})$ can be reduced, i.e., whether $\dot{c}_y(\mathbf{x}) > 0$.

**Case #1**: $\dot{c}_y(\mathbf{x}) > 0$.

In each iteration, a single submodel prediction is changed. Changing one submodel prediction $f_t(\mathbf{x})$ from label $y$ to label $\widetilde{y}$ maximally decreases $\dot{c}_y(\mathbf{x})$. Moreover, transferring the vote to $\widetilde{y}$ also increases $\dot{c}_{\widetilde{y}}(\mathbf{x})$. No other allocation of the votes could reduce $\Delta$ more in particular since the order of the votes being reallocated does not matter.

**Case #2**: $\dot{c}_y(\mathbf{x}) = 0$.

No label can have negative votes so $\dot{c}_y(\mathbf{x})$ cannot be further reduced. Reducing the margin exclusively entails maximally increasing $\dot{c}_{\widetilde{y}}(\mathbf{x})$. Def. 8 and Alg. 1 transfer a vote from the plurality label $y_{\text{pl}} := \arg\max_{y' \in \mathcal{Y}} \dot{c}_{y'}(\mathbf{x})$ to label $\widetilde{y}$. Transferring the vote from the plurality label guarantees that $\dot{c}_{\widetilde{y}}(\mathbf{x})$ monotonically increases and no vote is ever transferred twice since $k < T$. □

**Proof of Lemma 12**

*Proof.* This proof follows directly from the proof of Thm. 3 with one difference. When training labels $y_1, \ldots, y_n$ may not be pristine, an adversary can use malicious training labels to modify a submodel prediction.

Each training label is considered by exactly one submodel. An adversarial label change has the same worst-case effect as an adversarial feature perturbation, meaning the certified robustness derivation in Thm. 3's proof applies here unchanged (other than the definition of robustness). Hence, similar to Eq. (3),

$$\widetilde{r} = \left\lfloor \frac{\text{GAP}_{\text{vote}}(y_{\text{pl}}, y_{\text{ru}})}{2} \right\rfloor. \tag{23}$$

□

**Proof of Lemma 13**

*Proof.* This proof follows directly from Lem. 12's proof. As above, a single adversarial label flip or feature perturbation still changes at most one submodel prediction. Training submodels with (deterministic) semi-supervised learning does not change the mechanics of the ensemble decision. Therefore, Lem. 12's certified guarantee derivation remains unchanged between partitioning the training instances versus partitioning the training labels with semi-supervised learning. □

**Proof of Lemma 14**

*Proof.* This proof follows directly from Wang et al.'s [WLF22a] Theorem 2; we direct the reader to the original paper for Wang et al.'s complete derivation. For brevity, we directly apply Wang et al.'s result below.

Both FPA and Wang et al.'s deterministic finite aggregation (DFA) train an ensemble of $\phi T$ submodels, with each submodel considering the union of $\phi$ disjoint sets of objects. The only difference between the two formulations is that DFA considers sets of training instances while FPA considers sets of features; the differences in the two methods' certified guarantees arise solely out of this one difference in formulation. DFA provides guarantees w.r.t. training instances, i.e., w.r.t. overlapping objects in the sets. Since FPA's sets instead contain feature dimensions, FPA certifies feature robustness.

Eq. (31)'s robustness bound is identical to Wang et al.'s Theorem 2, albeit with slightly different notation.

Note that Wang et al. do not contextualize their Theorem 2 w.r.t. top-$k$ predictions. Rather Wang et al. specify their guarantees w.r.t. correct/incorrect predictions, which is equivalent to top-1 accuracy. □

# C Related Work: Extended Discussion

Section 3 briefly summarizes work closely related to our certified defense, feature partition aggregation (FPA). Due to space, we deferred this more extensive discussion of related work to the supplement.

## C.1 Summarized Comparison of Closely Related Work

Table 6 provides a summarized comparison of the certified defenses most relevant to this work.

Table 6: **Certified defense comparison** for the primary methods considered in this work, namely: feature partition aggregation (FPA), randomized ablation (RA), (de)randomized smoothing (DRS), and deep partition aggregation (DPA). This comparison covers the types of guarantees each method provides as well as each method's model architecture.

| Property | Method | | | |
|---|---|---|---|---|
| | FPA (ours) | Random. Ablate. | (De)Rand. Smooth. | DPA |
| Evasion Defense | ✓ | ✓ | ✓* | |
| Patch Defense | ✓ | ✓ | ✓ | |
| Poison Defense | ✓ | | | ✓ |
| Backdoor Defense | ✓ | | | |
| Guarantee Type | Deterministic | Probabilistic | Deterministic | Deterministic |
| Guarantee Dimension | Feature-wise | Feature-Wise | Square Patch* | Instance-wise |
| Model Type | Ensemble | Smoothed | Smoothed | Ensemble |

## C.2 $\ell_0$-Norm Certified Evasion Defenses

These defenses represent the most closely related work. Given (test) instance $(\mathbf{x}, y)$, $\ell_0$-norm defenses certify the number of features that change in test instance $\mathbf{x}$ without changing prediction $f(\mathbf{x})$ (Def. 2).

Originally proposed by Levine and Feizi [LF20b] and subsequently improved by Jia et al. [Jia+22b], randomized ablation (RA) is the current state-of-the-art $\ell_0$-norm certified defense. RA is smoothing-based [CRK19; Ros+20]. Given some feature vector $\mathbf{x}$, RA's underlying classifier labels multiple random perturbations of $\mathbf{x}$; the model's *smoothed prediction* is the plurality label across these randomly perturbed predictions. Also generated from the perturbed predictions is a lower bound on the probability of predicting the plurality label as well as upper bounds on probabilities all other labels.[7] These probability bounds are then used to calculate RA's certified probabilistic guarantee $\rho$.

The type of perturbation dictates the type of certified guarantee smoothing yields. For example, to certify robustness against *label-flipping attacks*, Rosenfeld et al. [Ros+20] train multiple submodels, each using a different set of randomly perturbed training labels ($\mathbf{y}$). Randomized ablation uses a novel ablation strategy customized for $\ell_0$ attacks; specifically, for each ablated input of $\mathbf{x}$, $(d - e)$ randomly-selected features are "turned off" (i.e., ablated),[8] with the remaining $e$ features left unchanged. If an attacker perturbs $m$ unknown features in $\mathbf{x}$, then via combinatorics, we can determine the probability that one or more perturbed features intersect with the ablated input's kept features; if the feature-set intersection is empty, then the adversarial perturbation had no effect on the ablated prediction. RA combines this insight with the Neyman-Pearson Lemma to calculate $\ell_0$-norm robustness $\rho$ [Jia+22b].

Levine and Feizi's [LF20b] RA guarantees are often loose in practice, particularly for larger values of $\rho$. More recently, Jia et al. [Jia+22b] propose improved certification analysis that generates tight RA guarantees for top-1 predictions and almost tight guarantees for top-$k$ predictions.

---

[7]These upper and lower bounds are probabilistic given some user-specific hyperparameter $\alpha \in (0, 1)$.

[8]To mark a feature as turned-off, randomized ablation relies on a custom feature encoding that doubles the number of features. For details, see the original randomized ablation paper [LF20b].

Given the looseness of Levine and Feizi's [LF20b] certified guarantees, RA's effectiveness as certified patch defense is limited. To that end, Levine and Feizi [LF20a] propose *(de)randomized smoothing* (DRS) – a specialized version of RA for patch attacks. The primary differences between RA and DRS are:

1. RA provides $\ell_0$-norm guarantees (Def. 2) while DRS provides patch guarantees. Both of these guarantees apply to evasion attacks only.

2. As its name indicates, randomized ablation's smoothing process selects the set of kept (i.e., non-ablated) features uniformly at random. By restricting consideration to just patches, DRS restricts the number of possible attacks from order $\mathcal{O}(\binom{d}{m})$ to $\mathcal{O}(d)$. More practically, exponentially fewer possible perturbations allow DRS to certify a prediction with far fewer ablated inputs – so few that DRS's ablation set can usually be tested exhaustively.

3. Since RA considers only a random subset of the possible ablations, RA provides only *probabilistic* guarantees. By exhaustively testing a deterministic set of possible ablations, DRS provides *deterministic* guarantees.

Levine and Feizi's [LF20a] empirical evaluation of DRS considers exclusively square patches. Table 4 details how some rectangular patch shapes drop DRS's certified accuracy to 0%. Table 6 above lists DRS as providing guarantees w.r.t. specifically square patches since as Metzen and Yatsura [MY21] state in the BAGCERT paper, "we do not consider [(de)randomized smoothing] with column smoothing...[a] general patch defense, despite good performance for square patches and efficient certification analysis..."

Levine and Feizi [LF22] generalize the idea of (de)randomized smoothing's deterministic guarantees to $\ell_0$ attacks. Unlike FPA which provides certified feature guarantees (Def. 1), Levine and Feizi's [LF22] alternate method only provides $\ell_0$-norm robustness guarantees (Def. 2) and still generally requires ablated training.

To summarize the differences between the various certified $\ell_0$ and patch defenses:

1. FPA provides guarantees over the union of $\ell_0$ evasion, backdoor, and poisoning attacks, while RA and DRS provide no training robustness guarantees.

2. FPA trains an ensemble of (non-smoothed) classifiers, while RA and DRS train a single smoothed classifier.

3. During both training and inference, feature ablation functionally marks any ablated feature as missing; this generally restricts RA and DRS to model architectures that are robust under missing data. In practice, feature ablation works best when combined with parametric model architectures (e.g., neural networks) that are trained using first-order methods. Ablated training and inference cannot be directly combined with tree-based methods such as gradient-boosted decision trees (GBDTs). By contrast, FPA supports any submodel architecture. Therefore, unlike RA and DRS, FPA can use whichever submodel architectures works best for a given application.

4. FPA and RA consider more general $\ell_0$ attacks, while DRS considers more restrictive patch attacks.

5. FPA and DRS provide deterministic guarantees, while RA provides only probabilistic guarantees.

6. DRS's deterministic ablation patterns (e.g., band smoothing and block smoothing) generally perform poorly when used as deterministic feature partitions.

Calzavara et al. [Cal+21] propose a *binary classification only* $\ell_0$-norm certified defense based on decision tree ensembles. Like FPA, Calzavara et al. use feature partitioning to bootstrap their guarantees. However, Calzavara et al.'s certification procedure is NP-complete in the worst case via reduction to partial set cover. Moreover, each of Calzavara et al.'s models certifies a single $\ell_0$-norm robustness level, potentially requiring a different model to be trained for each target robustness level $\rho$.

## C.3 Instance-wise Certified Poisoning Defenses

The second class of defenses related to FPA certify robustness against *instance-wise data poisoning*. Specifically, these methods provide pointwise guarantees on the number of arbitrary *instances* that can be inserted into or deleted from the training set without changing model prediction $f(\mathbf{x})$.[9] Def. 9 formalizes instance-wise poisoning guarantees as commonly defined in related work [LF21; WLF22b; WLF22a; Rez+23; WF23], where function $d_{\text{sym}}$ denotes the *symmetric difference*.

---

[9]Recall that FPA's *certified feature robustness* (Def. 1) provides guarantees on the number of *features* – training or test – an attacker can perturb. FPA does not certify robustness w.r.t. instances like DPA.

**Def. 9. Instance-wise Certified Poisoning Robustness** *Given model $f$ trained on training set $\mathcal{D} = \{(\mathbf{x}_i, y_i)\}_{i=1}^{n}$ and model $f'$ trained on $\mathcal{D}' = \{(\mathbf{x}_j, y_j)\}_{j=1}^{m}$, instance-wise certified poisoning robustness $\bar{\rho} \in \mathbb{N}$ is a pointwise, deterministic guarantee w.r.t. instance $\mathbf{x}$ where $|d_{\text{sym}}(\mathcal{D}, \mathcal{D}')| \leq \bar{\rho} \implies f(\mathbf{x}) = f'(\mathbf{x})$.*

The first poisoning defense to provide non-trivial instance-wise guarantees was *deep partition aggregation* (DPA) [LF21]. Described briefly, let $h_{\text{tr}} : \mathbb{R}^d \to [T]$ be a deterministic function that partitions the instance space into $T$ disjoint subregions. DPA trains an ensemble of $T$ deterministic submodels where each submodel's training set is drawn from a different $h_{\text{tr}}$ subregion. Levine and Feizi's [LF21] formulation of DPA relies on plurality voting as the decision function. FPA is heavily inspired by DPA, so we chose to name our method similarly.

Rezaei et al. [Rez+23] propose *run-off elections* – an alternate DPA decision function and certification procedure. Run-off elections require no retraining of the DPA ensemble, meaning run-off can increase DPA's certified guarantees essentially for free.

Additional instance-wise poisoning defenses include Jia et al.'s [Jia+22a] nearest-neighbor defense and Wang et al.'s [WLF22a] finite aggregation.

A major strength of FPA is its ability to directly leverage the properties implicit in existing voting-based techniques. More specifically, FPA can directly leverage both plurality voting and run-off election decision functions to maximize our certified guarantees.

## C.4 Certified Defenses against the Union of $\ell_p$ Attacks

Feature partition aggregation (FPA) is the first certified defense robust against the union of $\ell_0$ evasion, backdoor, and poisoning attacks. To our knowledge, the only other certified method robust over this union of attack types is Weber et al.'s [Web+20] *robustness against backdoors* (RAB) defense, which focuses on $\ell_2$ robustness. RAB extends randomized smoothing by training an ensemble of smoothed classifiers. Each smoothed RAB submodel is trained on a unique *smoothed training set* where i.i.d. random (Gaussian) noise is added to each training instance's feature vector.

To better understand RAB's certified guarantees, let $\delta_i \in \mathbb{R}^d$ denote the adversarial perturbation added to the $i$-th training instance, $\delta_\mathbf{x} \in \mathbb{R}^d$ denote the backdoor trigger added to target test instance $\mathbf{x}$, and $b \in \mathbb{R}_{\geq 0}$ denote RAB's certified guarantee. Then, RAB defines a prediction as pointwise certifiably robust whenever

$$\sqrt{\sum_{i=1}^{n} \|\delta_i\|_2^2} < b \tag{24}$$

implies (with high probability) that clean and poisoned training sets would have the same prediction for feature vector $\mathbf{x} + \delta_\mathbf{x}$.

In practice, RAB provides comparatively small robustness guarantees $b$. For example, consider CIFAR10 where RAB's maximum reported certified robustness is $b_{\text{max}} \leq 3$ [Web+20, Fig. 4b]. An attacker could violate this bound by arbitrarily modifying as few as *three RGB pixels across the entire training set*. In contrast, FPA can certify 41.0% of CIFAR10 predictions up to 1.25M arbitrarily perturbed pixels (see Takeaway #5 in Sec. 6.2).

# D  Certifying a Top-$k$ Prediction

In line with Jia et al.'s [Jia+22b] extension of randomized ablation to top-$k$ certification, below we generalize FPA with plurality voting to top-$k$ predictions below. For simplicity of presentation, we restrict consideration to the meaningful case where $k < T$.

**Updated Nomenclature**  $f$'s plurality-voting decision function generalizes to top-$k$ prediction as

$$f(\mathbf{x}; k) := \arg\max_{\mathcal{Y}_k \subset \mathcal{Y},\, |\mathcal{Y}_k|=k} \sum_{y \in \mathcal{Y}_k} \dot{c}_y(\mathbf{x}), \qquad (25)$$

where ties are broken by selecting the smallest class indices.

**Extending Plurality Voting to Top-$k$**  Intuitively, Thm. 3's certified feature robustness $r$ quantifies the number of submodel "votes" that can switch from plurality label $y_{\text{pl}}$ to runner-up label $y_{\text{ru}}$ without changing the model's prediction. The simplicity of top-1 predictions permits Eq. (3)'s neat closed form. Thm. 3's guarantee $r$ can also be calculated greedily, where submodel "votes" are switched, one at a time, from $y_{\text{pl}}$ to $y_{\text{ru}}$, with the vote-flipping stopping right before the plurality label changes. While top-$k$ feature robustness under plurality voting does not have a convenient closed form like Eq. (3), an (optimal) greedy strategy still applies.

Intuitively, a label $y$ is not in the top $k$ if there exist $k$ labels with more votes. Hence, two approaches to eject a label $y$ from the top $k$ are: (1) reduce $\dot{c}_y(\mathbf{x})$, the number of submodels that predict $y$, and (2) increase the number of votes for $\widetilde{y}$, i.e., the label with $(k+1)$-th most votes. Note that for $k > 1$, label $\widetilde{y}$ may change after each greedy iteration; it is this interaction that complicates providing a compact closed-form top-$k$ guarantee $r$ that is tight.

---

**Algorithm 1** Top-$k$ Greedy Robustness Certification under Plurality Voting

---

**Input:** Instance $\mathbf{x} \in \mathcal{X}$; target label $y \in \mathcal{Y}$; $k \in \mathbb{N}$; label vote counts $\forall_{y' \in \mathcal{Y}}\, \dot{c}_{y'}(\mathbf{x})$
**Output:** Certified feature robustness $r$
1: $r \leftarrow -1$
2: **while** $\dot{c}_y(\mathbf{x})$ is in the top $k$ **do**
3: $\quad \widetilde{y} \leftarrow$ Label with the $(k+1)$-th most votes
4: $\quad$ **if** $\dot{c}_y(\mathbf{x}) > 0$ **then**
5: $\quad\quad \dot{c}_y(\mathbf{x}) \leftarrow \dot{c}_y(\mathbf{x}) - 1$
6: $\quad$ **else**
7: $\quad\quad y_{\text{pl}} \leftarrow \arg\max_{y'} \dot{c}_{y'}(\mathbf{x})$ ▷ Plurality label
8: $\quad\quad \dot{c}_{y_{\text{pl}}}(\mathbf{x}) \leftarrow \dot{c}_{y_{\text{pl}}}(\mathbf{x}) - 1$
9: $\quad \dot{c}_{\widetilde{y}}(\mathbf{x}) \leftarrow \dot{c}_{\widetilde{y}}(\mathbf{x}) + 1$
10: $\quad r \leftarrow r + 1$ ▷ Update certified robustness
11: **return** $r$

---

Alg. 1 formalizes the above intuition into a complete method to calculate top-$k$ certified feature robustness $r$. With linear-time sorting (e.g., counting sort), Alg. 1 has $\mathcal{O}(T)$ time complexity – same as plurality-voting top-1 certification.[10]

**Theorem 10. Top-$k$ Greedy Strategy Optimality**  *Alg. 1 returns plurality voting's top-$k$ certified feature robustness $r$ that is tight under worst-case perturbations.*

Alg. 1 addresses an edge case to ensure $r$ is tight. Based on how ties are broken, a label $y$ can be in the top $k$ without receiving any votes (i.e., $\dot{c}_y(\mathbf{x}) = 0$). In such cases, Alg. 1 transfers votes from plurality label $y_{\text{pl}}$. Perturbing $y_{\text{pl}}$ ensures $\dot{c}_{\widetilde{y}}(\mathbf{x})$ is monotonically increasing. Like $\widetilde{y}$, the plurality label can change between loop iterations.

---

[10] With a more sophisticated greedy strategy, certifying a top-$k$ prediction under plurality voting requires no more than $\mathcal{O}(k)$ greedy iterations. We provide the less efficient Alg. 1 here for simplicity. Our source code implements both greedy algorithms.

**Generalizing our Top-$k$ Greedy Algorithm**  Observe that Alg. 1 deals only in submodel vote counts (i.e., $\dot{c}_{y'}(\mathbf{x})$) and is agnostic to how these independent votes are generated – be it over partitioned features or otherwise. Multiple existing certified defenses (e.g., deep partition aggregation [LF21] and the nearest neighbor-based instance-wise poisoning defense [Jia+22a]) are top-1 only and voting-based, with the votes independent. Alg. 1 can be directly reused to generalize those existing certified defenses to provide robustness guarantees over top-$k$ predictions. Alg. 1 also applies to alternate FPA formulations with non-pristine training labels (see suppl. Sec. E).

**Combining our Top-$k$ Greedy Algorithm with Run-Off**  Sec. 4.2 describes two possible ways an attacker can perturb run-off prediction $y_{\text{RO}}$. Consider Case #2 where the goal is to eject $y_{\text{RO}}$ from round #1's top-two labels. Observe that this case reduces to calculating $y_{\text{RO}}$'s top-2 robustness. Rezaei et al.'s [Rez+23] dynamic programming-based formulation in Eq. (7) could be directly replaced by Alg. 1's greedy approach. Sec. 4.2's presentation was chosen to better align with Rezaei et al.'s preprint formulation (while correcting an error in the definition of `dp`).

# E  On a Sparse Attacker that Modifies Training Labels

Sec. 2's base formulation of feature partition aggregation trains each submodel on a subset of the features from all training instances. Each submodel also considers full label vector $\mathbf{y} := [y_1, \ldots, y_n]$ (see Fig. 1). In the worst case, a single adversarial label flip could manipulate all $T$ predictions, invalidating FPA's guarantees. Whether an attacker is able to manipulate the training labels is application dependent. Previous work commonly views *clean-label attacks* (where $\mathbf{y}$ is pristine) as the stronger threat model [Che+17; Sha+18; Hua+20; Wal+21; YHL23]. To simplify the formulation and allow for a more direct comparison to existing work, we chose for our primary presentation to assume clean labels. Nonetheless, FPA's underlying formulation can be generalized to a threat model where an adversary can modify training labels. Def. 11 formalizes a joint robustness guarantee over feature perturbations and training-label flips.

**Def. 11. Certified Feature and Label-Flipping Robustness**  *Given training set $(\mathbf{X}, \mathbf{y})$, model $f'$ trained on $(\mathbf{X}', \mathbf{y}')$, and arbitrary feature vector $\mathbf{x}' \in \mathcal{X}$, certified feature and label-flipping robustness $\widetilde{r} \in \mathbb{N}$ is a pointwise, deterministic guarantee w.r.t. instance $(\mathbf{x}, y)$ where $|\mathbf{X} \ominus \mathbf{X}' \cup \mathbf{x} \ominus \mathbf{x}'| + |\mathbf{y} \ominus \mathbf{y}'| \leq \widetilde{r} \implies y = f'(\mathbf{x}')$.*

Similar to certified feature robustness $r$ in Def. 1, certified feature and label robustness $\widetilde{r}$ is <u>not</u> w.r.t. feature values. Instead, $\widetilde{r}$ provides a stronger guarantee allowing all values – training and test – for a feature to be perturbed. Robustness $\widetilde{r}$ arbitrarily divides between feature perturbations and training-label flips.

Below we propose two extended FPA formulations, which provide certified feature and label-flipping robustness guarantees. We focus on plurality voting below with the extension to run-off straightforward.

## E.1  Training Instance Partitioning

FPA's base formulation is particularly vulnerable to adversarial label flipping since each submodel considers full label vector $\mathbf{y}$. This vulnerability's fix is very simple: partition *both* the features and training instances across the $T$ submodels. Under this alternate formulation, a single adversarial label flip affects at most one submodel prediction, i.e., the submodel trained on that instance. Lem. 12 formalizes certified feature and label-flipping robustness for FPA under training instance partitioning. Like Thm. 3, Lem. 12 generalizes to certify top-$k$ predictions via Alg. 1. Alternatively, $\widetilde{r}$ could be certified using run-off elections similar to Theorem 4.

**Lemma 12. Certified Robustness with Partitioned Training Instances**  *Given feature partition $\mathcal{S}_1, \ldots, \mathcal{S}_T$, let $f$ be an ensemble of $T$ submodels using the plurality-voting decision function. Let $h_{tr} : \mathcal{X} \times \mathcal{Y} \to [T]$ be a deterministic function that partitions the instance space. The $t$-th submodel is trained exclusively on the features in set $\mathcal{S}_t$ as well as only those training instances $(\mathbf{x}_i, y_i)$ where $h_{tr}(\mathbf{x}_i, y_i) = t$. Then, for instance $(\mathbf{x}, y)$, the pointwise certified feature and label-flipping robustness is*

$$\widetilde{r} = \left\lfloor \frac{\mathrm{GAP}_{\mathrm{vote}}(y_{\mathrm{pl}}, y_{\mathrm{ru}})}{2} \right\rfloor. \tag{26}$$

While Lem. 12's guarantees appear *similar* to existing certified poisoning defenses such as *deep partition aggregation* (DPA) [LF21], there is a subtle yet important difference. As explained in Sec. 3, DPA's threat model encompasses only data poisoning attacks, meaning test instance $\mathbf{x}$ is assumed pristine. DPA does <u>not</u> certifiably improve the model's robustness under backdoor or evasion attacks when $\mathbf{x}$ is adversarially manipulated. By contrast, Lem. 12 provides certifiable robustness under sparse poisoning, backdoor, and evasion attacks – as well as adversarial label flipping. There exist backdoor attacks where Lem. 12 is provably robust but DPA is not (e.g., Gu et al.'s [Gu+19] pixel-based attacks) and vice versa.

Lem. 12 is no free lunch. Partitioning the training instances across the ensemble entails that each submodel is trained on even fewer data. This can degrade submodel performance, potentially degrading the certified robustness [WLF22b]. Next, we modify the above formulation to restore some of the feature information that is lost when the training instances are partitioned.

## E.2 Training Label Partitioning with Semi-Supervised Learning

Sec. 2's threat model places no constraint on the poisoning rate, i.e., the fraction of the training instances an attacker may adversarially perturb. In other words, under this threat model, perturbing a feature for one instance is equivalent, from a certification perspective, to perturbing that feature for all instances.

In Section E.1 above, our revised feature partition aggregation (FPA) formulation above discards significant feature information. Formally, for training instance $(\mathbf{x}_i, y_i)$ assigned to $t$-th submodel model (i.e., $h_{\text{tr}}(\mathbf{x}_i, y_i) = t$), features dimensions $[d] \setminus \mathcal{S}_t$ in $\mathbf{x}_i$ are *not used in the training of any submodel*. In other words, $\mathbf{x}_i$'s feature dimensions $[d] \setminus \mathcal{S}_t$ are totally ignored. Since our threat model allows a 100% poisoning rate, discarding these features does not improve the theoretical robustness.

Rethinking Sec. E.1, the primary motivation for partitioning the training instances was to ensure that a single adversarial label flip did not affect more than one submodel. To achieve that, the formulation above not only restricts each submodel's access to some training labels, it also restricts access to the corresponding training instance's feature information. This is heavy-handed, and a more careful partitioning is possible.

This section's revised FPA formulation is inspired by semi-supervised learning. The $t$-th submodel still considers the $\mathcal{S}_t$ columns of matrix $\mathbf{X}$. The sole difference is in the training-label vector used by each submodel. Rather than partitioning the training instances like in the previous section, our semi-supervised FPA uses function $h_{\text{tr}}$ to partition just the training *labels*. The $t$-th submodel treats as unlabeled any training instance $(\mathbf{x}_i, y_i)$ were $h_{\text{tr}}(\mathbf{x}_i, y_i) \neq t$. Put simply, the only difference between the submodel training sets of our base and semi-supervised formulations lies in the training labels available to each submodel. Both formulations train each submodel on the same feature submatrix.

Lem. 13 formalizes the certified feature and label-flipping robustness (Def. 11) for FPA under training label partitioning with semi-supervised learning. Observe that Eqs. (26) and (27) define the certified feature and label-flip robustness $\widetilde{r}$ identically. Like Thm. 3 and Lem. 13 above, Lem. 13 generalizes to certify top-$k$ predictions via Alg. 1. Again, Lem. 13 can be trivially modified to instead certify feature and label-flipping robustness using run-off elections similar to Theorem 4.

**Lemma 13. Certified Robustness with Partitioned Training Labels** *Given feature partition $\mathcal{S}_1, \ldots, \mathcal{S}_T$, let $f$ be an ensemble of $T$ submodels using the plurality-voting decision function. Let $h_{tr} : \mathcal{X} \times \mathcal{Y} \to [T]$ be a deterministic function that partitions the instance space. The $t$-th submodel is trained exclusively on the features in set $\mathcal{S}_t$ as well as the training labels for those training instances $(\mathbf{x}_i, y_i)$ where $h_{tr}(\mathbf{x}_i, y_i) = t$. For all training instances $(\mathbf{x}_i, y_i)$ where $h_{tr}(\mathbf{x}_i, y_i) \neq t$, the $t$-th submodel treats the instance as unlabeled. Then, for instance $(\mathbf{x}, y)$, the pointwise certified feature and label-flipping robustness is*

$$\widetilde{r} = \left\lfloor \frac{\text{GAP}_{\text{vote}}(y_{\text{pl}}, y_{\text{ru}})}{2} \right\rfloor. \tag{27}$$

Whether partitioning the training labels (Sec. E.2) or the training instances (Sec. E.1) yields larger certified guarantees is an empirical question, whose answer depends on the application and semi-supervised learning algorithm.

# F  On Overlapping Submodel Feature Sets

feature partition aggregation does not necessarily require that feature subsets $\mathcal{S}_1, \ldots, \mathcal{S}_T$ be a partition of the full feature set $[d]$. Rather, the feature subsets can partially overlap, but the certification analysis becomes NP-hard in the general case via reduction to (partial) set cover [HL22, Lem. 11].

Recall also that deep partition aggregation (DPA) is a certified defense against poisoning attacks under a limited poisoning rate. Like FPA, DPA trains submodels on partitioned sets – specifically, partitioned training instances. Wang et al.'s [WLF22a] *deterministic finite aggregation* (DFA) extends DPA where submodels are trained on *overlapping* instance sets. Just as FPA with partitioned feature sets can be viewed as the *transpose* of DPA, FPA with overlapping feature sets can be viewed as the transpose of Wang et al.'s DFA. Below we formulate FPA with overlapping feature sets as inspired by deterministic finite aggregation.

Rather than partitioning feature set $[d]$ into $T$ subsets, consider partitioning $[d]$ into $\phi T$ disjoint subsets where $\phi \in \mathbb{N}$. By definition, it should hold that $\phi T \leq d$. Otherwise, some subsets in the partition will be empty by the pigeonhole principle.

In our base FPA formulation, each submodel is trained on approximately $\frac{1}{T}$-th of the features, and each feature subset is assigned to exactly one submodel. For FPA with overlapping features, each submodel is still trained on $\frac{1}{T}$-th of the features. However, since each feature set is now $\frac{1}{\phi}$-th the size, each overlapping submodel is assigned $\phi$ feature subsets. Following Wang et al. [WLF22a], each feature subset is similarly assigned to $\phi$ submodels. Hence, $\phi$ is referred to as the feature subsets' *spread degree*.

Deterministic function $h_\mathcal{S} : [\phi T] \to [\phi T]^\phi$ maps the $\phi T$ feature subsets to the $\phi T$ submodels. Our overlapping features empirical evaluation below defines $h_\mathcal{S}$ identically to Wang et al.'s $h_{\text{spread}}$ function. Formally, let $\mathcal{T} \subset [\phi T]$ be a set drawn uniformly at random without replacement from $[\phi T]$ where $|\mathcal{T}| = \phi$. Then, the set of submodels that use feature partition $l \in [\phi T]$ is

$$h_\mathcal{S}(l) := \{\tau + l \mod \phi T : \tau \in \mathcal{T}\}. \tag{28}$$

Since $\mathcal{T}$ is constructed randomly, overlapping feature sets more closely resemble balanced random partitioning than deterministic partitioning.

There are two important differences in the analysis of FPA with partitioned versus overlapping feature sets. First, under partitioned feature sets, a single perturbed feature affects exactly one submodel. For overlapping features, each feature subset is used in the training of $\phi$ submodels, meaning a single perturbed feature affects $\phi$ submodel votes. Second, under partitioned feature sets, certification analysis exclusively considered the minimum number of models required for the runner-up label to overtake the plurality label. Under overlapping features, the runner-up label may not be the most efficient to perturb, meaning all labels must be considered in certification analysis.

The next section formalizes the certified feature robustness under overlapping feature sets with plurality voting.

## F.1  Certified Feature Robustness with Overlapping Feature Sets

Recall that for any $y \in \mathcal{Y}$ and $\mathbf{x} \in \mathcal{X}$,

$$\dot{c}_y(\mathbf{x}) := |\{t \in [T] : f_t(\mathbf{x}) = y\}|$$

denotes the number of submodels that predict label $y$ for $\mathbf{x}$. Given $\phi T$ disjoint feature subsets where $\bigsqcup_{l=1}^{\phi T} \mathcal{S}_l = [\phi T]$, let

$$\dot{c}_y(\mathbf{x}; l) := |\{t \in [T] : f_t(\mathbf{x}) = y \wedge t \in h_\mathcal{S}(l)\}|, \tag{29}$$

denote the number of submodels that both use feature subset $\mathcal{S}_l$ and predict label $y$ for $\mathbf{x}$. Define the multiset w.r.t. $\mathbf{x} \in \mathcal{X}$ as

$$\Delta_{(y,y')} := \{\phi + \dot{c}_y(\mathbf{x}; l) - \dot{c}_{y'}(\mathbf{x}; l) : l \in [\phi T]\}, \tag{30}$$

and let $\Delta_{(y,y')}^{r'}$ denote the sum of the $r' \in \mathbb{N}$ largest elements in multiset $\Delta_{(y,y')}$.

Lem. 14 defines the certified feature robustness with overlapping feature sets, plurality voting, and fixed spread degree $\phi$. Lem. 14 follows directly from Wang et al.'s [WLF22a] Thm. 2.

**Lemma 14. Certified Feature Robustness with Overlapping Feature Sets and Fixed Spread Degree**  *Given submodel feature partition* $\mathcal{S}_1, \ldots, \mathcal{S}_{\phi T}$ *and function* $h_\mathcal{S}$*, let* $f$ *be a voting-based ensemble of* $\phi T$ *submodels* *using plurality-voting, where each deterministic submodel $f_t$ uses the features in set*

$$\bigsqcup_{\substack{l \in [\phi T] \\ t \in h_{\mathcal{S}}(l)}} \mathcal{S}_l.$$

*Then the pointwise certified feature robustness of prediction is $y \coloneqq f(\mathbf{x})$ is $r = \min_{y' \neq y} r_{y'}$ where*

$$r_{y'} \coloneqq \underset{r' \in \mathbb{N}}{\arg\max} \text{ s.t. } \Delta_{(y,y')}^{r'} \leq \dot{c}_y(\mathbf{x}) - \dot{c}_{y'}(\mathbf{x}) - \mathbb{1}\big[y' < y\big] \tag{31}$$

The next section discusses the limitations of training FPA's submodels on overlapping feature subsets.

## F.2 Limitations of Overlapping Feature Sets

Combining FPA with overlapping feature sets has two primary limitations.

First, overlapping feature sets increase the computational cost versus Thm. 3's disjoint feature sets – even without an NP-hard optimization. One of FPA's key advantages over previous related methods like randomized ablation is FPA's computational efficiency (Tab. 3). FPA with disjoint feature sets has computational complexity in $\mathcal{O}(T)$. In contrast, FPA as formulated in Lem. 14 with overlapping feature sets has computation complexity in $\mathcal{O}(\phi T)$. Any performance gains derived from overlapping features need to be weighed against the multiplicative increase in training and certification time.

The other major limitation is that supplemental Sec. D's greedy algorithm does not apply to overlapping feature sets. Like any NP-hard problem, greedy methods may overestimate the solution necessitating an approximation factor to address any overestimation. A greedy-based, top-$k$ certification algorithm for overlapping feature sets is left as future work.

As an alternative to Wang et al.'s [WLF22a] closed-form lower bound for the certified robustness on overlapping sets of instances, Hammoudeh and Lowd [HL23b] use an integer linear program to find the optimal certified robustness. In short, Hammoudeh and Lowd's formulation trades a better certified bound for a potentially (significantly) more complex optimization. Hammoudeh and Lowd's [HL23b] linear program could be modified to determine overlapping FPA's optimal top-$k$ robustness.

## F.3 Empirical Evaluation of Overlapping Feature Sets for Certified Feature Robustness

This section evaluates FPA's performance with disjoint and overlapping feature sets under plurality voting. The results for CIFAR10 are in Tables 7 and 8. MNIST's results are in Tables 9, 10, and 11. Weather's results are in Tables 12 and 13. Beyond the overlapping feature sets, the evaluation setup is identical to Sec. 6.

Recall that under overlapping features, the total number of feature partitions is $\phi T$. As discussed above, this quantity is functionally bounded by the dataset dimension $d$. For each model configuration below, we evaluate performance with spread degree $\phi$ set as large as possible given $T$ without exceeding the dataset's corresponding dimension $d$.

We briefly summarize these experiments' takeaways.

**Takeaway #1**: *The benefits of overlapping feature sets is largest for smaller $T$ values.* We see this trend for all three datasets. For example with CIFAR10, overlapping feature sets improved random partitioning's performance by up to 3.5 percentage points when $T = 25$. By contrast, for CIFAR10 with $T = 115$, overlapping feature sets improved the performance by only 0.6 percentage points. We conjecture that the primary cause of this behavior is that $T$ and the maximum spread degree are inversely related. Since feature dimension $d$ is fixed, larger $T$ restricts $\phi$ and in turn the potential benefits of overlapping feature sets.

By comparison, the spread degree of Wang et al.'s [WLF22a] DFA is capped by the number of training instances. For modern datasets, the training set's size is much larger than the feature dimension. We believe this partially explains why overlapping sets are more useful for certified poisoning defenses than FPA.

**Takeaway #2**: *For vision datasets, deterministic partitioning generally outperforms overlapping feature sets.* The trend is most visible for CIFAR10 where overlapping feature sets only marginally outperformed strided partitioning under one small case. By contrast, CIFAR10 deterministic partitioning outperformed overlapping feature sets by multiple percentage points in many cases. For MNIST, overlapping feature sets did outperform strided deterministic partitioning in particular when $r$ is small. In many of those cases, random partitioning also performed as well as or better than strided partitioning.

**Takeaway #3**: *Overlapping feature sets reduce the certified accuracy's variance for random partitioning.* For Weather [Mal+21], we report both the certified accuracy's mean and standard deviation. As spread degree $\phi$ increased, the certified accuracy's variance decreased by up to two-thirds. In short, overlapping feature sets mitigate the effect of poor feature partitions, which can severely degrade random partitioning's performance.

**Takeaway #4**: *The benefits of overlapping feature sets decrease as $r$ increases.* This trend is consistent across all three datasets over all $T$ values. At the largest certified robustness values, overlapping feature sets can even significantly *underperform* random partitioning. We theorize the primary cause for this phenomenon is that while guarantees for disjoint feature sets are tight, Lem. 14 only lower bounds overlapping feature set's maximum certifiable robustness. As $r$ increases, this looseness becomes increasingly visible.

Table 7: **CIFAR10 Overlapping Feature Sets** ($T = 25$): CIFAR10 certified accuracy for our sparse defense, feature partition aggregation (FPA), with $T = 25$. "Random" denotes balanced random partitioning with disjoint submodel feature sets (i.e., spread degree $\phi = 1$). "Overlapping" denotes that the submodel feature sets were trained using Sec. F.1's overlapping feature set formulation with the corresponding spread degree ($\phi$) specified above each column. "Strided" denotes deterministic strided partitioning with disjoint submodel feature sets (Eq. (33)). The configuration with the best mean certified accuracy is shown in **bold**.

| Cert. Robust. | Random | Overlapping | | | Strided |
|---|---|---|---|---|---|
| | | $\phi = 10$ | $\phi = 20$ | $\phi = 40$ | |
| 1 | 72.1 | 73.2 | 73.6 | 73.7 | **76.1** |
| 4 | 60.8 | 62.4 | 63.6 | 64.3 | **67.6** |
| 8 | 42.5 | 43.6 | 44.4 | 45.8 | **53.0** |
| 12 | 14.2 | 13.1 | 12.8 | 12.7 | **25.0** |

Table 8: **CIFAR10 Overlapping Feature Sets** ($T = 115$): CIFAR10 certified accuracy for our sparse defense, feature partition aggregation (FPA), with $T = 115$. "Random" denotes balanced random partitioning with disjoint submodel feature sets (i.e., spread degree $\phi = 1$). "Overlapping" denotes that the submodel feature sets were trained using Sec. F.1's overlapping feature set formulation with the corresponding spread degree ($\phi$) specified above each column. "Strided" denotes deterministic strided partitioning with disjoint submodel feature sets (Eq. (33)). The configuration with the best mean certified accuracy is shown in **bold**.

| Cert. Robust. | Random | Overlapping | | Strided |
|---|---|---|---|---|
| | | $\phi = 4$ | $\phi = 8$ | |
| 1 | 61.3 | 61.5 | **61.6** | 61.2 |
| 10 | 49.6 | 49.6 | 50.2 | **51.2** |
| 20 | 36.9 | 36.8 | 37.3 | **40.0** |
| 30 | 25.1 | 24.7 | 24.8 | **29.1** |
| 40 | 14.7 | 14.1 | 14.0 | **18.9** |
| 50 | 5.7 | 5.5 | 5.4 | **8.9** |

Table 9: **MNIST Overlapping Feature Sets** ($T = 25$): MNIST certified accuracy for our sparse defense, feature partition aggregation (FPA), with $T = 25$. "Random" denotes balanced random partitioning with disjoint submodel feature sets (i.e., spread degree $\phi = 1$). "Overlapping" denotes that the submodel feature sets were trained using Sec. F.1's overlapping feature set formulation with the corresponding spread degree ($\phi$) specified above each column. "Strided" denotes deterministic strided partitioning with disjoint submodel feature sets (Eq. (33)). The configuration with the best mean certified accuracy is shown in **bold**.

| Cert. Robust. | Random | Overlapping | | | Strided |
|---|---|---|---|---|---|
| | | $\phi = 10$ | $\phi = 20$ | $\phi = 30$ | |
| 1 | 93.6 | 94.7 | 94.9 | **95.0** | 94.1 |
| 4 | 84.0 | 86.5 | 87.4 | **87.6** | 86.5 |
| 8 | 57.5 | 59.9 | 60.6 | 61.8 | **66.4** |
| 12 | 11.3 | 11.5 | 10.5 | 10.8 | **20.1** |

Table 10: **MNIST Overlapping Feature Sets** ($T = 60$): MNIST certified accuracy for our sparse defense, feature partition aggregation (FPA), with $T = 60$. "Random" denotes balanced random partitioning with disjoint submodel feature sets (i.e., spread degree $\phi = 1$). "Overlapping" denotes that the submodel feature sets were trained using Sec. F.1's overlapping feature set formulation with the corresponding spread degree ($\phi$) specified above each column. "Strided" denotes deterministic strided partitioning with disjoint submodel feature sets (Eq. (33)). The configuration with the best mean certified accuracy is shown in **bold**.

| Cert. Robust. | Random | Overlapping | | Strided |
|---|---|---|---|---|
| | | $\phi = 6$ | $\phi = 12$ | |
| 1 | 80.8 | 82.6 | **82.7** | 80.8 |
| 5 | 64.9 | 67.3 | **68.4** | 66.6 |
| 10 | 43.1 | 43.9 | 46.5 | **46.9** |
| 15 | 26.1 | 25.9 | 27.1 | **29.2** |
| 20 | 14.2 | 14.2 | 14.6 | **16.1** |
| 25 | 5.2 | 5.2 | 5.7 | **6.3** |

Table 11: **MNIST Overlapping Feature Sets** ($T = 80$): MNIST certified accuracy for our sparse defense, feature partition aggregation (FPA), with $T = 80$. "Random" denotes balanced random partitioning with disjoint submodel feature sets (i.e., spread degree $\phi = 1$). "Overlapping" denotes that the submodel feature sets were trained using Sec. F.1's overlapping feature set formulation with the corresponding spread degree ($\phi$) specified above each column. "Strided" denotes deterministic strided partitioning with disjoint submodel feature sets (Eq. (33)). The configuration with the best mean certified accuracy is shown in **bold**.

| Cert. Robust. | Random | Overlapping | | Strided |
|---|---|---|---|---|
| | | $\phi = 6$ | $\phi = 9$ | |
| 1 | 72.2 | 73.8 | **74.5** | 68.0 |
| 8 | 46.3 | 47.2 | **48.3** | 46.2 |
| 16 | 24.0 | 24.0 | 24.5 | **25.5** |
| 24 | 12.0 | 12.1 | 12.1 | **13.2** |
| 32 | 3.1 | 2.6 | 3.2 | **5.3** |

Table 12: **Weather Overlapping Feature Sets** ($T = 11$): Certified accuracy mean and standard deviation for the Weather tabular dataset for FPA (FPA) with $T = 11$. "Random" denotes balanced random partitioning with disjoint submodel feature sets (i.e., spread degree $\phi = 1$). "Overlapping" denotes that the submodel feature sets were trained using Sec. F.1's overlapping feature set formulation with the corresponding spread degree ($\phi$) specified above each column. The configuration with the best mean certified accuracy is shown in **bold**. Results averaged over 10 trials.

| Cert. Robust. | Random | Overlapping | | | |
|---|---|---|---|---|---|
| | | $\phi = 3$ | $\phi = 7$ | $\phi = 9$ | $\phi = 11$ |
| 1 | 78.9 ± 1.5 | 80.1 ± 1.1 | 79.8 ± 0.4 | 80.1 ± 0.4 | **80.7 ± 0.5** |
| 2 | 70.6 ± 2.5 | 72.6 ± 1.9 | **73.2 ± 0.9** | 72.1 ± 0.7 | **73.2 ± 0.9** |
| 3 | 58.9 ± 3.6 | 61.2 ± 3.0 | 61.8 ± 1.7 | 61.7 ± 1.1 | **61.9 ± 1.5** |
| 4 | 42.5 ± 4.4 | 43.7 ± 3.8 | 40.7 ± 2.7 | 43.9 ± 1.5 | **44.2 ± 1.9** |
| 5 | **19.4 ± 4.4** | 18.2 ± 2.9 | 17.2 ± 2.6 | 17.3 ± 1.5 | 17.5 ± 1.3 |

Table 13: **Weather Overlapping Feature Sets** ($T = 31$): Certified accuracy mean and standard deviation for the Weather tabular dataset for FPA (FPA) with $T = 31$. "Random" denotes balanced random partitioning with disjoint submodel feature sets (i.e., spread degree $\phi = 1$). "Overlapping" denotes that the submodel feature sets were trained using Sec. F.1's overlapping feature set formulation with the corresponding spread degree ($\phi$) specified above each column. The configuration with the best mean certified accuracy is shown in **bold**. Results averaged over 10 trials.

| Cert. Robust. | Random | Overlapping |
|---|---|---|
| | | $\phi = 3$ |
| 1 | **61.9 ± 1.4** | 61.0 ± 0.9 |
| 3 | 52.7 ± 1.4 | **53.3 ± 0.9** |
| 6 | 36.8 ± 1.6 | **37.6 ± 1.0** |
| 9 | **18.3 ± 2.4** | 17.7 ± 1.9 |
| 12 | 3.0 ± 1.7 | **3.1 ± 1.1** |

# G  Evaluation Setup

This section details the evaluation setup used in the experiments in Sections 6, F, and H. Below, we provide our experiments' implementation details, dataset configurations, and hyperparameter settings. The evaluation setup details below apply irrespective of whether the decision function uses plurality voting or run-off.

Our source code can be downloaded from https://github.com/ZaydH/feature-partition. All experiments were implemented and tested in either Python 3.7.13 or 3.10.10. All neural networks were implemented in `PyTorch` version 1.12.0 [Pas+19]. LightGBM decision forests were trained using the official `lightgbm` Python module, version 3.3.3.99 [Ke+17].

## G.1  Hardware Setup

Experiments were performed on a desktop system with a single AMD 5950X 16-core CPU, 64GB of 3200MHz DDR4 RAM, and a single NVIDIA 3090 GPU.

## G.2  Baselines

To the extent of our knowledge, no existing method considers certified feature robustness guarantees (Def. 1). *Randomized ablation* – our most closely related method – considers $\ell_0$-norm certified robustness (Def. 2) [LF20b]. RA is a specialized form of randomized smoothing [CRK19; LXL23] targeted towards sparse evasion attacks. In terms of the state of the art, Jia et al. [Jia+22b] provide the tightest certification analysis for randomized ablation.

Recall that feature partition aggregation (FPA) provides strictly stronger certified guarantees than baseline RA. Put simply, FPA is solving a harder task than baseline randomized ablation. Therefore, when FPA achieves the same certified accuracy as the baseline, FPA is performing provably better, given FPA's stronger guarantees.

We also compare FPA to three certified patch defenses, namely: (de)randomized smoothing (DRS) [LF20a], patch interval bound propagation (IBP) [Chi+20b], and BAGCERT [MY21]. Note that BAGCERT's implementation is not open source, and Metzen and Yatsura [MY21] have indicated they do not plan to open source the code in the future.[11] As such, BAGCERT's results in the main paper were provided by Metzen and Yatsura via personal correspondence. BAGCERT's closed source code prohibited the collection of its certification time. Nonetheless, comparing FPA's certification time to that of BAGCERT provides only limited insight since FPA and BAGCERT certify very different types of guarantees.

## G.3  Datasets

Our empirical evaluation considers four datasets. First, MNIST [LeC+98] and CIFAR10 [KNH14] are vision classification datasets with 10 classes each.

Although all certified sparse defenses considered in this work are exclusively proposed in the context of classification, Hammoudeh and Lowd [HL23b] prove that certified regression *reduces* to voting-based certified classification. Hence, it is straightforward to transform FPA and randomized ablation into certified regression defenses. We reuse this reduction and evaluate two tabular regression datasets, Weather [Mal+21] and Ames [De 11].

For Weather, we follow Hammoudeh and Lowd's [HL23b] empirical evaluation, where the objective is to predict ground temperature within $\pm 3°$C using features that include the date, time of day, longitude, and latitude. Similarly, we follow Hammoudeh and Lowd's [HL23b]'s empirical evaluation for Ames, where the objective is to predict a property's sale price within $\pm 15\%$ of the actual price. Since ablated training requires a custom feature encoding to differentiate ablated and non-ablated features, min-max scaling was applied to both datasets' features for RA to normalize all feature values to the range $[0, 1]$.

---

[11]The author's comments regarding open-sourcing their code can be found on BAGCERT's OpenReview page (https://openreview.net/forum?id=hr-3PMvDpil).

We chose these two regression datasets as a stand-in for vertically partitioned data, which are commonly tabular and particularly vulnerable to sparse backdoor and evasion attacks.

Table 14 provides basic information about the four datasets, including their sizes and feature dimension. Table 15 provides summary statistics for the regression datasets' test target-value (i.e., $y$) distribution.

Table 14: Evaluation dataset information

| Dataset | # Classes | # Feats | # Train | # Test |
|---|---|---|---|---|
| CIFAR10 | 10 | 1,024 | 50,000 | 10,000 |
| MNIST | 10 | 784 | 60,000 | 10,000 |
| Weather | N/A | 128 | 3,012,917 | 531,720 |
| Ames | N/A | 352 | 2,637 | 293 |

Table 15: **Target Value Test Distribution Statistics**: Mean ($\bar{y}$), standard deviation ($\sigma_y$), minimum value ($y_{\min}$) and maximum value ($y_{\max}$) for the test instances' target $y$ value for regression datasets Weather and Ames.

| | $\bar{y}$ | $\sigma_y$ | $y_{\min}$ | $y_{\max}$ |
|---|---|---|---|---|
| Weather | 14.9°C | 10.3°C | −44.0°C | 54.0°C |
| Ames | $184k | $83.4k | $12.8k | $585k |

Our source code automatically downloads all necessary dataset files.

### G.4 Network Architectures

Table 16 details the CIFAR10 neural network architecture. Specifically, we follow previous work on CIFAR10 data poisoning [HL22] and use Page's [Pag20] ResNet9 architecture. ResNet9 is ideal for our experiments since it is very fast to train, as ranked on DAWNBench [Col+17]. ResNet9's fast training significantly reduces the overhead of training $T$ submodels for FPA.

We directly adapt Page's [Pag20] published implementation[12] including the use of ghost batch normalization [SD20] and the CELU activation function with $\alpha = 0.075$ [Bar17].

Three forms of data augmentation were also used in line with Page's [Pag20] implementation. First, a random crop with four pixels of padding was performed. Next, the image was flipped horizontally with a 50% probability. Finally, a random $8 \times 8$ pixel portion of the image was randomly erased. Note that these transformations were performed *after* the pixels were disabled in the image, meaning these transformations do not result in a network seeing additional pixel information.

In a separate paper, Levine and Feizi [LF21] propose *deep partition aggregation* (DPA), a certified defense against poisoning attacks. Here, we follow Levine and Feizi's [LF21] public implementation[13] and use the Network-in-Network (NiN) architecture [LCY14] when evaluating our method on MNIST. Table 17 visualizes the MNIST NiN architecture.

### G.5 Hyperparameters

For simplicity, FPA used the same hyperparameter settings for a given dataset irrespective of $T$. Therefore, FPA's results could be further improved in practice by tuning the hyperparameter settings to optimize the ensemble's performance for a specific submodel count.

Table 18 details the CIFAR10 and MNIST hyperparameter settings for feature partition aggregation.

---
[12]Source code: https://github.com/davidcpage/cifar10-fast.
[13]Source code: https://github.com/alevine0/DPA.

Table 16: ResNet9 neural network architecture

|  | Layer | | | | |
|---|---|---|---|---|---|
|  | Conv1 | In=3 | Out=64 | Kernel=3×3 | Pad=1 |
|  | BatchNorm2D | Out=64 | | | |
|  | CELU | | | | |
|  | Conv2 | In=64 | Out=128 | Kernel=3×3 | Pad=1 |
|  | BatchNorm2D | Out=128 | | | |
|  | CELU | | | | |
|  | MaxPool2D | 2×2 | | | |
| ↑ ResNet1 ↓ | ConvA | In=128 | Out=128 | Kernel=3×3 | Pad=1 |
|  | BatchNorm2D | Out=128 | | | |
|  | CELU | | | | |
|  | ConvB | In=128 | Out=128 | Kernel=3×3 | Pad=1 |
|  | BatchNorm2D | Out=128 | | | |
|  | CELU | | | | |
|  | Conv3 | In=128 | Out=256 | Kernel=3×3 | Pad=1 |
|  | BatchNorm2D | Out=256 | | | |
|  | CELU | | | | |
|  | MaxPool2D | 2×2 | | | |
|  | Conv4 | In=256 | Out=512 | Kernel=3×3 | Pad=1 |
|  | BatchNorm2D | Out=512 | | | |
|  | CELU | | | | |
|  | MaxPool2D | 2×2 | | | |
| ↑ ResNet2 ↓ | ConvA | In=512 | Out=512 | Kernel=3×3 | Pad=1 |
|  | BatchNorm2D | Out=512 | | | |
|  | CELU | | | | |
|  | ConvB | In=512 | Out=512 | Kernel=3×3 | Pad=1 |
|  | BatchNorm2D | Out=512 | | | |
|  | CELU | | | | |
|  | MaxPool2D | 4×4 | | | |
|  | Linear | Out=10 | | | |

For CIFAR10 and MNIST, we directly used Levine and Feizi's [LF20b] published randomized ablation training source code, which includes pre-specified hyperparameter settings for the learning rate, weight decay, and optimizer hyperparameters.

Recall from Sec. 6 that for the Weather and Ames datasets, FPA's submodels are LightGBM [Ke+17] gradient-boosted decision tree (GBDT) regressors. Table 19 details FPA's LightGBM hyperparameter settings. For a more direct comparison with randomized ablation which cannot use a GBDT, we also evaluated FPA with linear submodels. FPA's linear submodel hyperparameter settings for the regression datasets are in Table 20.

Levine and Feizi [LF20b] only evaluate classification datasets in their original paper. As such, there are no existing hyperparameter settings for randomized ablation on Weather and Ames. We manually tuned randomized ablation's learning rate for the regression datasets considering all values in the set $\{10^{-2}, 10^{-3}, 10^{-4}\}$. We also tested numerous different settings for the number of training epochs. To ensure a strong baseline, we report the best performing randomized ablation hyperparameter settings.

Recall from Sec. 3 that randomized ablation only provides probabilistic guarantees. By contrast, feature partition aggregation provides deterministic guarantees. To facilitate a more direct comparison between certified feature and $\ell_0$-norm guarantees, $\alpha = 0.0001$ in all experiments.

## G.6 Overview of the Certified Regression to Certified Classification Reduction

Hammoudeh and Lowd [HL23b] provide a reduction from certified regression to (voting-based) certified classification. Hammoudeh and Lowd [HL23b] frame this reduction primarily in the context of poisoning attacks, but the reduction generalizes to other voting-based certified classifiers. For full details on the reduction from certified regression to certified classification, we direct the reader to Hammoudeh and Lowd's [HL23b] original paper. We briefly summarize the reduction below.

Consider a multiset of real-valued "votes" $\mathcal{V} \in \mathbb{R}^T$, where Hammoudeh and Lowd [HL23b] assume for simplicity that $T$ is odd. These "votes" could be generated from an ensemble of independent submodels in the case of deep partition aggregation [LF21] and FPA. These votes could also be generated from a smoothing-based classifier such as randomized ablation. Regardless, for voting-based real-valued regression, model $f$'s decision function for arbitrary instance $\mathbf{x} \in \mathcal{X}$ is

$$f(\mathbf{x}) \coloneqq \operatorname{med} \mathcal{V}, \tag{32}$$

where med denotes the median operator.

Let $y \in \mathbb{R}$ denote the true *target* value for $\mathbf{x}$ and let $\xi_l, \xi_u \in \mathbb{R}_{\geq 0}$ be arbitrary non-negative constants. Hammoudeh and Lowd's [HL23b] formulation seeks to certify the pointwise robustness of $\xi_l \leq f(\mathbf{x}) \leq \xi_u$.[14] Below, we discuss certifying a one-sided upper bound $f(\mathbf{x}) \leq \xi_u$. As Hammoudeh and Lowd [HL23b] explain, certifying a two-sided bound is equivalent to taking the minimum robustness of the one-sided lower and upper bounds.

Consider binarizing multiset $\mathcal{V}$ as $\mathcal{V}_{\pm 1} \coloneqq \{\operatorname{sgn}(v - \xi_u) : v \in \mathcal{V}\}$, where $\operatorname{sgn}(\cdot)$ is the signum function. Intuitively, our goal is to transform each real-valued instance in the multiset into a binary label, either $-1$ or $+1$. Certified defenses such as deep partition aggregation [LF20b], our sparse defense feature partition aggregation (FPA), and randomized ablation (RA) turn a multiset of votes into certified guarantees. Hammoudeh and Lowd's [HL23b] key insight is that the median and plurality labels of a binary multiset (e.g., $\mathcal{V}_{\pm 1}$) with odd-valued cardinality are always equal. In short, certifying when a multiset's median exceeds some threshold (e.g., $\xi_u$) is equivalent to certifying the perturbation of the plurality label of binarized multiset $\mathcal{V}_{\pm 1}$ [HL23b, Lem 6]. Hammoudeh and Lowd's [HL23b] reduction allows us to change the underlying prediction mechanism from a classifier to a regressor and directly reuse a voting-based certified classifier's robustness certification mechanism.

Hence, while our feature partition aggregation (FPA) and baseline randomized ablation are formulated as certified classifiers, both can be reformulated as certified regressors using the reduction of Hammoudeh and Lowd [HL23b]. In practice, the primary change made to both defenses is that the underlying learner(s) predict a real value instead of a label.

For regression, certified accuracy denotes that the model prediction satisfies $\xi_l \leq f(\mathbf{x}) \leq \xi_u$, even after $r$ feature perturbations.

For smoothing-based methods like randomized ablation, the reduction of Hammoudeh and Lowd [HL23b] is functionally very similar to Chiang et al.'s [Chi+20a] *median smoothing*. The two methods have slightly different formulations depending on the specification of the bounds.

---

[14]We use the exact same definitions for $\xi_l$ and $\xi_u$ as Hammoudeh and Lowd [HL23b]. Specifically for the Weather dataset, our experiments used $\xi_l = y - 3°C$ and $\xi_u = y + 3°C$. For the Ames dataset, our experiments used $\xi_l = y - 15\% y$ and $\xi_u = y + 15\% y$.

Table 17: Network-in-Network neural network architecture

| | | | | | |
|---|---|---|---|---|---|
| | Conv1 | In=3 | Out=192 | Kernel=5×5 | Pad=2 |
| | BatchNorm2D | Out=192 | | | |
| | ReLU | | | | |
| | Conv2 | In=192 | Out=160 | Kernel=1×1 | Pad=1 |
| Block 1 | BatchNorm2D | Out=160 | | | |
| | ReLU | | | | |
| | Conv3 | In=160 | Out=96 | Kernel=1×1 | Pad=1 |
| | BatchNorm2D | Out=96 | | | |
| | ReLU | | | | |
| | MaxPool2D | 3×3 | | | |
| | Conv1 | In=96 | Out=192 | Kernel=5×5 | Pad=2 |
| | BatchNorm2D | Out=192 | | | |
| | ReLU | | | | |
| | Conv2 | In=192 | Out=192 | Kernel=1×1 | Pad=1 |
| Block 2 | BatchNorm2D | Out=192 | | | |
| | ReLU | | | | |
| | Conv3 | In=192 | Out=192 | Kernel=1×1 | Pad=1 |
| | BatchNorm2D | Out=192 | | | |
| | ReLU | | | | |
| | AvgPool2D | 3×3 | | | |
| | Conv1 | In=192 | Out=192 | Kernel=3×3 | Pad=1 |
| | BatchNorm2D | Out=192 | | | |
| | ReLU | | | | |
| | Conv2 | In=192 | Out=192 | Kernel=1×1 | Pad=1 |
| Block 3 | BatchNorm2D | Out=192 | | | |
| | ReLU | | | | |
| | Conv3 | In=192 | Out=192 | Kernel=1×1 | Pad=1 |
| | BatchNorm2D | Out=192 | | | |
| | ReLU | | | | |
| | GlobalAvgPool2D | Out=192 | | | |
| | Linear | Out=10 | | | |

Table 18: FPA's neural network training hyperparameters

|  | CIFAR10 | MNIST |
|---|---|---|
| Data Augmentation? | ✓ | |
| Validation Split | N/A | 5% |
| Optimizer | SGD | AdamW |
| Batch Size | 512 | 128 |
| # Epochs | 80 | 25 |
| Learning Rate (Peak) | $1 \cdot 10^{-3}$ | $3.16 \cdot 10^{-4}$ |
| Learning Rate Scheduler | One cycle | Cosine |
| Weight Decay ($L_2$) | $1 \cdot 10^{-1}$ | $1 \cdot 10^{-3}$ |

Table 19: Regression datasets LightGBM submodel training hyperparameters

|  | Weather | Ames |
|---|---|---|
| Boosting Type | GBDT | GBDT |
| # Estimators | 500 | 1,000 |
| Max. Depth | 10 | 6 |
| Min. Child Samples | 20 | 5 |
| Max. # Leaves | 127 | 127 |
| $L_1$ Regularizer | 0 | $1 \cdot 10^{-3}$ |
| $L_2$ Regularizer | 0 | $1 \cdot 10^2$ |
| Objective | Huber | MAE |
| Learning Rate | 0.5 | $1 \cdot 10^2$ |
| Subsampling | 0.9 | 0.9 |

Table 20: Regression datasets linear submodel training hyperparameters

|  | Weather | Ames |
|---|---|---|
| $L_1$ Regularizer | $3.16 \cdot 10^{-3}$ | $4.15 \cdot 10^{-5}$ |
| Max. # Iterations | $1 \cdot 10^4$ | $1 \cdot 10^6$ |
| Tolerance | $1 \cdot 10^{-3}$ | $1 \cdot 10^{-8}$ |

# H  Additional Experiments

Limited space prevents us from including all experimental results in the main paper. We provide additional results below.

## H.1  Non-Robust Accuracy

Table 21 provides the non-robust (i.e., uncertified) accuracy when training a single model ($T = 1$) on each of Sec. 6's four datasets. The non-robust accuracy provides an upper-bound reference for the maximum achievable accuracy given the training set and the model architectures we used.

For regression, the "non-robust accuracy" denotes the single model's prediction satisfies the error bounds, i.e., $\xi_l \leq f(\mathbf{x}) \leq \xi_u$. Given arbitrary instance $(\mathbf{x}, y)$, we follow Hammoudeh and Lowd [HL23b] and use for Weather $\xi_l = y - 3°C$ and $\xi_u = y + 3°C$ as well as for Ames $\xi_l = y - 15\%y$ and $\xi_u = y + 15\%y$.

Table 21: **Non-Robust Accuracy**: Prediction accuracy when training a single model on all model features, i.e., $T = 1$. These values represent an upper bound on the potential accuracy of our method given the training set, model architecture, and hyperparameters.

| Dataset | Accuracy |
|---|---|
| CIFAR10 | 95.40% |
| MNIST | 99.57% |
| Weather | 92.61% |
| Ames | 88.05% |

## H.2 Detailed Median Certified Robustness Results

In Section 6.2 of the main paper, Tables 1 and 2 summarize the median certified robustness and classification accuracies of feature partition aggregation (FPA) and baseline randomized ablation [LF20b; Jia+22b]. In the tables, "(LF'20b)" denotes Levine and Feizi's [LF20b] original version of RA, and "(Jia'22b)" denotes Jia et al.'s [Jia+22b] improved RA; "Plural" denotes FPA using plurality voting as the decision function (Sec. 4.1) while "Run-Off" denotes FPA with Sec. 4.2's run-off elections.

Recall that FPA's primary hyperparameter is $T$ – the number of ensemble submodels. RA's primary hyperparameter is $e$ – the number of kept (unchanged) pixels in each ablated input. $T$ and $e$ control the corresponding method's accuracy-robustness trade-off where smaller $T$ and larger $e$ entail better accuracy. As a rule of thumb, the fairest comparison across methods sets $T \approx \frac{d}{e}$, since this relationship entails that each FPA and RA prediction uses approximately the same number of features from instance $\mathbf{x}$.

This section explores the relationship between each method's hyperparameter settings and the corresponding median robustness and classification accuracy. Each dataset's results are split into separate tables similar to Levine and Feizi's [LF20b, Tables 1 and 2] presentation in the original RA paper.

For CIFAR10 and MNIST, FPA uses deterministic partitioning. Specifically, we use a striding strategy as Section 5.1 details. Depending on the image dimensions, some stride lengths are substantially worse than others, leading to non-monotonic changes in median robustness as a function of $T$. Tables 22 and 23 do not report the particularly poor choices of $T$ that severely degrade median robustness, e.g., when $T$ is evenly divisible by the image width.

Below, any misclassified prediction is assigned robustness of $-\infty$, meaning the median certified robustness can in some cases be negative.

Table 22: **CIFAR10 Detailed Results**: Classification accuracy (%) and median certified robustness (larger is better) for the CIFAR10 [KNH14] dataset ($d = 1024$) for our certified sparse defense, feature partition aggregation (FPA), and baseline randomized ablation (RA) across various hyperparameter settings. Each certification method's hyperparameter setting with the best median robustness is shown in **bold**. The best overall median robustness is shown in **blue**.

(a) Feature Partition Aggregation (FPA – Ours)

| $T$ | Plural | | Run-Off | |
|---|---|---|---|---|
| | Acc. (%) | $r_{\text{med}}$ | Acc. (%) | $r_{\text{med}}$ |
| 5 | 91.46 | 2 | 91.77 | 2 |
| 10 | 86.09 | 4 | 86.20 | 4 |
| 20 | 81.38 | 7 | 81.40 | 7 |
| 25 | 78.65 | 8 | 78.58 | 8 |
| 40 | 74.74 | 9 | 74.95 | 10 |
| 55 | 70.44 | 10 | 70.34 | 11 |
| 70 | 67.46 | 9 | 67.47 | 11 |
| 85 | 66.24 | 10 | 66.61 | 12 |
| 105 | 63.55 | 10 | 63.61 | 12 |
| 115 | **62.39** | **11** | **62.35** | **13** |
| 140 | 60.35 | 10 | 60.57 | 12 |
| 165 | 57.91 | 8 | 58.48 | 10 |
| 185 | 56.08 | 7 | 56.39 | 9 |
| 200 | 55.80 | 7 | 56.43 | 9 |
| 225 | 56.27 | 6 | 56.56 | 8 |
| 250 | 53.30 | 4 | 53.46 | 5 |

(b) Randomized Ablation (RA – Baseline)

| $e$ | [LF20b] | | [Jia+22b] | |
|---|---|---|---|---|
| | Acc. (%) | $\rho_{\text{med}}$ | Acc. (%) | $\rho_{\text{med}}$ |
| 250 | 88.77 | 2 | 88.56 | 2 |
| 225 | 88.05 | 2 | 87.90 | 2 |
| 200 | 86.76 | 3 | 86.54 | 3 |
| 175 | 86.16 | 3 | 85.94 | 3 |
| 150 | 84.23 | 4 | 84.08 | 4 |
| 125 | 82.66 | 5 | 82.49 | 5 |
| 100 | 80.43 | 6 | 80.05 | 6 |
| 75 | **78.48** | **7** | 78.11 | 7 |
| 50 | 73.26 | 7 | 72.79 | 8 |
| 35 | 70.34 | 7 | 69.72 | 9 |
| 30 | 69.62 | 7 | 69.01 | 9 |
| 25 | 68.81 | 6 | 68.08 | 9 |
| 20 | 67.01 | 5 | 66.15 | 9 |
| 15 | 65.68 | 3 | **64.74** | **10** |
| 12 | 63.93 | 0 | 62.91 | 10 |
| 10 | 62.73 | 0 | 61.71 | 10 |
| 8 | 60.24 | 0 | 59.12 | 9 |
| 7 | 59.08 | 0 | 57.83 | 8 |
| 5 | 53.20 | 0 | 51.84 | 3 |

Table 23: **MNIST Detailed Results**: Classification accuracy (%) and median certified robustness (larger is better) for the MNIST [LeC+98] dataset ($d = 784$) for our certified sparse defense, feature partition aggregation (FPA), and baseline randomized ablation (RA) across various hyperparameter settings. Each certification method's hyperparameter setting with the best median robustness is shown in **bold**. The best overall median robustness is shown in **blue**.

(a) Feature Partition Aggregation (FPA – Ours)

| $T$ | Plural Acc. (%) | $r_{\text{med}}$ | Run-Off Acc. (%) | $r_{\text{med}}$ |
|---|---|---|---|---|
| 5 | 99.50 | 2 | 99.51 | 2 |
| 10 | 98.64 | 4 | 98.67 | 4 |
| 15 | 96.82 | 7 | 97.02 | 7 |
| 20 | 96.36 | 8 | 96.53 | 8 |
| 25 | **95.77** | **9** | 96.06 | 10 |
| 35 | 91.70 | 9 | 93.05 | 11 |
| 40 | 89.37 | 9 | 91.32 | 11 |
| 50 | 84.54 | 8 | 88.46 | 11 |
| 60 | 83.54 | 9 | **87.22** | **12** |
| 70 | 79.71 | 8 | 85.87 | 11 |
| 80 | 71.29 | 6 | 79.05 | 9 |
| 90 | 69.94 | 6 | 79.25 | 9 |
| 105 | 62.53 | 4 | 74.45 | 8 |
| 120 | 63.03 | 3 | 74.09 | 7 |
| 130 | 57.48 | 2 | 69.93 | 7 |
| 150 | 52.51 | 0 | 67.30 | 5 |

(b) Randomized Ablation (RA – Baseline)

| $e$ | [LF20b] Acc. (%) | $\rho_{\text{med}}$ | [Jia+22b] Acc. (%) | $\rho_{\text{med}}$ |
|---|---|---|---|---|
| 100 | 98.78 | 4 | 98.75 | 4 |
| 95 | 98.75 | 5 | 98.72 | 5 |
| 90 | 98.62 | 5 | 98.56 | 5 |
| 85 | 98.60 | 5 | 98.52 | 5 |
| 80 | 98.46 | 6 | 98.40 | 6 |
| 75 | 98.35 | 6 | 98.27 | 6 |
| 70 | 98.14 | 6 | 98.07 | 6 |
| 65 | 98.04 | 7 | 97.98 | 7 |
| 60 | 97.85 | 7 | 97.78 | 7 |
| 55 | 97.58 | 7 | 97.39 | 8 |
| 50 | 97.26 | 7 | 97.07 | 8 |
| 45 | **96.88** | **8** | 96.68 | 8 |
| 40 | 96.42 | 8 | 96.13 | 9 |
| 35 | 95.69 | 8 | 95.32 | 9 |
| 30 | 94.87 | 7 | 94.47 | 9 |
| 25 | 93.55 | 6 | **93.09** | **10** |
| 20 | 90.99 | 3 | 90.07 | 9 |
| 15 | 86.71 | 0 | 85.24 | 8 |
| 10 | 76.78 | 0 | 74.69 | 6 |
| 5 | 35.54 | $-\infty$ | 32.89 | $-\infty$ |

Table 24: **Weather Detailed Results**: Classification accuracy (%) and median certified robustness (larger is better) for the Weather [Mal+21] dataset ($d = 128$) for our certified sparse defense, feature partition aggregation (FPA), and baseline randomized ablation (RA) across various hyperparameter settings. FPA considers only plurality voting-based certification (Sec. 4.1) since Hammoudeh and Lowd's [HL23b] reduction is from certified regression to certified *binary* classification (see Sec. G.6 for details). FPA results are reported using both GBDTs [Ke+17] and linear submodels. Median robustness "$-\infty$" denotes that the classification accuracy was less than 50%. Each approach's hyperparameter setting with the best median robustness is shown in **bold**. The best overall median robustness is shown in **blue**. *Takeaway*: FPA with both GBDT and linear submodels achieved better median robustness than baseline RA.

(a) Feature Partition Aggregation (FPA – Ours)

| $T$ | LightGBM | | Linear | |
|---|---|---|---|---|
| | Acc. (%) | $r_{\text{med}}$ | Acc. (%) | $r_{\text{med}}$ |
| 1 | 92.70 | 0 | 86.05 | 0 |
| 5 | 85.29 | 2 | 83.34 | 2 |
| 11 | 82.48 | 3 | 79.55 | 2 |
| 15 | 81.09 | 3 | **76.15** | **3** |
| 21 | **76.10** | **4** | 67.09 | 2 |
| 25 | 71.40 | 3 | 64.77 | 2 |
| 31 | 67.06 | 3 | 58.71 | 2 |
| 35 | 62.56 | 3 | 55.95 | 1 |
| 41 | 60.19 | 2 | 51.57 | 0 |
| 51 | 55.34 | 1 | 45.84 | $-\infty$ |
| 75 | 42.20 | $-\infty$ | 26.93 | $-\infty$ |
| 101 | 28.67 | $-\infty$ | 21.26 | $-\infty$ |

(b) Randomized Ablation (RA – Baseline)

| $e$ | [LF20b] | | [Jia+22b] | |
|---|---|---|---|---|
| | Acc. (%) | $\rho_{\text{med}}$ | Acc. (%) | $\rho_{\text{med}}$ |
| 65 | **80.70** | **0** | 78.63 | 0 |
| 60 | 80.33 | 0 | 78.01 | 0 |
| 55 | 79.52 | 0 | 77.05 | 0 |
| 50 | 78.62 | 0 | 76.59 | 0 |
| 45 | 77.20 | 0 | **75.19** | **1** |
| 40 | 76.56 | 0 | 74.82 | 1 |
| 35 | 74.76 | 0 | 73.22 | 1 |
| 30 | 72.04 | 0 | 70.74 | 1 |
| 25 | 69.77 | 0 | 68.72 | 1 |
| 20 | 66.94 | 0 | 65.87 | 1 |
| 16 | 63.89 | 0 | 63.10 | 1 |
| 12 | 58.59 | 0 | 57.74 | 1 |
| 8 | 53.44 | 0 | 52.82 | 0 |
| 6 | 47.94 | $-\infty$ | 47.25 | $-\infty$ |
| 4 | 40.70 | $-\infty$ | 39.91 | $-\infty$ |

Table 25: **Ames Detailed Results**: Classification accuracy (%) and median certified robustness (larger is better) for the Ames [De 11] dataset ($d = 352$) for our certified sparse defense, feature partition aggregation (FPA), and baseline randomized ablation (RA) across various hyperparameter settings. FPA considers only plurality voting-based certification (Sec. 4.1) since Hammoudeh and Lowd's [HL23b] reduction is from certified regression to certified *binary* classification (see Sec. G.6 for details). FPA results are reported using both GBDTs [Ke+17] and linear submodels. Median robustness "$-\infty$" denotes that the classification accuracy was less than 50%. Each approach's hyperparameter setting with the best median robustness is shown in **bold**. The best overall median robustness is shown in blue. *Takeaway*: FPA with both GBDT and linear submodels achieved better median robustness than baseline RA.

(a) Feature Partition Aggregation (FPA – Ours)

| $T$ | LightGBM Acc. (%) | LightGBM $r_{\text{med}}$ | Linear Acc. (%) | Linear $r_{\text{med}}$ |
|---|---|---|---|---|
| 1 | 88.05 | 0 | 89.25 | 0 |
| 5 | 84.64 | 1 | 82.08 | 1 |
| 11 | 78.50 | 2 | 74.40 | 1 |
| 15 | 73.04 | 2 | **66.55** | **2** |
| 21 | **65.53** | **3** | 61.60 | 2 |
| 25 | 61.77 | 2 | 57.34 | 1 |
| 31 | 57.68 | 2 | 53.58 | 0 |
| 35 | 55.97 | 1 | 50.34 | 0 |
| 41 | 52.90 | 1 | 46.42 | $-\infty$ |
| 51 | 47.10 | $-\infty$ | 40.10 | $-\infty$ |
| 75 | 36.86 | $-\infty$ | 35.15 | $-\infty$ |

(b) Randomized Ablation (RA – Baseline)

| $e$ | [LF20b] Acc. (%) | [LF20b] $\rho_{\text{med}}$ | [Jia+22b] Acc. (%) | [Jia+22b] $\rho_{\text{med}}$ |
|---|---|---|---|---|
| 70 | 68.60 | 0 | 66.89 | 0 |
| 60 | 68.94 | 0 | **67.24** | **1** |
| 50 | **67.58** | **1** | 66.89 | 1 |
| 40 | 61.77 | 1 | 61.77 | 1 |
| 35 | 61.09 | 0 | 60.07 | 1 |
| 30 | 57.68 | 0 | 57.00 | 1 |
| 25 | 53.58 | 0 | 52.56 | 1 |
| 20 | 51.54 | 0 | 49.49 | $-\infty$ |
| 15 | 45.05 | $-\infty$ | 44.37 | $-\infty$ |
| 10 | 37.20 | $-\infty$ | 37.54 | $-\infty$ |
| 5 | 33.79 | $-\infty$ | 33.79 | $-\infty$ |

## H.3 Feature Partition Aggregation and Randomized Ablation Certified Accuracy Comparison

Levine and Feizi [LF20b] use median certified robustness and classification accuracy as the two primary metrics by which they compare RA against previous work. In this section, we present an alternative evaluation strategy comparing the methods' certified accuracy across a range of robustness levels.

Specifically, we consider the same four datasets from Section 6, namely classification datasets CIFAR10 [KNH14] and MNIST [LeC+98] as well as regression datasets Weather [Mal+21] and Ames [De 11]. Like in Section 6, we report FPA's performance using both the plurality-voting and run-off decision functions for classification and only plurality voting for regression. For baseline randomized ablation (RA), we again report the performance of Levine and Feizi's [LF20b] original version of RA as well as the improved version by Jia et al. [Jia+22b].

This section also compares FPA and RA against a *naive baseline* that is generally low accuracy but maximally robust. For classification, the naive baseline always predicts $f(\mathbf{x}) = 1$; for regression, the naive baseline always predicts the training set's median target value.

Recall that hyperparameters $T$ for FPA and $e$ for baseline randomized ablation control the corresponding method's accuracy versus robustness trade-off. Specifically, a smaller value of $T$ and a larger value of $e$ entails better accuracy. As a **rule of thumb**, the fairest comparison between FPA and RA is when $T \approx \frac{d}{e}$ as each FPA and RA prediction, in expectation, uses a comparable amount of information (i.e., number of features). For each dataset, we report each method's certified accuracy across 10 hyperparameter settings, roughly following the rule of thumb above. Section H.3.1 presents the experimental results in tabular form, and Section H.3.2 visualizes the methods' certified accuracy graphically.

### H.3.1 Numerical Comparison of Feature Partition Aggregation and Randomized Ablation

*Certified accuracy* w.r.t. $\psi \in \mathbb{N}$ quantifies the fraction of correctly-classified test instances with certified robustness at least $\psi$.

Tables 26, 27, 28, and 29 numerically display the certified accuracies for our certified feature defense, feature partition aggregation (FPA), and baseline randomized ablation (RA) for CIFAR10, MNIST, Weather, and Ames, respectively. For each dataset, the corresponding table lists the certified accuracy at 11 equally spaced certified robustness levels.

Recall that RA's $\ell_0$-norm robustness (Def. 2) is a strictly weaker guarantee than FPA's certified feature robustness (Def. 1). Put simply, a true direct comparison is not possible here since FPA provides stronger certified guarantees than the baseline. Despite that, FPA can achieve larger certified accuracies than the baseline while simultaneously providing stronger guarantees.

Table 26: **CIFAR10 Certified Accuracy Comparison**: CIFAR10 ($d = 1024$) certified accuracy (% – larger is better) for our certified feature defense, feature partition aggregation (FPA), and baseline randomized ablation (RA). "Plurality" denotes FPA with plurality voting as the decision function (Sec. 4.1) while "Run-Off" denotes FPA using run-off elections as the decision function (Sec. 4.2). "(LF'20b)" denotes Levine and Feizi's [LF20b] original version of randomized ablation while "(Jia'22b)" denotes Jia et al.'s [Jia+22b] improved version of RA that is tight for top-1 predictions. We also consider an additional naive baseline that always predicts $f(\mathbf{x}) = 1$, where, for correct predictions, the feature robustness equals $d$. For each certified robustness level, each method's best performing hyperparameter setting is shown in **bold** with the overall best performing method shown in blue. These numerical results are visualized graphically as envelope plots in Figure 2.

| Method | Cert. Alg. | Hyper. Setting | Certified Robustness | | | | | | | | | | |
|---|---|---|---|---|---|---|---|---|---|---|---|---|---|
| | | | 0 | 13 | 26 | 39 | 52 | 65 | 78 | 91 | 104 | 117 | 130 |
| Always $f(\mathbf{x}) = 1$ | | N/A | **10.00** | **10.00** | **10.00** | **10.00** | **10.00** | **10.00** | **10.00** | **10.00** | **10.00** | **10.00** | **10.00** |
| FPA ($T$) (ours) | Plurality | 5 | **91.46** | 0.00 | 0.00 | 0.00 | 0.00 | 0.00 | 0.00 | 0.00 | 0.00 | 0.00 | 0.00 |
| | | 25 | 78.65 | 0.00 | 0.00 | 0.00 | 0.00 | 0.00 | 0.00 | 0.00 | 0.00 | 0.00 | 0.00 |
| | | 35 | 69.62 | 36.35 | 0.00 | 0.00 | 0.00 | 0.00 | 0.00 | 0.00 | 0.00 | 0.00 | 0.00 |
| | | 55 | 70.44 | 44.06 | 10.06 | 0.00 | 0.00 | 0.00 | 0.00 | 0.00 | 0.00 | 0.00 | 0.00 |
| | | 85 | 66.24 | 46.67 | 26.87 | 7.77 | 0.00 | 0.00 | 0.00 | 0.00 | 0.00 | 0.00 | 0.00 |
| | | 115 | 62.39 | **47.74** | 33.48 | 19.67 | 6.97 | 0.00 | 0.00 | 0.00 | 0.00 | 0.00 | 0.00 |
| | | 160 | 60.94 | 42.27 | 27.77 | 16.95 | 9.00 | 3.89 | 0.52 | 0.00 | 0.00 | 0.00 | 0.00 |
| | | 250 | 53.30 | 43.98 | **35.63** | 28.37 | 21.54 | 15.57 | 10.91 | 7.04 | 4.02 | 1.62 | 0.00 |
| | | 500 | 43.79 | 38.75 | 33.63 | **28.86** | **24.65** | **20.86** | **17.56** | 14.32 | 11.56 | 9.38 | 7.66 |
| | | 1024 | 33.01 | 29.70 | 26.95 | 24.14 | 21.68 | 19.33 | 17.24 | **15.41** | **13.92** | **12.29** | **11.05** |
| | Run-Off | 5 | <span style="color:blue">**91.77**</span> | 0.00 | 0.00 | 0.00 | 0.00 | 0.00 | 0.00 | 0.00 | 0.00 | 0.00 | 0.00 |
| | | 25 | 78.58 | 0.00 | 0.00 | 0.00 | 0.00 | 0.00 | 0.00 | 0.00 | 0.00 | 0.00 | 0.00 |
| | | 35 | 69.92 | 37.45 | 0.00 | 0.00 | 0.00 | 0.00 | 0.00 | 0.00 | 0.00 | 0.00 | 0.00 |
| | | 55 | 70.34 | 46.71 | 11.18 | 0.00 | 0.00 | 0.00 | 0.00 | 0.00 | 0.00 | 0.00 | 0.00 |
| | | 85 | 66.61 | 49.26 | 30.25 | 8.91 | 0.00 | 0.00 | 0.00 | 0.00 | 0.00 | 0.00 | 0.00 |
| | | 115 | 62.35 | <span style="color:blue">**50.04**</span> | 36.76 | 22.64 | 8.21 | 0.00 | 0.00 | 0.00 | 0.00 | 0.00 | 0.00 |
| | | 160 | 61.34 | 45.54 | 32.71 | 21.16 | 11.96 | 5.06 | 0.56 | 0.00 | 0.00 | 0.00 | 0.00 |
| | | 250 | 53.46 | 45.48 | <span style="color:blue">**38.40**</span> | **31.70** | 25.24 | 19.02 | 13.48 | 8.94 | 4.99 | 1.88 | 0.00 |
| | | 500 | 44.58 | 39.58 | 35.25 | 31.17 | <span style="color:blue">**27.60**</span> | <span style="color:blue">**24.21**</span> | <span style="color:blue">**20.57**</span> | **17.62** | 14.74 | 12.33 | 10.25 |
| | | 1024 | 35.50 | 32.01 | 28.80 | 25.89 | 23.22 | 20.74 | 18.63 | 16.85 | <span style="color:blue">**15.20**</span> | <span style="color:blue">**13.80**</span> | <span style="color:blue">**12.57**</span> |
| RA ($e$) | (LF'20b) | 250 | **88.77** | 0.00 | 0.00 | 0.00 | 0.00 | 0.00 | 0.00 | 0.00 | 0.00 | 0.00 | 0.00 |
| | | 75 | 78.48 | 0.00 | 0.00 | 0.00 | 0.00 | 0.00 | 0.00 | 0.00 | 0.00 | 0.00 | 0.00 |
| | | 50 | 73.26 | 25.80 | 0.00 | 0.00 | 0.00 | 0.00 | 0.00 | 0.00 | 0.00 | 0.00 | 0.00 |
| | | 25 | 68.81 | **38.82** | 11.25 | 0.00 | 0.00 | 0.00 | 0.00 | 0.00 | 0.00 | 0.00 | 0.00 |
| | | 15 | 65.68 | 38.81 | 23.59 | 9.42 | 0.00 | 0.00 | 0.00 | 0.00 | 0.00 | 0.00 | 0.00 |
| | | 10 | 62.73 | 37.60 | **27.46** | 17.72 | 9.74 | 1.89 | 0.00 | 0.00 | 0.00 | 0.00 | 0.00 |
| | | 7 | 59.08 | 33.44 | 25.65 | **18.58** | 12.56 | 7.77 | 3.71 | 1.09 | 0.00 | 0.00 | 0.00 |
| | | 5 | 53.20 | 28.47 | 22.80 | 17.85 | **14.04** | **10.10** | 6.87 | 4.20 | 2.31 | 0.94 | 0.05 |
| | | 2 | 40.44 | 14.03 | 12.37 | 10.62 | 9.12 | 7.91 | **6.96** | **5.95** | **5.16** | **4.51** | **3.98** |
| | | 1 | 21.16 | 4.37 | 3.87 | 3.37 | 2.91 | 2.58 | 2.35 | 1.90 | 1.68 | 1.42 | 1.21 |
| | (Jia'22b) | 250 | **88.56** | 0.00 | 0.00 | 0.00 | 0.00 | 0.00 | 0.00 | 0.00 | 0.00 | 0.00 | 0.00 |
| | | 75 | 78.11 | 0.00 | 0.00 | 0.00 | 0.00 | 0.00 | 0.00 | 0.00 | 0.00 | 0.00 | 0.00 |
| | | 50 | 72.79 | 26.98 | 0.00 | 0.00 | 0.00 | 0.00 | 0.00 | 0.00 | 0.00 | 0.00 | 0.00 |
| | | 25 | 68.08 | 43.10 | 12.28 | 0.00 | 0.00 | 0.00 | 0.00 | 0.00 | 0.00 | 0.00 | 0.00 |
| | | 15 | 64.74 | 46.17 | 28.17 | 11.25 | 0.00 | 0.00 | 0.00 | 0.00 | 0.00 | 0.00 | 0.00 |
| | | 10 | 61.71 | **47.54** | 34.36 | 22.44 | 11.99 | 2.31 | 0.00 | 0.00 | 0.00 | 0.00 | 0.00 |
| | | 7 | 57.83 | 46.43 | **35.75** | 26.23 | 17.70 | 10.79 | 4.96 | 1.33 | 0.00 | 0.00 | 0.00 |
| | | 5 | 51.84 | 43.08 | 34.70 | **27.14** | 20.77 | 15.27 | 10.36 | 6.32 | 3.34 | 1.21 | 0.06 |
| | | 2 | 38.70 | 33.84 | 29.15 | 25.01 | **21.22** | **17.95** | **14.90** | **12.49** | **10.33** | **8.54** | **7.03** |
| | | 1 | 19.64 | 17.96 | 15.83 | 14.06 | 12.48 | 11.18 | 10.17 | 9.06 | 8.24 | 7.35 | 6.48 |

A35

Table 27: **MNIST Certified Accuracy Comparison**: MNIST ($d = 784$) certified accuracy (% – larger is better) for our certified feature defense, feature partition aggregation (FPA), and baseline randomized ablation (RA). "Plurality" denotes FPA with plurality voting as the decision function (Sec. 4.1) while "Run-Off" denotes FPA using run-off elections as the decision function (Sec. 4.2). "(LF'20b)" denotes Levine and Feizi's [LF20b] original version of randomized ablation while "(Jia'22b)" denotes Jia et al.'s [Jia+22b] improved version of RA that is tight for top-1 predictions. We also consider an additional naive baseline that always predicts $f(\mathbf{x}) = 1$, where, for correct predictions, the feature robustness equals $d$. For each certified robustness level, each method's best performing hyperparameter setting is shown in **bold** with the overall best performing method shown in blue. These numerical results are visualized graphically as envelope plots in Figure 2.

| Method | Cert. Alg. | Hyper. Setting | Certified Robustness | | | | | | | | | | |
|---|---|---|---|---|---|---|---|---|---|---|---|---|---|
| | | | 0 | 4 | 8 | 12 | 16 | 20 | 24 | 28 | 32 | 36 | 40 |
| Always $f(\mathbf{x}) = 1$ | | N/A | **11.35** | **11.35** | **11.35** | **11.35** | **11.35** | **11.35** | **11.35** | **11.35** | **11.35** | **11.35** | **11.35** |
| FPA ($T$) (ours) | Plurality | 5 | **99.50** | 0.00 | 0.00 | 0.00 | 0.00 | 0.00 | 0.00 | 0.00 | 0.00 | 0.00 | 0.00 |
| | | 10 | 98.64 | **87.16** | 0.00 | 0.00 | 0.00 | 0.00 | 0.00 | 0.00 | 0.00 | 0.00 | 0.00 |
| | | 25 | 95.77 | 86.48 | **66.42** | 20.05 | 0.00 | 0.00 | 0.00 | 0.00 | 0.00 | 0.00 | 0.00 |
| | | 35 | 91.70 | 79.49 | 59.53 | 35.95 | 13.18 | 0.00 | 0.00 | 0.00 | 0.00 | 0.00 | 0.00 |
| | | 60 | 83.54 | 70.30 | 54.72 | **39.10** | **26.26** | 16.08 | 7.95 | 1.78 | 0.00 | 0.00 | 0.00 |
| | | 75 | 74.99 | 61.44 | 47.75 | 34.97 | 25.34 | 17.90 | 12.43 | 8.11 | 3.89 | 0.42 | 0.00 |
| | | 90 | 69.94 | 57.11 | 43.89 | 33.01 | 24.52 | 17.89 | 12.99 | 9.16 | 6.24 | 3.22 | 0.71 |
| | | 105 | 62.53 | 50.33 | 39.10 | 29.27 | 22.13 | 16.52 | 13.04 | 10.51 | 8.42 | 6.61 | 4.63 |
| | | 130 | 57.48 | 46.68 | 36.45 | 28.38 | 22.70 | **18.52** | **15.23** | 12.54 | 10.45 | 8.38 | 6.30 |
| | | 240 | 28.13 | 24.67 | 21.81 | 19.57 | 17.63 | 16.33 | 15.16 | **14.40** | **13.79** | **13.00** | **12.30** |
| | Run-Off | 5 | 99.51 | 0.00 | 0.00 | 0.00 | 0.00 | 0.00 | 0.00 | 0.00 | 0.00 | 0.00 | 0.00 |
| | | 10 | 98.67 | 87.50 | 0.00 | 0.00 | 0.00 | 0.00 | 0.00 | 0.00 | 0.00 | 0.00 | 0.00 |
| | | 25 | 96.06 | 88.72 | 71.52 | 20.28 | 0.00 | 0.00 | 0.00 | 0.00 | 0.00 | 0.00 | 0.00 |
| | | 35 | 93.05 | 83.56 | 67.58 | 44.72 | 14.36 | 0.00 | 0.00 | 0.00 | 0.00 | 0.00 | 0.00 |
| | | 60 | 87.22 | 76.59 | 63.67 | 50.52 | 37.10 | 23.91 | 12.14 | 2.97 | 0.00 | 0.00 | 0.00 |
| | | 75 | 81.74 | 68.54 | 56.44 | 44.65 | 34.68 | 25.48 | 17.82 | 11.09 | 5.28 | 0.45 | 0.00 |
| | | 90 | 79.25 | 66.38 | 53.93 | 43.35 | 33.92 | 26.20 | 20.14 | 14.71 | 9.98 | 6.02 | 2.34 |
| | | 105 | 74.45 | 61.76 | 50.73 | 40.32 | 31.38 | 24.57 | 19.00 | 14.85 | 11.80 | 9.05 | 6.46 |
| | | 130 | 69.93 | 58.88 | 48.44 | 38.73 | 31.04 | 25.06 | 20.82 | 17.47 | 14.69 | 12.00 | 9.85 |
| | | 240 | 48.33 | 40.31 | 33.37 | 28.30 | 24.57 | 21.29 | 18.71 | 17.17 | 15.82 | 14.82 | 13.81 |
| RA ($e$) | (LF'20b) | 100 | **98.78** | 84.16 | 0.00 | 0.00 | 0.00 | 0.00 | 0.00 | 0.00 | 0.00 | 0.00 | 0.00 |
| | | 85 | 98.60 | **86.08** | 0.00 | 0.00 | 0.00 | 0.00 | 0.00 | 0.00 | 0.00 | 0.00 | 0.00 |
| | | 60 | 97.85 | 84.30 | 35.32 | 0.00 | 0.00 | 0.00 | 0.00 | 0.00 | 0.00 | 0.00 | 0.00 |
| | | 50 | 97.26 | 81.56 | 49.77 | 0.00 | 0.00 | 0.00 | 0.00 | 0.00 | 0.00 | 0.00 | 0.00 |
| | | 40 | 96.42 | 76.53 | **51.99** | 16.85 | 0.00 | 0.00 | 0.00 | 0.00 | 0.00 | 0.00 | 0.00 |
| | | 30 | 94.87 | 66.97 | 46.33 | **26.88** | 7.30 | 0.00 | 0.00 | 0.00 | 0.00 | 0.00 | 0.00 |
| | | 20 | 90.99 | 48.11 | 34.38 | 23.77 | **15.23** | 7.50 | 0.96 | 0.00 | 0.00 | 0.00 | 0.00 |
| | | 10 | 76.78 | 20.36 | 16.22 | 13.08 | 10.62 | 8.40 | 5.99 | 3.72 | 1.54 | 0.16 | 0.00 |
| | | 5 | 35.54 | 10.85 | 10.31 | 9.75 | 9.17 | 8.69 | 7.86 | 6.90 | 5.73 | 4.42 | 3.23 |
| | | 3 | 16.91 | 11.13 | 10.96 | 10.70 | 10.51 | **10.19** | **9.84** | **9.41** | **8.87** | **8.21** | **7.04** |
| | (Jia'22b) | 100 | **98.75** | 86.10 | 0.00 | 0.00 | 0.00 | 0.00 | 0.00 | 0.00 | 0.00 | 0.00 | 0.00 |
| | | 85 | 98.52 | 88.21 | 0.00 | 0.00 | 0.00 | 0.00 | 0.00 | 0.00 | 0.00 | 0.00 | 0.00 |
| | | 60 | 97.78 | **88.45** | 39.75 | 0.00 | 0.00 | 0.00 | 0.00 | 0.00 | 0.00 | 0.00 | 0.00 |
| | | 50 | 97.07 | 87.28 | 57.28 | 0.00 | 0.00 | 0.00 | 0.00 | 0.00 | 0.00 | 0.00 | 0.00 |
| | | 40 | 96.13 | 85.69 | **62.37** | 21.15 | 0.00 | 0.00 | 0.00 | 0.00 | 0.00 | 0.00 | 0.00 |
| | | 30 | 94.47 | 82.47 | 62.32 | 36.45 | 11.20 | 0.00 | 0.00 | 0.00 | 0.00 | 0.00 | 0.00 |
| | | 20 | 90.07 | 76.29 | 58.26 | **39.39** | **24.36** | 12.98 | 2.70 | 0.00 | 0.00 | 0.00 | 0.00 |
| | | 10 | 74.69 | 59.11 | 44.55 | 32.87 | 23.94 | **17.91** | **13.49** | 10.38 | 7.33 | 3.73 | 0.80 |
| | | 5 | 32.89 | 26.17 | 21.19 | 17.56 | 15.76 | 14.46 | 13.43 | **12.52** | **11.51** | 10.77 | 10.05 |
| | | 3 | 15.91 | 14.97 | 13.90 | 13.10 | 12.46 | 12.01 | 11.71 | 11.50 | 11.40 | **11.30** | **11.30** |

A36

Table 28: **Weather Certified Accuracy Comparison**: Weather [Mal+21] dataset ($d = 128$) certified accuracy (% – larger is better) for our certified feature defense, feature partition aggregation (FPA), and baseline randomized ablation (RA). "(LF'20b)" denotes Levine and Feizi's [LF20b] original version of randomized ablation while "(Jia'22b)" denotes Jia et al.'s [Jia+22b] improved version of RA that is tight for top-1 predictions. Hammoudeh and Lowd's [HL23b] reduction is from certified regression to certified binary classification. Run-off is identical to plurality voting under binary classification, so we report only the plurality voting results below. We also consider an additional naive baseline that always predicts the median training set target value (i.e., $f(\mathbf{x}) = \text{med}\{y_i\}_{i=1}^n$), where, for correct predictions, the feature robustness equals $d$. For each certified robustness level, each method's best performing hyperparameter setting is shown in **bold** with the overall best performing method shown in **blue**. These numerical results are visualized graphically as envelope plots in Figure 3.

| Method | Cert. Alg. | Hyper. Setting | Certified Robustness | | | | | | | | | | |
|---|---|---|---|---|---|---|---|---|---|---|---|---|---|
| | | | 0 | 1 | 2 | 3 | 4 | 5 | 6 | 7 | 8 | 9 | 10 |
| Always $f(\mathbf{x}) = \text{med}\{y_i\}_{i=1}^n$ | | N/A | **21.90** | **21.90** | **21.90** | **21.90** | **21.90** | **21.90** | **21.90** | **21.90** | **21.90** | **21.90** | **21.90** |
| FPA ($T$) (ours) | Plurality | 5 | **85.29** | **77.38** | 62.69 | 0.00 | 0.00 | 0.00 | 0.00 | 0.00 | 0.00 | 0.00 | 0.00 |
| | | 11 | 82.48 | 76.34 | 67.59 | 55.50 | 39.02 | 18.42 | 0.00 | 0.00 | 0.00 | 0.00 | 0.00 |
| | | 15 | 81.09 | 75.23 | **68.16** | **58.98** | 48.08 | 35.81 | 19.92 | 7.77 | 0.00 | 0.00 | 0.00 |
| | | 21 | 76.10 | 70.78 | 64.73 | 57.69 | **50.01** | 41.48 | 33.04 | 23.78 | 14.30 | 6.47 | 0.91 |
| | | 25 | 71.40 | 66.29 | 60.70 | 55.03 | 49.17 | 42.93 | 35.88 | 28.92 | 21.58 | 14.29 | 7.12 |
| | | 31 | 67.06 | 62.80 | 58.18 | 53.39 | 48.76 | **43.85** | 38.49 | 32.77 | 27.12 | 21.51 | 15.81 |
| | | 35 | 62.56 | 58.84 | 54.93 | 50.72 | 46.54 | 42.03 | 37.62 | 33.08 | 28.10 | 22.76 | 17.18 |
| | | 41 | 60.19 | 56.83 | 53.34 | 49.72 | 45.99 | 42.34 | 38.55 | 34.60 | 30.44 | 26.09 | 21.47 |
| | | 45 | 57.96 | 54.99 | 51.94 | 48.81 | 45.57 | 42.26 | **38.78** | **35.11** | **31.29** | **27.23** | **22.91** |
| | | 127 | 23.43 | 22.95 | 22.49 | 22.04 | 21.61 | 21.19 | 20.77 | 20.38 | 20.00 | 19.61 | 19.23 |
| RA ($e$) | (LF'20b) | 50 | **78.62** | 22.32 | 0.00 | 0.00 | 0.00 | 0.00 | 0.00 | 0.00 | 0.00 | 0.00 | 0.00 |
| | | 40 | 76.56 | 31.26 | 0.00 | 0.00 | 0.00 | 0.00 | 0.00 | 0.00 | 0.00 | 0.00 | 0.00 |
| | | 30 | 72.04 | 39.64 | 9.53 | 0.00 | 0.00 | 0.00 | 0.00 | 0.00 | 0.00 | 0.00 | 0.00 |
| | | 20 | 66.94 | 45.11 | 20.61 | 6.82 | 0.00 | 0.00 | 0.00 | 0.00 | 0.00 | 0.00 | 0.00 |
| | | 16 | 63.89 | **45.77** | 26.67 | 11.64 | 3.83 | 0.04 | 0.00 | 0.00 | 0.00 | 0.00 | 0.00 |
| | | 12 | 58.59 | 45.19 | 31.87 | 18.36 | 9.67 | 4.37 | 1.06 | 0.00 | 0.00 | 0.00 | 0.00 |
| | | 9 | 54.68 | 44.55 | **35.11** | 25.05 | 15.88 | 9.48 | 5.26 | 2.26 | 0.61 | 0.01 | 0.00 |
| | | 6 | 47.94 | 41.22 | 34.84 | **28.60** | 22.32 | 16.45 | 11.82 | 8.60 | 6.00 | 3.90 | 2.37 |
| | | 3 | 36.88 | 33.32 | 30.57 | 27.90 | **25.63** | **23.08** | **20.58** | 18.16 | 15.97 | 13.91 | 11.87 |
| | | 1 | 21.00 | 20.68 | 20.61 | 20.48 | 20.35 | 20.19 | 20.05 | **19.93** | **19.77** | **19.67** | **19.43** |
| | (Jia'22b) | 50 | **76.59** | 47.32 | 0.00 | 0.00 | 0.00 | 0.00 | 0.00 | 0.00 | 0.00 | 0.00 | 0.00 |
| | | 40 | 74.82 | 53.84 | 0.00 | 0.00 | 0.00 | 0.00 | 0.00 | 0.00 | 0.00 | 0.00 | 0.00 |
| | | 30 | 70.74 | 56.18 | 31.24 | 0.00 | 0.00 | 0.00 | 0.00 | 0.00 | 0.00 | 0.00 | 0.00 |
| | | 20 | 65.87 | **56.66** | 44.24 | 26.06 | 3.94 | 0.00 | 0.00 | 0.00 | 0.00 | 0.00 | 0.00 |
| | | 16 | 63.10 | 55.29 | **46.24** | 34.49 | 19.75 | 5.20 | 0.00 | 0.00 | 0.00 | 0.00 | 0.00 |
| | | 12 | 57.74 | 51.96 | 45.73 | 38.47 | 29.53 | 19.26 | 10.88 | 0.00 | 0.00 | 0.00 | 0.00 |
| | | 9 | 53.97 | 49.95 | 45.97 | **41.18** | 35.62 | 29.11 | 21.44 | 14.51 | 9.10 | 2.63 | 0.00 |
| | | 6 | 47.25 | 44.86 | 41.94 | 39.16 | **36.21** | **33.00** | **29.54** | 25.82 | 21.18 | 16.82 | 13.31 |
| | | 3 | 36.01 | 34.97 | 33.59 | 32.19 | 31.02 | 29.72 | 28.46 | **27.33** | **26.28** | **25.21** | **23.99** |
| | | 1 | 20.84 | 20.76 | 20.72 | 20.63 | 20.58 | 20.50 | 20.41 | 20.31 | 20.25 | 20.14 | 20.03 |

Table 29: **Ames Certified Accuracy Comparison**: Ames [De 11] dataset ($d = 352$) certified accuracy (% – larger is better) for our certified feature defense, feature partition aggregation (FPA), and baseline randomized ablation (RA). "(LF'20b)" denotes Levine and Feizi's [LF20b] original version of randomized ablation while "(Jia'22b)" denotes Jia et al.'s [Jia+22b] improved version of RA that is tight for top-1 predictions. Hammoudeh and Lowd's [HL23b] reduction is from certified regression to certified binary classification. Run-off is identical to plurality voting under binary classification, so we report only the plurality voting results below. We also consider an additional naive baseline that always predicts the median training set target value (i.e., $f(\mathbf{x}) = \mathrm{med}\{y_i\}_{i=1}^n$), where, for correct predictions, the feature robustness equals $d$. For each certified robustness level, each method's best performing hyperparameter setting is shown in **bold** with the overall best performing method shown in **blue**. These numerical results are visualized graphically as envelope plots in Figure 3.

| Method | Cert. Alg. | Hyper. Setting | Certified Robustness | | | | | | | | | | |
|---|---|---|---|---|---|---|---|---|---|---|---|---|---|
| | | | 0 | 1 | 2 | 3 | 4 | 5 | 6 | 7 | 8 | 9 | 10 |
| Always $f(\mathbf{x}) = \mathrm{med}\{y_i\}_{i=1}^n$ | N/A | | **31.40** | **31.40** | **31.40** | **31.40** | **31.40** | **31.40** | **31.40** | **31.40** | **31.40** | **31.40** | **31.40** |
| FPA ($T$) (ours) | Plurality | 5 | **84.64** | **72.01** | 39.93 | 0.00 | 0.00 | 0.00 | 0.00 | 0.00 | 0.00 | 0.00 | 0.00 |
| | | 11 | 78.50 | 70.99 | **58.70** | 40.96 | 22.53 | 5.12 | 0.00 | 0.00 | 0.00 | 0.00 | 0.00 |
| | | 21 | 65.53 | 60.41 | 54.95 | **50.17** | 41.64 | 32.42 | 22.87 | 12.63 | 5.46 | 1.37 | 0.00 |
| | | 25 | 61.77 | 58.36 | 54.27 | 49.83 | **43.69** | 35.84 | 28.67 | 20.82 | 12.63 | 6.14 | 2.39 |
| | | 31 | 57.68 | 54.95 | 51.54 | 48.12 | 42.66 | 37.20 | 32.08 | 26.28 | 20.82 | 15.02 | 10.24 |
| | | 35 | 55.97 | 52.56 | 48.81 | 45.73 | 42.32 | **38.23** | 33.79 | 29.01 | 24.57 | 19.45 | 14.68 |
| | | 41 | 52.90 | 50.51 | 47.10 | 43.34 | 40.96 | 37.20 | **34.47** | 31.06 | 27.65 | 24.23 | 20.82 |
| | | 51 | 47.10 | 44.37 | 41.98 | 39.25 | 37.88 | 35.49 | 34.13 | 32.08 | 30.03 | 28.33 | 26.28 |
| | | 65 | 41.64 | 39.25 | 37.88 | 37.20 | 36.01 | 34.47 | 33.45 | **32.42** | 31.40 | 30.38 | 29.69 |
| | | 101 | 33.45 | 33.11 | 32.76 | 32.76 | 32.42 | 32.08 | 32.08 | 31.74 | **31.74** | **31.74** | 31.40 |
| RA ($e$) | (LF'20b) | 60 | **68.94** | 43.34 | 11.95 | 0.00 | 0.00 | 0.00 | 0.00 | 0.00 | 0.00 | 0.00 | 0.00 |
| | | 50 | 67.58 | **52.56** | 32.08 | 7.85 | 0.00 | 0.00 | 0.00 | 0.00 | 0.00 | 0.00 | 0.00 |
| | | 40 | 61.77 | 50.17 | 38.23 | 18.09 | 4.10 | 0.00 | 0.00 | 0.00 | 0.00 | 0.00 | 0.00 |
| | | 35 | 61.09 | 49.49 | **39.93** | 20.48 | 10.24 | 1.71 | 0.00 | 0.00 | 0.00 | 0.00 | 0.00 |
| | | 30 | 57.68 | 48.46 | 39.59 | 26.96 | 16.38 | 5.46 | 0.00 | 0.00 | 0.00 | 0.00 | 0.00 |
| | | 25 | 53.58 | 47.78 | 38.91 | 27.65 | 20.82 | 15.02 | 4.10 | 0.34 | 0.00 | 0.00 | 0.00 |
| | | 20 | 51.54 | 43.34 | 38.23 | 32.76 | 26.28 | 20.48 | 15.02 | 7.85 | 2.39 | 0.00 | 0.00 |
| | | 15 | 45.05 | 39.25 | 36.18 | **34.81** | 29.69 | 27.99 | 23.21 | 19.45 | 13.99 | 9.90 | 5.80 |
| | | 10 | 37.20 | 36.18 | 35.15 | 33.11 | **32.76** | 31.40 | 28.67 | 26.62 | 25.26 | 24.57 | 22.87 |
| | | 5 | 33.79 | 33.11 | 32.76 | 32.08 | 32.08 | **32.08** | 31.74 | 31.40 | 31.06 | 30.38 | 30.38 |
| | (Jia'22b) | 60 | **67.24** | 59.73 | 46.76 | 13.99 | 0.00 | 0.00 | 0.00 | 0.00 | 0.00 | 0.00 | 0.00 |
| | | 50 | 66.89 | **59.73** | 48.81 | 31.40 | 7.17 | 0.00 | 0.00 | 0.00 | 0.00 | 0.00 | 0.00 |
| | | 40 | 61.77 | 55.63 | **49.49** | 38.57 | 25.60 | 6.48 | 0.00 | 0.00 | 0.00 | 0.00 | 0.00 |
| | | 35 | 60.07 | 52.90 | 48.12 | 38.91 | 31.06 | 16.38 | 2.39 | 0.00 | 0.00 | 0.00 | 0.00 |
| | | 30 | 57.00 | 51.88 | 47.10 | **41.30** | 34.81 | 26.96 | 15.36 | 2.39 | 0.00 | 0.00 | 0.00 |
| | | 25 | 52.56 | 50.17 | 45.39 | 40.27 | 35.84 | 31.06 | 24.91 | 17.06 | 6.48 | 0.34 | 0.00 |
| | | 20 | 49.49 | 45.73 | 44.03 | **41.30** | **37.54** | 33.79 | 30.38 | 25.94 | 22.53 | 13.99 | 6.83 |
| | | 15 | 44.37 | 42.32 | 40.96 | 39.93 | 35.84 | **35.49** | 32.76 | 30.72 | 27.65 | 24.91 | 22.18 |
| | | 10 | 37.54 | 36.52 | 35.84 | 33.79 | 33.79 | 33.45 | 32.42 | 31.06 | 30.38 | 29.35 | 29.01 |
| | | 5 | 33.79 | 33.45 | 33.45 | 33.11 | 33.11 | 33.11 | **32.76** | **32.76** | **32.42** | **32.08** | **32.08** |

### H.3.2 Graphical Comparison of Feature Partition Aggregation and Randomized Ablation

Recall that hyperparameters $T$ for FPA and $e$ for baseline randomized ablation control the corresponding method's accuracy-robustness trade-off. Specifically, a smaller value of $T$ and a larger value of $e$ entails better accuracy. This section emulates a defender that tunes FPA's and randomized ablation's hyperparameters to maximize the certified accuracy *at each individual robustness level individually*.

Tables 26 through 29 above report each method's certified accuracy across 10 comparable hyperparameter settings. For a given method, each hyperparameter setting provides a certified accuracy versus certified robustness curve (example curves are shown in Figures 4 and 5). This section considers each defense's certified accuracy *envelope*. Specifically, an envelope in mathematics represents the supremum of a set of curves. Intuitively, taking the certified accuracy envelope emulates maximizing a method's performance at each certified robustness level individually across the 10 hyperparameter settings.

Figures 2 and 3 visualize the certified accuracy envelopes in two ways. First, Figures 2a, 2b, 3a, and 3b visualize the envelope curves themselves. These figures also visualize the same naive baselines considered in Sec. H.3.1 above (e.g., always predict label 1 for classification and median $\text{med}\{y_i\}_{i=1}^n$ for regression). Second, Figures 2c, 2d, 3c, and 3d visualize the improvement in certified accuracy between FPA and the two versions of randomized ablation across the range of certified robustness levels. A positive value in these four subfigures entails that FPA outperformed the corresponding baseline (i.e., FPA had a larger certified accuracy), while a negative value entails the baseline outperformed FPA.

For CIFAR10 and MNIST, FPA with run-off's envelope had larger certified accuracy than the envelope of both versions of baseline RA across the entire certified robustness range (x-axis). Specifically, for Levine and Feizi's [LF20b] version of RA, FPA with run-off's certified accuracy advantage was as large as 14.17 and 24.28 percentage points (pp) for CIFAR10 and MNIST, respectively. For Jia et al.'s [Jia+22b] version of RA, FPA with run-off's certified accuracy advantage was as large as 6.54pp and 12.74pp for CIFAR10 and MNIST, respectively.

For regression datasets Weather and Ames, FPA's envelope had larger certified accuracy than the envelope of both versions of baseline RA across most of the certified accuracy range. At the largest robustness values, [Jia+22b] marginally outperformed both FPA and the naive baseline by <2pp. At smaller certified robustness values, FPA outperformed Jia et al.'s [Jia+22b] version of RA by up to 21.9pp and 17.4pp for Weather and Ames, respectively.

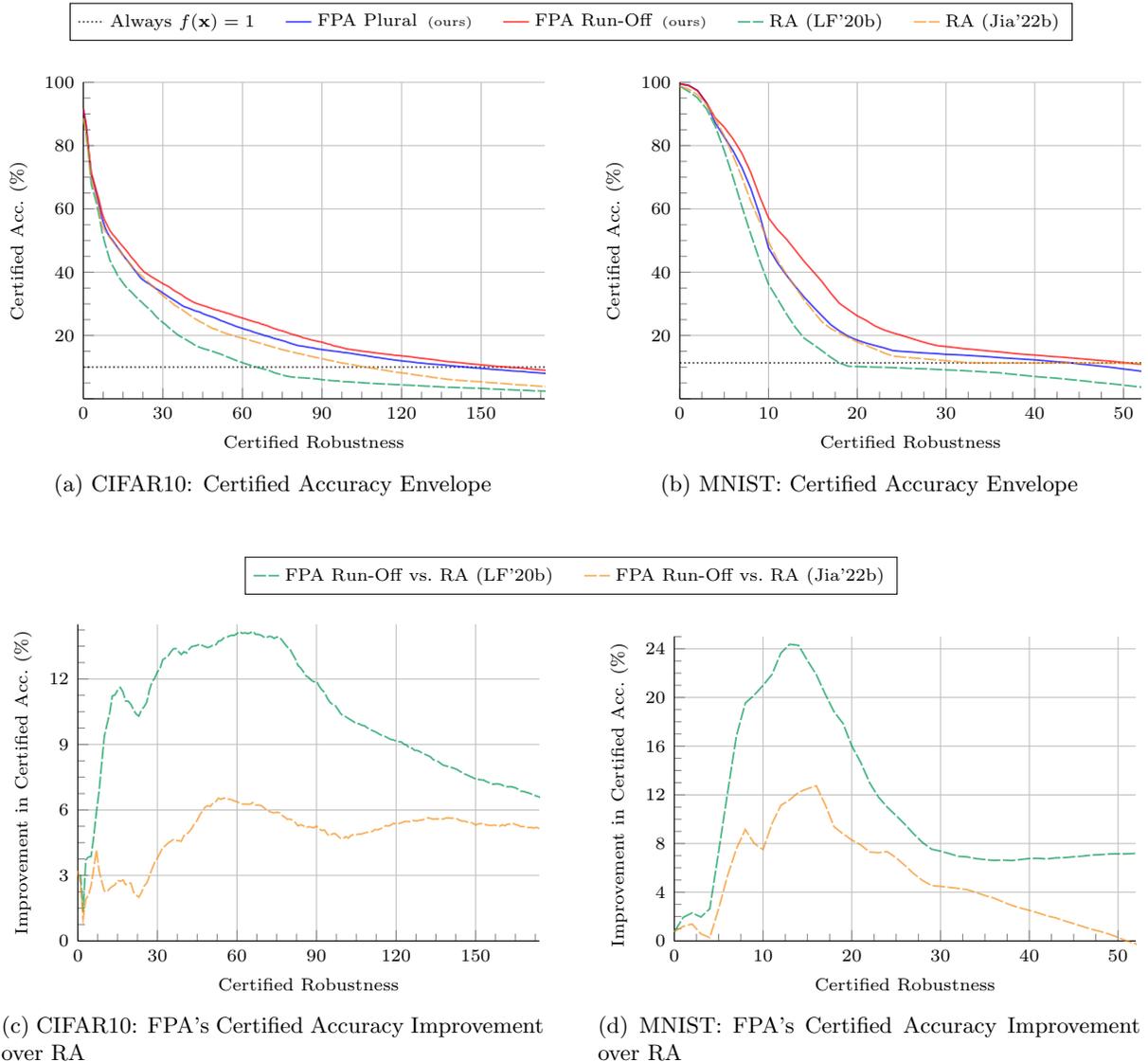

Figure 2: **Classification certified accuracy envelope** for datasets CIFAR10 ($d = 1024$) and MNIST ($d = 784$) for feature partition aggregation (FPA) and baseline randomized ablation (RA). Each method's envelope considers the corresponding hyperparameters in Tables 26 and 27, emulating a certified defense where the hyperparameters are roughly tuned to maximize the certified accuracy at each robustness level. Subfigures 2a and 2b visualize each method's certified accuracy envelope (larger is better); also shown in these subfigures is a naive baseline where the decision function always predicts label $f(\mathbf{x}) = 1$. Subfigures 2c and 2d visualize the improvement in certified accuracy when using FPA with the run-off decision function over the two randomized ablation baselines from Levine and Feizi [LF20b] and Jia et al. [Jia+22b]. FPA with run-off's certified accuracy advantage over Jia et al.'s version of RA was as large as 6.54pp and 12.74pp for CIFAR10 and MNIST, respectively. FPA's performance advantage was even larger over Levine and Feizi's [LF20b] version of RA. The envelope plots' underlying numerical values are provided in Table 26 for CIFAR10 and Table 27 for MNIST.

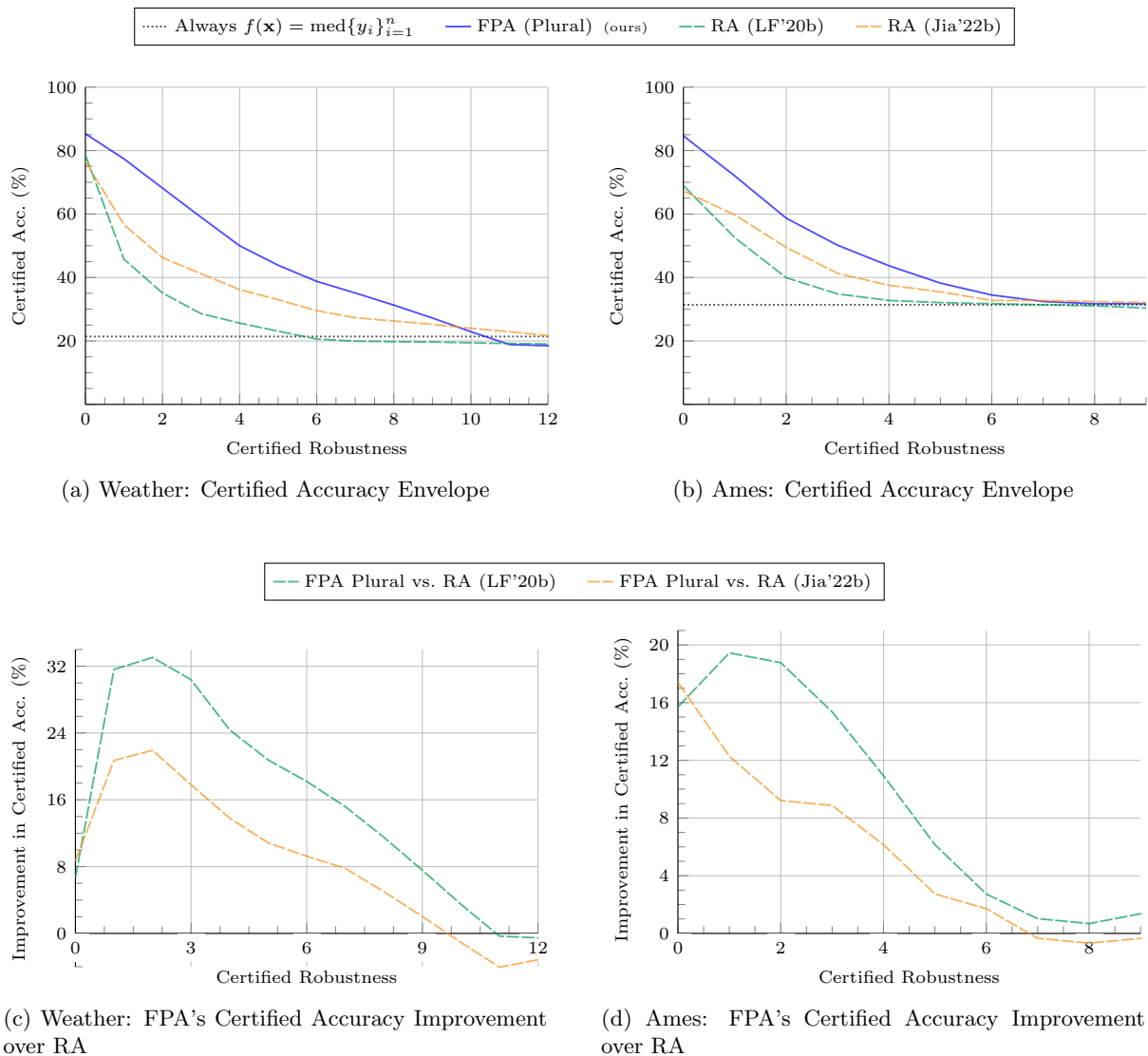

Figure 3: **Regression certified accuracy envelope** for the Weather [Mal+21] ($d = 128$) and Ames [De 11] ($d = 352$) datasets for feature partition aggregation (FPA) and baseline randomized ablation (RA). Each method's envelope considers the corresponding hyperparameters in Tables 28 and 29, emulating a certified defense where the hyperparameters are tuned to maximize each robustness level's certified accuracy. Subfigures 3a and 3b visualize each method's certified accuracy envelope (larger is better); also shown in these subfigures is a naive baseline that always predicts the median training data target value. Subfigures 3c and 3d visualize the improvement in certified accuracy when using FPA (with plurality voting) as the decision function over the two randomized ablation baselines from Levine and Feizi [LF20b] and Jia et al. [Jia+22b]. FPA with run-off's certified accuracy advantage over Jia et al.'s version of RA was as large as 21.9pp and 17.4pp for Weather and Ames, respectively. FPA's performance advantage was even larger over Levine and Feizi's [LF20b] version of RA. FPA outperforms randomized ablation for smaller certified robustness values, while Jia et al.'s [Jia+22b] version of RA marginally outperformed both FPA and the naive baseline at larger robustness values. The envelope plots' underlying numerical values are provided in Table 28 for Weather and Table 29 for Ames.

## H.4 Feature Partition Aggregation Model Count Hyperparameter Analysis

Figure 4 visualizes the certified accuracy[15] of FPA for multiple $T$ values for all four datasets in Section 6. Figure 4 also visualizes each dataset's non-robust (i.e., uncertified) accuracy (······), where a single model is trained on all features.

These experiments used the same evaluation setup as Section 6. For classification datasets CIFAR10 [KNH14] and MNIST [LeC+98], results using plurality voting and run-off decisions are provided. For regression datasets Weather [Mal+21] and Ames [De 11], plurality voting and run-off are identical; we provide regression results for both LightGBM [Ke+17] and linear submodels.

The exact effect of $T$ differs by dataset. As a general rule, increasing $T$ decreases the ensemble's classification accuracy (although not necessarily monotonically in the case of deterministic partitioning). Figure 4 visualizes this basic relationship where increasing $T$ generally increases the maximum certified robustness.

---

[15]*Certified accuracy* w.r.t. $\psi \in \mathbb{N}$ quantifies the fraction of correctly-classified test instances with certified robustness at least $\psi$.

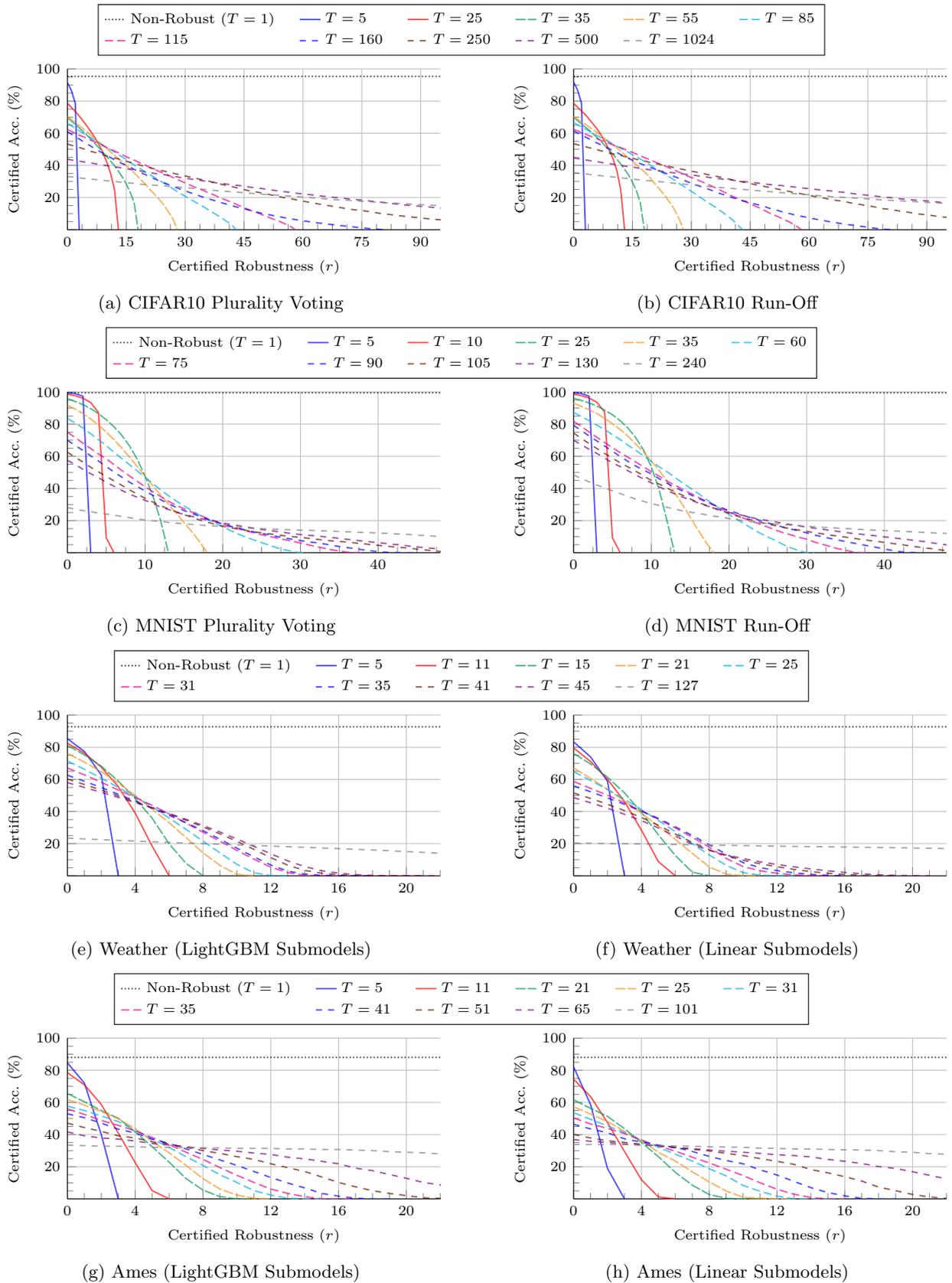

Figure 4: **Effect of Submodel Count $T$ on the Certified Feature Robustness**: Mean certified accuracy (%) for our sparse defense, feature partition aggregation (FPA), across different submodel counts ($T$). The non-robust accuracy (⋯) visualizes the classification accuracy of a single model ($T = 1$) trained on all features; these single model prediction results are provided only for reference. For all four datasets, increasing $T$ decreases the classification accuracy but increases the maximum certifiable robustness.

A43

## H.5 Randomized Ablation Number of Kept Features ($e$) Hyperparameter Analysis

As discussed in Sections 3, 6, and C, $\ell_0$-norm certified defense randomized ablation (RA) is based on randomized smoothing where predictions are averaged across multiple randomly perturbed inputs. For each input, $e \in \mathbb{N}$ features in $\mathbf{x} \in \mathcal{X}$ are randomly selected to be kept at their original value, and the rest of the features are ablated, i.e., marked as unused or "turned off." In short, $e$ controls RA's accuracy versus robustness tradeoff where larger $e$ increases the classifier's accuracy at the expense of a smaller maximum achievable robustness ($\rho$). By contrast, a small $e$ decreases the model's accuracy but increases the maximum achievable certified robustness.

Figure 5 visualizes RA's certified accuracy[16] for a range of $e$ settings for all four datasets in Sec. 6, namely CIFAR10 [KNH14], MNIST [LeC+98], Weather [Mal+21], and Ames [De 11]. Fig. 5 also visualizes each dataset's non-robust accuracy (······), where a single non-smoothed model is trained on all features.

---

[16] *Certified accuracy* w.r.t. $\psi \in \mathbb{N}$ quantifies the fraction of correctly-classified test instances with certified robustness at least $\psi$.

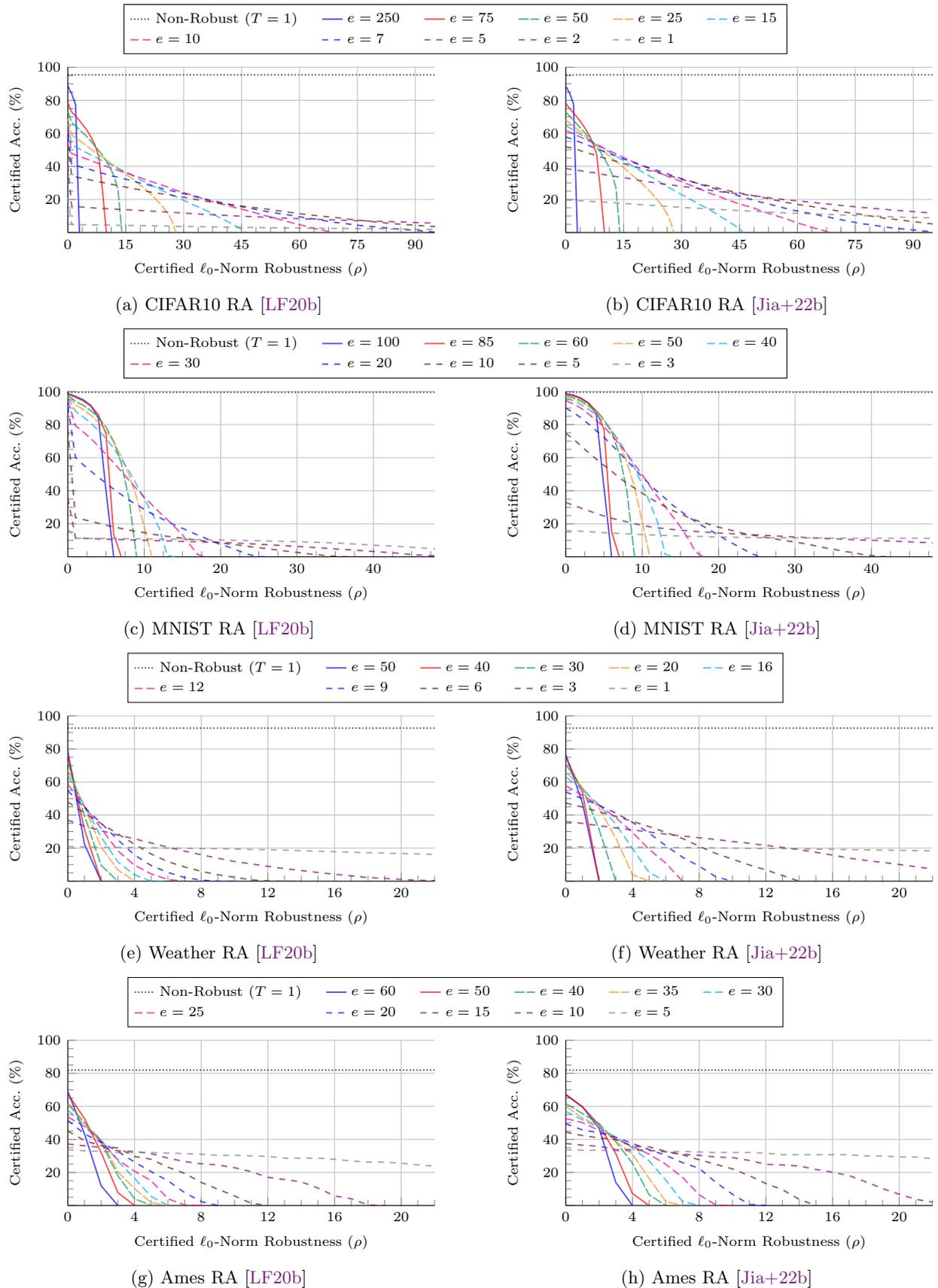

Figure 5: **Effect of the Number of Kept Features ($e$) on RA's Certified $\ell_0$-Norm Robustness**: Mean certified accuracy (%) for baseline randomized ablation across different quantities of kept pixels ($e$). Non-robust accuracy (······) visualizes the peak accuracy of a single model ($T = 1$) trained on all features; these single model predictions are provided only for reference.

## H.6 Comparing FPA Plurality Voting and Run-Off Certification

Sec. 4 proposes two decision functions for FPA, namely *plurality voting* (4.1) and *run-off elections* (4.2). Both decision functions can be used to certify feature robustness (Def. 1). However, the two decision functions' guarantees may differ significantly in *size*.[17]

Below, Figures 6, 7, and 8 show the improvement in FPA's *certified accuracy*[18] for CIFAR10 and MNIST when robustness certification is enhanced using run-off elections. Specifically, Figure 6 visualizes the improvement in certified accuracy when run-off is used instead of plurality voting for each certified robustness value $r$, where a positive value denotes that run-off performed better, while a negative value entails that plurality voting had better performance. Across almost all values of $r$ and submodel counts $T$, combining FPA with run-off improved the certified accuracy, with performance improvements as large as 12.3 percentage points (pp) for MNIST and 3.8pp for CIFAR10.

Figures 7 and 8 visualize the performance of FPA with plurality voting directly against that of FPA with run-off for CIFAR10 and MNIST, respectively.

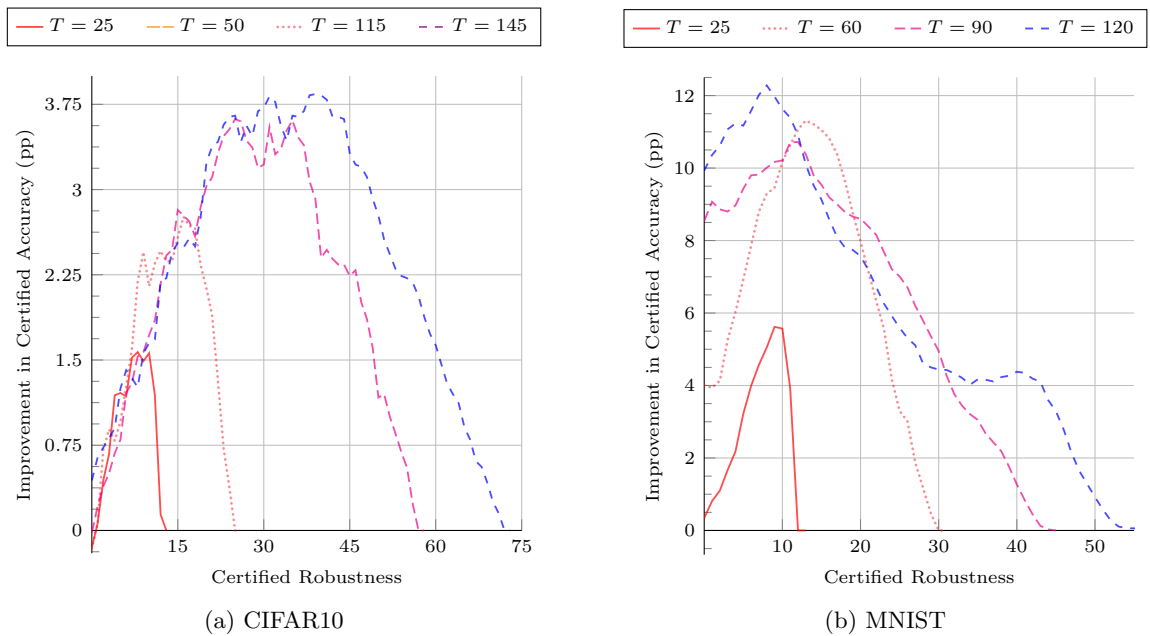

Figure 6: **Improvement in FPA's Certified Accuracy with Run-Off Elections for CIFAR10 and MNIST**: Effect of the decision function on FPA's certified accuracy. 0 on the y-axis denotes the baseline performance of FPA using plurality voting-based (Sec. 4.1). A positive value denotes that run-off-based certification improves FPA's certified accuracy, while a negative value denotes run-off degrades performance. Across almost all submodel counts $T$ and certified robustness levels $r$, run-off improves FPA's certified accuracy, with improvements up to 12.3 percentage points (pp) on MNIST and 3.8pp on CIFAR10.

---

[17]Recall that run-off and plurality voting are identical for regression datasets Weather [Mal+21] and Ames [De 11] since Hammoudeh and Lowd's [HL23b] reduction is from certified regression to certified *binary* classification.

[18]*Certified accuracy* w.r.t. $\psi \in \mathbb{N}$ quantifies the fraction of correctly-classified test instances with certified robustness at least $\psi$.

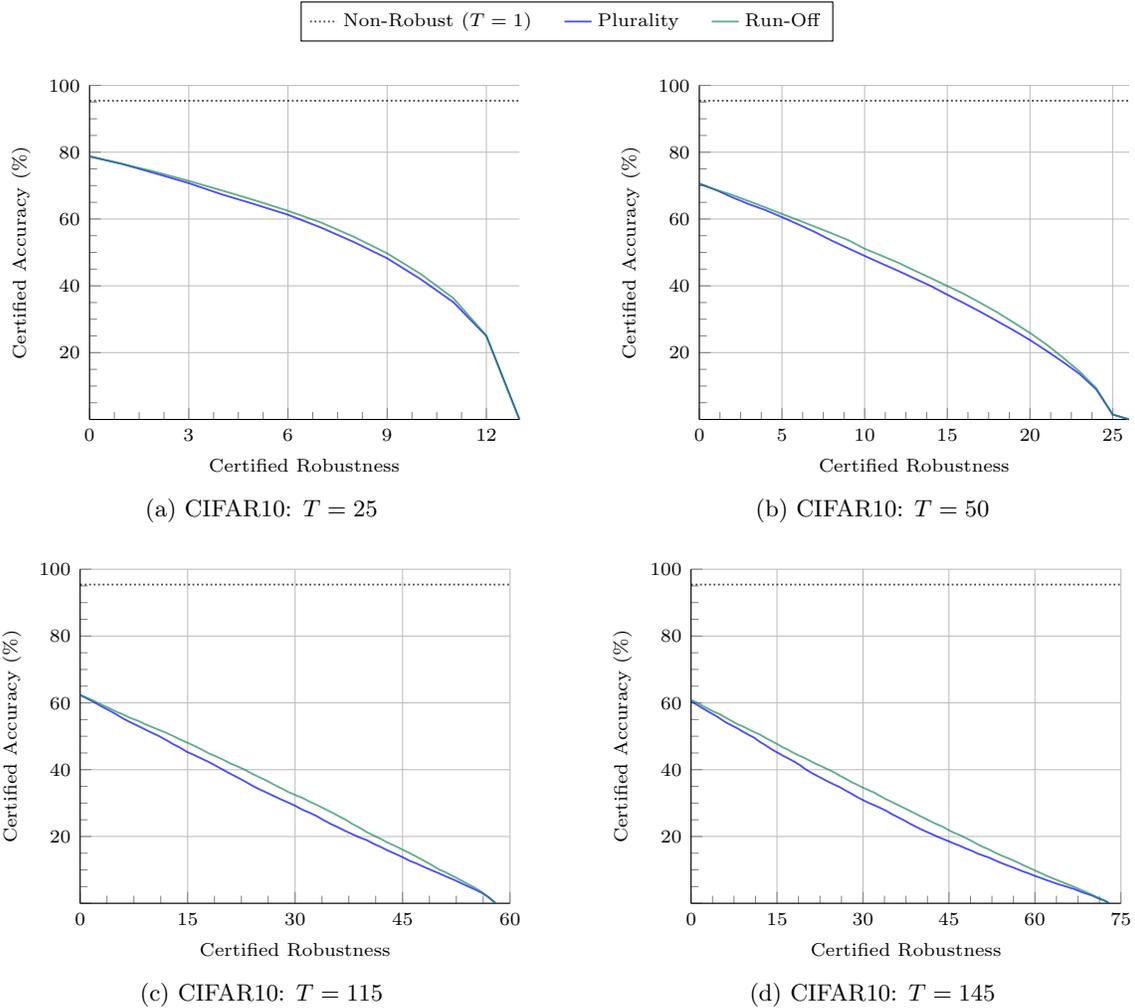

Figure 7: **Effect of the Decision Function on FPA's CIFAR10 Certified Accuracy**: Comparison of the certified accuracy of FPA when using the plurality-voting decision function (Sec. 4.1) versus the run-off decision function (Sec. 4.2). Across all model counts ($T$) and certified robustness levels ($r$), run-off improved the certified accuracy, with the maximum improvement up to 3.8 percentage points on CIFAR10.

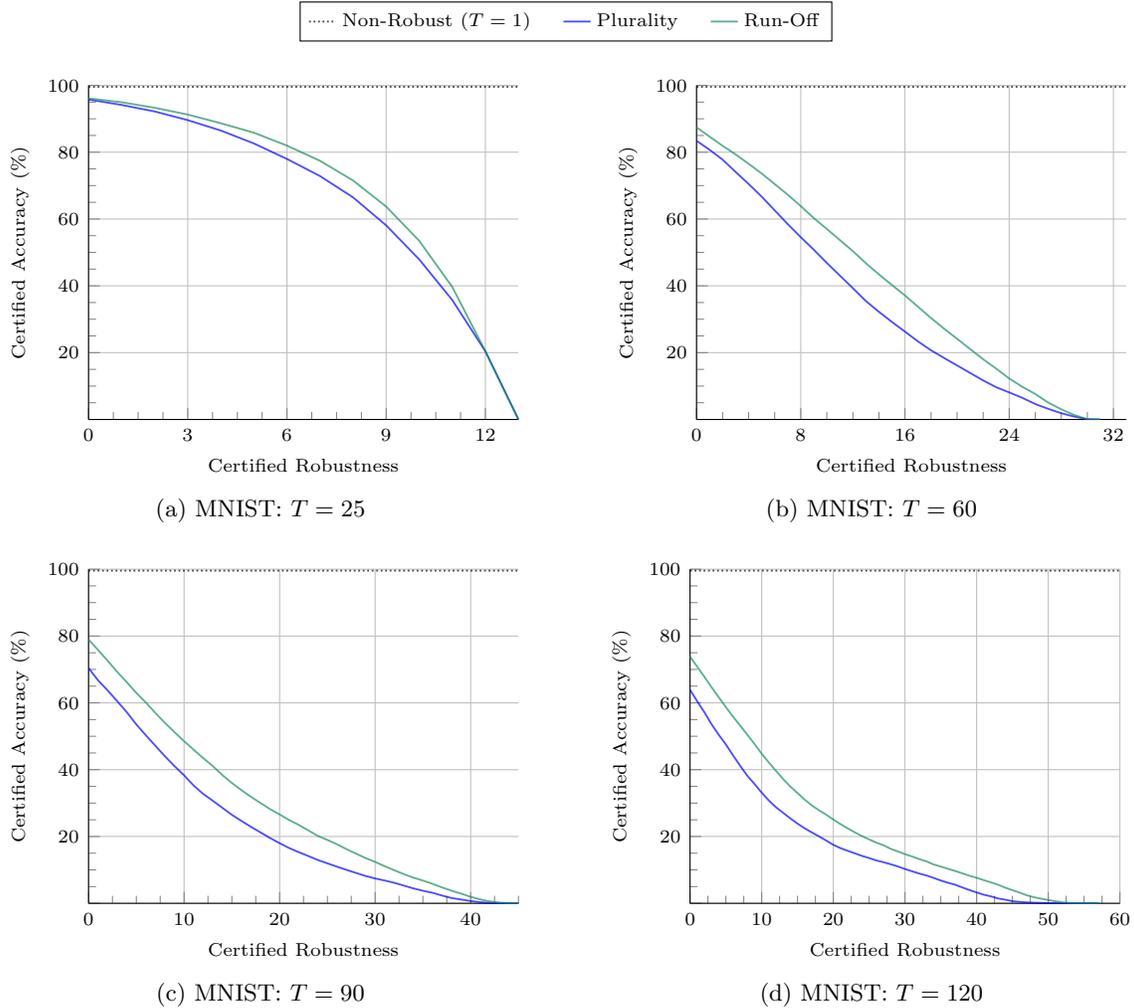

Figure 8: **Effect of the Decision Function on FPA's MNIST Certified Accuracy**: Comparison of the certified accuracy of FPA when using the plurality-voting decision function (Sec. 4.1) versus the run-off decision function (Sec. 4.2). Across all model counts ($T$) and certified robustness levels ($r$), run-off improved the certified accuracy, with the maximum improvement up to 12.3 percentage points on MNIST.

## H.7 Random vs. Deterministic Feature Partitioning

Sec. 5 proposes two paradigms for partitioning the $d$ features between the $T$ submodels. The first option, *balanced random partitioning*, assigns each submodel approximately the same number of features uniformly at random. The second option, *deterministic partitioning*, uses a deterministic scheme to decide the set of features assigned to each submodel.

In the main paper, we propose what we term "strided partitioning," a deterministic partitioning strategy where for submodel $f_t$, the corresponding feature set is

$$\mathcal{S}_t = \{j \in [d] : j \mod T = t - 1\}. \tag{33}$$

Strided partitioning is specifically targeted toward structured, two-dimensional feature sets (e.g., images). Striding ensures that each subset $\mathcal{S}_t$ contains feature information across the 2D grid.

Alternative deterministic strategies we considered include "patching," where the image is broken up into a grid of disjoint 2D patches. Each submodel is then trained on a different subpatch. Patching performed exceptionally poorly (much worse than random partitioning) because, in short, each submodel is trained on highly correlated pixels limiting the information available to each. Moreover, many of the submodel patches contained no information from the highly salient center pixels.

A third deterministic partitioning strategy we considered assigned pixels to each submodel starting from the center of the image. In essence, this "spiral" strategy renumbers the pixels, defining the center pixel as feature 1 and then assigning pixels indices in order based on their Manhattan distance from the center. The intuition behind the "spiral" strategy is to maximize the number of highly-salient center pixels used by each submodel.

Figure 9 compares FPA with plurality voting's certified accuracy using random partitioning versus the consistently best performing deterministic strategy – striding. We consider three datasets from Sec. 6. CIFAR10 [KNH14] ($d = 1024$) and MNIST [LeC+98] ($d = 784$) are image classification datasets, while Weather [Mal+21] is a tabular regression dataset. For all three datasets, the partitioning strategy used in Sec. 6 is shown as a solid line, while the other partitioning strategy is shown as a dashed line. Below we briefly summarize the key takeaways from Fig. 9.

**Takeaway #1**: *Deterministic feature partitioning significantly improves FPA's performance on vision datasets.* For both CIFAR10 and MNIST, deterministic (strided) feature partitioning significantly outperforms random partitioning. For example, on CIFAR10 and MNIST $T = 25$, strided partitioning improves the mean certified accuracy by up to 15.6% and 11.9%, respectively.

**Takeaway #2**: *Deterministic partitioning's benefits decrease with increasing submodel count.* For CIFAR10 with $T = 115$ submodels, deterministic partitioning improved FPA's mean certified accuracy by at most 5.8%; in contrast, for CIFAR10 with $T = 25$ submodels, deterministic partitioning improved performance by up to 15.6%. A similar trend is observed for MNIST. As $T$ increases, each submodel is trained on (substantially) fewer pixels. As feature sparsity increases, the benefit of a regular pixel pattern decreases.

**Takeaway #3**: *Deterministic and random partitioning perform comparably for the Weather dataset.* Tabular features are generally unstructured or, in some cases, loosely structured. Intuitively, there is no consistent advantage in ensuring that the tabular features considered by each submodel are well-spaced. A deterministic tabular feature partition can be viewed as a random variable drawn from the set of all random partitions. Some deterministic partitions outperform the mean random partition; other deterministic partitions underperform the mean random partition. We see this behavior in Fig. 9c, where for $T = 11$, strided partitioning outperforms balanced random while for $T = 21$, balanced random is better. For $T = 31$, strided and random partitioning perform similarly.

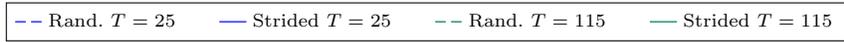
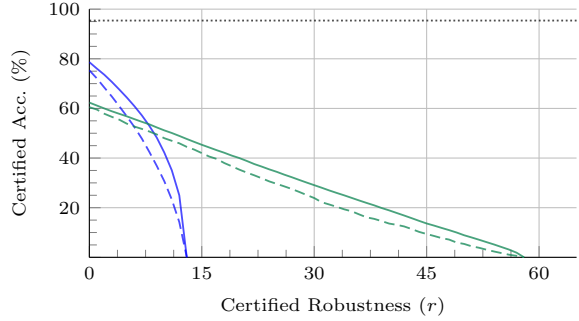

(a) CIFAR10

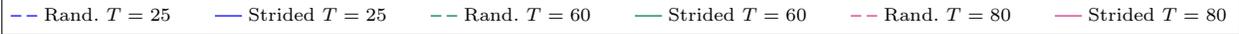
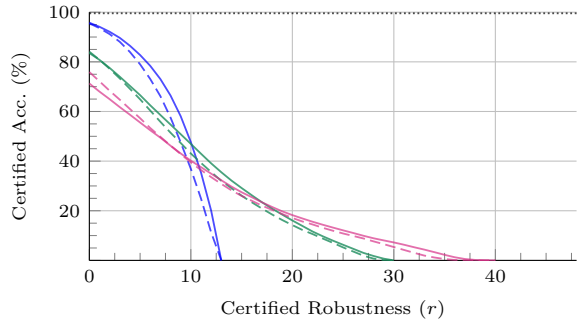

(b) MNIST

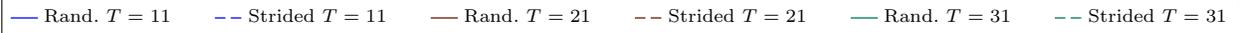
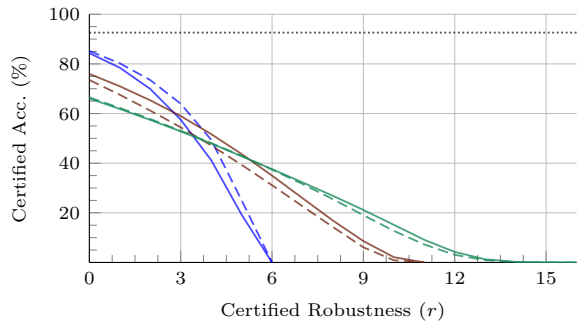

(c) Weather (LightGBM)

Figure 9: **Effect of the Feature Partitioning Paradigm on Certified Feature Robustness**: Certified accuracy for feature partition aggregation (FPA) with plurality voting across different feature partitioning paradigms. Uncertified accuracy (······) visualizes the peak accuracy of a single model ($T = 1$) trained on all features; these single model predictions are completely non-robust and provided only for reference. For each dataset, the feature partitioning strategy used in Sec. 6 is shown as a solid line. The alternate feature partitioning strategy is shown in the same color but as dashed lines.

## H.8 Model Training Time

This section summarizes the (sub)model training times of feature partition aggregation (FPA) and baseline randomized ablation (RA). These experiments were performed on a desktop system with a single AMD 5950X 16-core CPU, 64GB of 3200MHz DDR4 RAM, and a single NVIDIA 3090 GPU.

Recall that certified defenses against sparse attacks – both ours and randomized ablation – trade off accuracy against robustness. Put simply, larger certified guarantees are generally achieved at the expense of reduced accuracy (and vice versa). To capture the nature of this trade-off, supplemental Sec. H.3 reports performance at various hyperparameter settings.

Hyperparameter settings can affect (sub)model training times so Table 30 reports the mean training times for two hyperparameter settings per method – one a higher accuracy setting and the other a more robust setting. For FPA, we separately report the mean training time for a single submodel as well as the total training time of the entire ensemble. Model training for randomized ablation used Levine and Feizi's [LF20b] original source code for MNIST and CIFAR10. Levine and Feizi's code was modified to support the Weather and Ames datasets, which are not included in RA's published implementation.

For the tabular Weather and Ames dataset, FPA was 18× to 90× faster to train than randomized ablation. Randomized ablation is only compatible with model types that support stochastic, ablated training. By contrast, FPA supports any submodel type, including LightGBM gradient-boosted decision trees (GBDTs) used here.

For vision datasets MNIST and CIFAR10, FPA's total ensemble training times are 2.1× to 11× slower than randomized ablation. Note that the training of each FPA submodel is fully independent. In other words, FPA ensemble training is embarrassingly parallel with up to $T$ degrees of parallelism. Provided sufficient hardware, an FPA ensemble can be (significantly) faster to train in parallel than a randomized ablation model, as evidenced by Table 30's single FPA submodel training times.

Training is identical for both Levine and Feizi's [LF20b] and Jia et al.'s [Jia+22b] versions of randomized ablation (RA).

Table 30: **Model Training Time**: Mean model training time (in seconds) for feature partition aggregation (FPA) and baseline randomized ablation. For each dataset, we report the training times for two hyperparameter settings – one that achieves higher certified accuracy and the other that achieves larger certified robustness. For FPA, the time to train a single submodel and the total time to train the entire ensemble are reported. "<1" denotes that training took less than 1 second.

| Dataset | Random. Abl. | | FPA (ours) | | |
|---|---|---|---|---|---|
| | $e$ | Time | $T$ | Single | Total |
| CIFAR10 | 75 | 6,278 | 25 | 541 | 13,526 |
| | 25 | 6,085 | 115 | 544 | 62,613 |
| MNIST | 45 | 904 | 25 | 153 | 3,834 |
| | 20 | 883 | 60 | 161 | 9,669 |
| Weather | 20 | 5,186 | 11 | 13 | 141 |
| | 8 | 5,210 | 31 | 9 | 278 |
| Ames | 50 | 63 | 11 | <1 | 1 |
| | 15 | 64 | 51 | <1 | <1 |